\documentclass[11pt]{article}

\usepackage[final]{acl}

\usepackage{times}
\usepackage{latexsym}

\usepackage[T1]{fontenc}

\usepackage[utf8]{inputenc}

\usepackage{microtype}

\usepackage{inconsolata}

\usepackage{graphicx}

\usepackage{booktabs}
\usepackage{array}
\usepackage{tcolorbox}
\usepackage{cuted}
\tcbuselibrary{breakable}
\usepackage{amsmath}
\usepackage{enumitem}
\usepackage{multirow}
\usepackage{graphicx}
\usepackage{pdflscape}
\usepackage{rotating}
\usepackage{float}
\usepackage{subcaption}
\usepackage{hyperref}
\usepackage{xcolor}

\author{
  \textbf{Xiangyu Wang}\textsuperscript{1,*}, 
  \textbf{Jin Wu}\textsuperscript{2,3,*}, 
  \textbf{Xiaoyu Li}\textsuperscript{4}, 
  \textbf{Chanjin Zheng}\textsuperscript{2,$\dagger$},
  \textbf{Yifeng Zhou}\textsuperscript{5} \\
  \textsuperscript{1}Department of Educational Psychology, East China Normal University, \\
  \textsuperscript{2}Shanghai Institute of Artificial Intelligence for Education, East China Normal University, \\
  \textsuperscript{3}School of Computer Science and Technology, East China Normal University \\
  \textsuperscript{4}
School of Education and Intelligent Education Research Center, Yangzhou University \\
  \textsuperscript{5}School of Data Science and Engineering, East China Normal University \\
  \small{\texttt{\{51274118009, 52275901018\}@stu.ecnu.edu.cn}, \texttt{chjzheng@dep.ecnu.edu.cn}}
}

\title{Decoupled Analysis-Judging: An Automated Creativity Evaluator Using LLMs in Complex Multi-step Creativity Tasks}

\begin{document}

\maketitle

\makeatletter
\newcommand\authorfootnote[2]{%
  \def\@thefnmark{#1}%
  \@footnotetext{#2}%
}
\makeatother

\authorfootnote{*}{Equal contribution.}
\authorfootnote{$\dagger$}{Corresponding author.}

\begin{abstract}
Automated evaluation of creativity tasks remains challenging for LLM-as-a-Judge, as LLM is susceptible to biases such as verbosity bias and leniency bias. Such limitations are particularly evident in Contextually-Grounded and Procedurally-Structured Tasks (CGPST), a complex multi-step creativity task where inter-step dependencies, highly subjectivity, and wide scoring ranges lead to more unstable and biased judgments. Existing approaches either rely on task-specific training or directly apply LLM-as-a-Judge, both of which struggle to ensure reliable evaluation under such complexity. To bridge these gaps, we propose \textbf{CreaEval}, an automated \textbf{crea}tivity \textbf{eval}uator for CGPST that decouples typical LLM-as-a-Judge into analysis and judging. Correspondingly, CreaEval involves two critical phases: \textbf{Memory-augmented Analysis}, a SoT-LLM converts multi-step responses into structured evaluation evidence, incorporating cross-step memory; and \textbf{Evidence-based Judging}, a Judge-LLM uses the extracted evidence for judging without accessing raw responses. 
Comprehensive experiments show that CreaEval achieves an average performance improvement of 22.74\% over the second-best baselines across CGPST and two classic simple creativity tasks, demonstrating its generalizability. The code is available at \url{https://github.com/Jaong/CreaEval}.
\end{abstract}

\section{Introduction}
The creativity of Large Language Models (LLMs) has attracted increasing attention, with applications in creative writing, novel mathematical reasoning, and other creative domains \citep{10.1145/3706598.3714198, lin-etal-2025-creativity, Ye_Gu_Zhao_Yin_Wang_2025}. However, current automated evaluation methods mainly focus on simple creativity tasks, such as the Alternate Uses Task (AUT) \citep{lu2024llmdiscussionenhancingcreativity} and the Torrance Test of Creative Thinking (TTCT) \citep{10.1145/3706598.3714198}, while the automated evaluation of complex multi-step creativity tasks remains underexplored.

Among creativity tasks, Contextually-Grounded and Procedurally-Structured tasks (\textbf{CGPST}) \citep{wang2026teamllmhumanliketeamorientedcollaboration} is particularly challenging due to its high complexity, including multi-step and scenario-based features. Each task on CGPST is grounded in a complete future scenario and needs to solve multiple interdependent steps \citep{Treffinger1995, treffinger2012four}.

Existing automated evaluation methods for the above creativity tasks mainly fall into two categories. The first involves training task-specific models, which require additional training resources and annotated datasets, making them costly to develop \citep{do-etal-2024-autoregressive, wang-liu-2025-mes, Li_Pan_2025}. The other is LLM-as-a-Judge, a training-free approach that has become increasingly dominant due to the strong capabilities of LLMs \citep{NEURIPS2023_91f18a12, li-etal-2025-automated-creativity, Ye_Gu_Zhao_Yin_Wang_2025}. However, directly applying LLM-as-a-Judge to CGPST yields low agreement, as LLMs are often susceptible to biases such as verbosity bias and leniency bias, making it difficult to accurately capture subtle score differences in subjective dimensions \citep{wang2026teamllmhumanliketeamorientedcollaboration}.

\begin{table*}
    \centering
    \scriptsize
    \renewcommand{\arraystretch}{1.5}
    \begin{tabular}{>{\footnotesize}p{2cm}>{\footnotesize}p{5.1cm}>{\footnotesize}p{4.7cm}>{\footnotesize}p{2.2cm}}
        \toprule
        \textbf{Step} & \textbf{Requirement} & \textbf{Dimensions (Score Range)} & \textbf{Targeted Ability} \\
        \midrule
        Step-1: Identify Challenges & Identify up to 8 reasonable challenges based on the future scenario. & Fluency (0-8), Flexibility (0-8), Elaboration (0-16), Originality (0-16) & Divergent Thinking \\
        Step-2: Select an Underlying Problem & Select the most promising and meaningful challenge from Step-1 as the underlying problem. & Integrity (0-10), Focus (1-10), Adequacy (1-10) & Convergent Thinking \\
        Step-3: Produce Solutions & Generate up to 8 solutions for the underlying problem from Step-2. & Fluency (0-8), Flexibility (0-8), Elaboration (0-16), Originality (0-16) & Divergent Thinking \\
        Step-4: Select Criteria & Generate 5 evaluation criteria for the solutions from Step-3. & Correctly Written (0-5), Relevance (0-15) & Critical Thinking \\
        Step-5: Apply Criteria to Top Solution & Rank the solutions from Step-3 using the criteria from Step-4 and select the highest-scoring solution. & Correctly Used (0-5) & Logical Thinking \\
        Step-6: Develop an Action Plan & Develop the top solution from Step-5 into a comprehensive action plan to address the underlying problem from Step-2. & Relevance (1-5), Effectiveness (1-5), Criteria (1-5), Impact (1-5), Humaneness (1-5), Development (1-10) & Comprehensive Problem-Solving \\
        \bottomrule
    \end{tabular}
    \caption{Step-wise Information on CGPST. See Appendix~\ref{App:CGPST Benchmark Details} for a full description.}
    \label{Table:Step-wise Information on CGPST}
\end{table*}

Applying automated evaluation to CGPST presents several key challenges compared to other simple creativity tasks. \textbf{(1)} The task is \textbf{highly complex} compared to traditional creativity tasks. Each task requires evaluating multiple interdependent steps based on a scenarios, with each step assessed across multiple dimensions. \textbf{(2)} The task is \textbf{inherently subjective}, as there are no fixed answers and earlier responses influence subsequent ones, resulting in highly diverse outputs \citep{Zhao2025, wang2026teamllmhumanliketeamorientedcollaboration}. \textbf{(3)} Some dimensions involve \textbf{large scoring ranges} (e.g., 10-level scoring), which further increases scoring instability.

To address these challenges, we propose \textbf{CreaEval}, an automated \textbf{crea}tivity \textbf{eval}uator using LLMs in complex multi-step creativity tasks such as CGPST. Inspired by human evaluation practices, where raters first analyze responses across each dimension before assigning rubric-based scores \citep{klein1998analytic, harsch2013comparing}, CreaEval decouples the evaluation process into two phases: \textbf{Memory-augmented Analysis} and \textbf{Evidence-based Judging}. In the first phase, a SoT-LLM incrementally organizes raw responses into structured intermediate evaluation evidence in the form of Structure-of-Thought (SoT) \citep{qi-etal-2025-sot, wang2026t2sbenchstructureofthoughtbenchmarking}. Considering the interdependencies among CGPST steps, CreaEval further introduces a memory mechanism to maintain cross-step coherence during analysis. In the second phase, each Judge-LLM assigns scores based on the extracted evidence and predefined rubrics, without accessing the original responses, thereby enabling evidence-grounded scoring.

Compared with typical LLM-as-a-Judge, this decoupled design constrains judging with structured evidence, thereby narrowing the range of plausible judgments. This not only improves scoring accuracy and stability, but also mitigates verbosity and leniency biases. The contributions are threefold:

\begin{itemize}
\item We propose \textbf{CreaEval}, an novel automated \textbf{crea}tivity \textbf{eval}uator for complex and subjective creativity tasks such as CGPST that decouples typical LLM-as-a-Judge into Memory-augmented Analysis and Evidence-based Judging.

\item Extensive experiments show that CreaEval achieves an average human-LLM agreement of 0.64 (quadratic weighted kappa, QWK) on CGPST, outperforming the second-best baseline (supervised training) by 0.17.

\item Further analysis reveals that the decoupled design enhances scoring stability across dimensions and mitigates verbosity and leniency biases in LLM-as-a-Judge, providing new insights into evaluating creativity tasks.
\end{itemize}

\section{Related Work}

\subsection{Creativity Tasks}
Traditional creativity tasks originate from educational and psychological studies, such as AUT, which requires participants to generate as many novel uses as possible for a common object \citep{lu2024llmdiscussionenhancingcreativity, Zhao2025, AUT_1, AUT_2}, and TTCT, which assesses creativity through responses to open-ended and unconventional scenarios \citep{torrance1966torrance, 10.1145/3706598.3714198}. However, these tasks are typically single-step and structurally simple, limiting their ability to capture complex creative processes. Recently, \citet{wang2026teamllmhumanliketeamorientedcollaboration} proposed CGPST, a multi-step, scenario-based benchmark for evaluating creative problem-solving abilities of LLMs. As shown in Table~\ref{Table:Step-wise Information on CGPST}, each task is grounded in a complete future scenario and requires LLMs to sequentially solve six interdependent steps, involving diverse abilities \citep{Treffinger1995, treffinger2012four}. Compared to traditional creativity tasks, CGPST exhibits significantly higher complexity.

\begin{figure*}
    \centering
    \includegraphics[width=\textwidth]{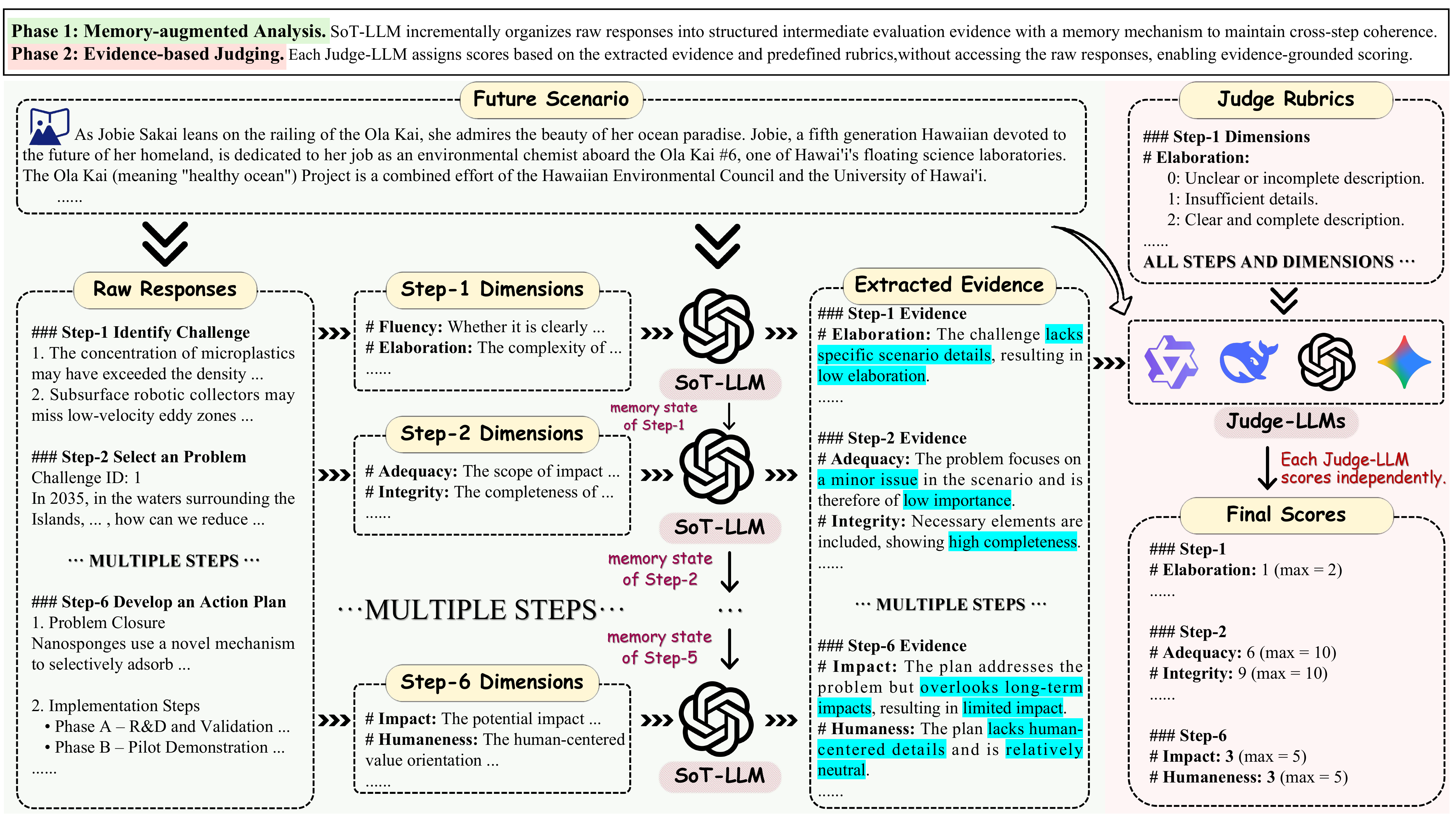} 
    \caption{Our proposed CreaEval framework. CreaEval decouples the evaluation process into two phases: \textbf{1) Memory-augmented Analysis.} SoT-LLM first extracts intermediate evaluation evidence (blue) based on the responses and dimensions in a step-by-step manner. \textbf{2) Evidence-based Judging.} Each Judge-LLM then performs scoring based on the extracted evidence and predifined rubrics without accessing raw responses.}
    \label{Figure:Our proposed CreaEval framework}
\end{figure*} 

\subsection{LLM-as-a-Judge for Creativity Evaluation}
Recently, several studies have explored automated evaluation methods for creativity tasks \citep{NEURIPS2023_91f18a12, liang-etal-2024-encouraging, li-etal-2025-automated-creativity, Ye_Gu_Zhao_Yin_Wang_2025}. For example, \citet{Zhao2025} leverages GPT-4 to generate TTCT-inspired datasets and employs LLMs for scoring responses. \citet{lu2024llmdiscussionenhancingcreativity} applies LLM-as-a-Judge to AUT tasks, achieving an average human–LLM agreement of 0.49 (Kendall’s $\tau$) across four dimensions, even surpassing inter-human agreement (0.39), suggesting the potential of LLM-as-a-Judge to improve scoring reliability in creativity tasks. However, existing studies mainly focus on relatively simple and traditional creativity benchmarks. Directly applying LLM-as-a-Judge to more complex tasks such as CGPST remains challenging. For instance, \citet{wang2026teamllmhumanliketeamorientedcollaboration} shows that a direct LLM-as-a-Judge approach with few-shot prompting achieves only 0.31 (pearson correlation coefficient, PCC) human–LLM agreement, highlighting the need for more reliable and effective evaluation frameworks for complex multi-step creativity tasks.

\section{Methodology}
Existing LLM-as-a-Judge fail to decouple analysis from judging, instead directly mapping from raw responses $R^s=\{R^s_t\}_{t=1}^{N}$ ($N$=6) to a score list $S$. This leads to the LLM's inability to effectively capture key intermediate evidence when handling complex creativity tasks. To address this limitation, we propose \textbf{CreaEval}, a novel evaluation framework that explicitly models the evaluation process in a structured manner, decoupling analysis and judging. As illustrated in Figure~\ref{Figure:Our proposed CreaEval framework}, CreaEval consists of two sequential phases: (1) Memory-augmented Analysis; and (2) Evidence-based Judging.

\subsection{Phase 1: Memory-augmented Analysis}
Given the raw response $R^s_t$ and the corresponding dimension description $D_t$ in Step-$t$, SoT-LLM performs structured evidence extraction in a step-by-step manner. Specifically, it iteratively processes each step of the CGPST by jointly considering the task scenario $\textit{Scenario}$, the step response $R^s_t$, and the evaluation dimensions $D_t$. Within Step-$t$, SoT-LLM in CreaEval extracts multi-dimensional evidence represented as a structured mapping 
$e_t = \left\{ \langle k_i, v_i \rangle \right\}_{i=1}^{|\mathcal{D}_t|}$, where each pair $\langle k_i, v_i \rangle$ denotes an evaluation dimension $k_i$ and its corresponding evidence $v_i$. This progressive design prevents SoT-LLM from being overwhelmed by all subjective information at once, thereby improving the reliability and consistency of the extracted evidence.

The steps of CGPST exhibit strong temporal dependencies, as responses in later steps are conditioned on decisions made in earlier ones. For instance, the solution proposed in Step-3 is designed to address the problem identified in Step-2. To model these dependencies, CreaEval introduces a memory mechanism that preserves step-relevant memory state $m_t$. Specifically, in addition to generating evidence $e_t$, SoT-LLM produces a corresponding $m_t = Summary(R^s_t, M_{t-1})$, which summarizes the key information from $R^s_t$ and is passed as contextual input to subsequent steps as shown in Figure~\ref{Figure:Our proposed CreaEval framework}. This mechanism retains cross-step contextual dependencies, thereby augmenting the evidence extraction process and improving accuracy, defined as:
\begin{equation}
(e_t, m_t) =
f_{SoT}
(Scenario, R^s_t, D_t, M_{t-1})
\end{equation}

\noindent where \(e_t\) and \(m_t\) denote the evaluation evidence and memory state extracted by SoT-LLM in Step-\(t\), respectively. $D=\{D_t\}_{t=1}^{N}$ and $R^s=\{R^s_t\}_{t=1}^{N}$ denote the dimension information and raw responses across all steps, respectively. ${M_{t-1}}=\{m_j\}_{j=1}^{t-1}$ represents the accumulated memory of previous steps to preserve cross-step dependency.

\subsection{Phase 2: Evidence-based Judging}
In this phase, Judge-LLM performs scoring grounded in the structured evidence generated during Phase 1. Rather than accessing the raw responses $R^s$ directly, Judge-LLM takes the aggregated evidence $E$ , the scoring rubrics $R^u$, and \textit{Scenario} as input, and generates multi-step and multi-dimensional scores $S$ over the entire response in a single pass, defined as:
\begin{equation}
S = f_{Judge}(Scenario, E, D, R^u)
\end{equation}

\noindent where $E=\{e_t\}_{t=1}^{N}$ and $R^u=\{R^u_t\}_{t=1}^{N}$ denote the evidence extracted by SoT-LLM and the scoring rubrics for all steps, respectively.

As illustrated in Figure~\ref{Figure:Our proposed CreaEval framework}, consider the \textit{Adequacy} dimension in Step-2: the extracted evidence (e.g., ``\textit{low importance}'' and ``\textit{a minor issue}'') indicates weak performance along this dimension, upon which Judge-LLM assigns a relatively low score of 6 out of 10. This evidence-grounded judging enhances both the accuracy and stability of the final scoring.

\section{Experiment}
\subsection{Challenging Dataset}

We conduct experiments on the Contextually-Grounded and Procedurally-Structured Tasks (CGPST) dataset \citep{wang2026teamllmhumanliketeamorientedcollaboration}, a complex multi-step creativity benchmark. It possesses a strong psychological foundation based on the Future Problem Solving Program International (FPSPI) \citep{FPSPI_1, FPSPI_2, wang2026teamllmhumanliketeamorientedcollaboration}, an active international creativity competition framework with over 50 years of history founded by psychologist Ellis Paul Torrance \citep{torrance1966torrance}. Each task on CGPST is grounded in a future scenario and needs to complete six interdependent steps, each associated with multiple scoring dimensions, as shown in Table~\ref{Table:Step-wise Information on CGPST}. The dataset contains 10 different scenarios, each with 20 samples, resulting in a total of 200 samples. Each sample includes complete responses of six steps and is annotated with calibrated scores from two human evaluators. The inter-rater reliability between the two human evaluators reaches 0.84, indicating that the CGPST dataset is of high quality. Due to its strong subjectivity, cross-step dependencies, fine-grained multi-dimensional scoring, and wide score ranges, CGPST poses significant challenges for automated evaluation. More details are shown in Appendix~\ref{App:CGPST Benchmark Details}.

To further demonstrate the generalizability of CreaEval beyond complex multi-step creativity task, we also evaluate our method on two classic creativity benchmarks with different task structures: 
(1) \textbf{Alternative Uses Task (AUT)} \citep{AUT_1, AUT_2}, a widely used creativity test requiring participants to generate novel uses for a given object, scored exclusively on \textit{Originality}. 
(2) \textbf{Torrance Test of Creative Writing (TTCW)} \citep{10.1145/3613904.3642731}, a narrative generation task requiring creative story writing based on a given plot, evaluated across four dimensions (\textit{Fluency}, \textit{Flexibility}, \textit{Originality}, \textit{Elaboration}). While we have incorporated AUT and TTCW dataset to ensure generalizability, CGPST remains our primary benchmark due to its comprehensive assessment and structural complexity.

\begin{table*}[t]
\centering

\begin{subtable}{\textwidth}
\centering
\resizebox{\textwidth}{!}{
\begin{tabular}{l|
cccc
ccc
cccc}
\toprule

\multirow{2}{*}{\textbf{Method}}
& \multicolumn{4}{c}{Step-1}
& \multicolumn{3}{c}{Step-2}
& \multicolumn{4}{c}{Step-3} \\

\cmidrule(lr){2-5}
\cmidrule(lr){6-8}
\cmidrule(lr){9-12}

& Fluency & Flexibility & Elaboration & Originality
& Integrity & Focus & Adequacy
& Fluency & Flexibility & Elaboration & Originality \\

\midrule

Direct Score
& \underline{0.5577} & 0.1771 & 0.1074 & 0.3552
& 0.184 & 0.042 & 0.0339
& 0.7138 & 0.2965 & 0.3177 & 0.3161 \\

CoT
& 0.4906 & 0.1779 & 0.1298 & 0.2974
& 0.1729 & 0.027 & 0.038
& 0.7182 & 0.2747 & 0.324 & 0.3193 \\

ToT
& 0.5311 & 0.1745 & 0.1269 & 0.3149
& 0.1554 & 0.0573 & 0.0349
& 0.7147 & 0.3013 & 0.315 & 0.2983 \\

GoT
& 0.5244 & 0.1775 & 0.1281 & 0.3364
& 0.1656 & 0.022 & 0.0013
& \underline{0.7464} & 0.3189 & 0.2909 & 0.3251 \\

TaT
& 0.4766 & 0.1853 & 0.1336 & 0.2894
& 0.1954 & 0.0384 & 0.0171
& 0.6365 & 0.3026 & 0.3307 & 0.264 \\

SaMer
& 0.1589 & 0.2668 & 0.2821 & 0.276
& 0.204 & 0.1342 & 0.107
& 0.2117 & 0.1698 & 0.2112 & 0.2461 \\

SFT
& 0.5027 & \underline{0.4324} & \underline{0.4765} & \underline{0.5353}
& \underline{0.476} & \underline{0.4869} & \underline{0.4194}
& 0.4723 & \underline{0.4662} & \underline{0.4756} & \underline{0.4541} \\

CreaEval
& \textbf{0.7505$^{*}$} & \textbf{0.6665$^{*}$} & \textbf{0.656$^{*}$} & \textbf{0.665$^{*}$}
& \textbf{0.7756$^{*}$} & \textbf{0.554$^{*}$} & \textbf{0.4831}
& \textbf{0.8315$^{*}$} & \textbf{0.6975$^{*}$} & \textbf{0.6969$^{*}$} & \textbf{0.6965$^{*}$} \\

\bottomrule
\end{tabular}
}
\end{subtable}

\vspace{1mm}

\begin{subtable}{\textwidth}
\centering
\resizebox{\textwidth}{!}{
\begin{tabular}{l|
cc
c
cccccc
c}
\toprule

\multirow{2}{*}{\textbf{Method}}
& \multicolumn{2}{c}{Step-4}
& \multicolumn{1}{c}{Step-5}
& \multicolumn{6}{c}{Step-6}
& \multirow{2}{*}{AVG} \\

\cmidrule(lr){2-3}
\cmidrule(lr){4-4}
\cmidrule(lr){5-10}

& Correctly Written & Relevance
& Correctly Used
& Relevance & Effectiveness & Criteria & Impact & Humaneness & Development
& \\

\midrule

Direct Score
& \underline{0.4832} & 0.0798
& 0.4627
& 0.0744 & 0.1217 & 0.1491 & 0.1104 & 0.0756 & 0.1714
& 0.2415 \\

CoT
& 0.3723 & 0.1018
& 0.2836
& 0.0328 & 0.1094 & 0.1641 & 0.1038 & 0.0548 & 0.1825
& 0.2187 \\

ToT
& 0.379 & 0.1
& 0.3128
& 0.0415 & 0.1031 & 0.1634 & 0.0975 & 0.0542 & 0.1816
& 0.2229 \\

GoT
& 0.4551 & 0.0438
& 0.0828
& 0.0201 & 0.1077 & 0.183 & 0.1137 & 0.0559 & 0.1821
& 0.214 \\

TaT
& 0.3191 & 0.0679
& 0.3713
& 0.0324 & 0.1108 & 0.1279 & 0.0735 & 0.0419 & 0.15
& 0.2082 \\

SaMer
& 0.1661 & 0.1238
& 0.3594
& 0.1829 & 0.2044 & 0.1749 & 0.1583 & 0.1326 & 0.1052
& 0.1938 \\

SFT & 0.4599 & \underline{0.4572} & \underline{0.5317} & \underline{0.4408}
& \underline{0.4469} & \underline{0.4582} & \underline{0.4694}
& \underline{0.4065} & \underline{0.4617} & \underline{0.4665} \\

CreaEval
& \textbf{0.7933$^{*}$} & \textbf{0.535$^{*}$}
& \textbf{0.9439$^{*}$}
& \textbf{0.5366} & \textbf{0.5214$^{*}$} & \textbf{0.5276$^{*}$} & \textbf{0.4952} & \textbf{0.4388} & \textbf{0.5119}
& \textbf{0.6388$^{*}$} \\

\bottomrule
\end{tabular}
}
\end{subtable}

\caption{Consistency (QWK) results across different methods and dimensions on the CGPST benchmark. \textbf{Bold} indicates the highest score and \underline{underline} indicates the second highest score. The AVG column summarizes the overall average score. The asterisk (*) marks statistically significant improvements (p < 0.05, t-test) over the second-best method.}
\label{Table:QWK_results}
\end{table*}

\subsection{Baselines}

We compare our CreaEval with the following baselines. Detailed implementation details of all baselines are provided in Appendix~\ref{App:Baseline Implementation Details}.

\begin{itemize}
\item \textbf{Direct Score}: This method directly assigns a score to a complete response without intermediate steps or decomposition.
\item \textbf{Chain-of-Thought} (CoT) \citep{NEURIPS2022_9d560961}: This method decomposes the problem into intermediate steps and solve each before giving the final answer.
\item \textbf{Tree-of-Thought} (ToT) \citep{NEURIPS2023_271db992}: This method actively maintains a tree of thoughts, where each thought is a coherent language sequence that serves as an intermediate step toward problem solving.
\item \textbf{Graph-of-Thought} (GoT) \citep{Besta_2024}: This method models the problem-solving process as a graph with more flexible thought transformations compared to ToT.
\item \textbf{Table as Thought} (TaT) \citep{sun-etal-2025-tables}: This method uses a table to represent structured thoughts. Although this method shares the same structured representation, it does not decouple analysis and judging, making it an \textbf{undecoupled variant of CreaEval}.
\item \textbf{Supervised Fine-Tuning} (SFT): We fine-tunes \texttt{Qwen3.5-9B}\footnote{\url{https://huggingface.co/Qwen/Qwen3.5-9B}} on a 7:3 train-test split of the dataset, and reports results on test set.


\item \textbf{SaMer} \citep{ICLR2025_646ca7b9}: SaMer is a scenario-aware multi-dimensional LLM evaluator that adaptively identifies and weights dimensions according to different scenarios.

\end{itemize}

Unlike CGPST, both AUT and TTCW are single-turn generation tasks without procedural process and dependencies. Consequently, we omit structure-based baselines like ToT and GoT for these benchmarks. Additionally, we additionally incorporate a reference-based approach \citep{li-etal-2025-automated-creativity} using human-written stories as evaluation references as a train-free baseline for TTCW.

\subsection{Experimental Setup}
We employ four LLMs (\texttt{qwen3.6-plus}, \texttt{gpt-5.4}, \texttt{deepseek-v4-pro}, \texttt{gemini-3.1-pro}) as Judge-LLMs for all methods. We use \texttt{qwen3.6-plus} as SoT-LLM for structured evidence extraction. Full LLM details are provided in Appendix~\ref{App:LLM Details}. We report three agreement metrics: Pearson Correlation Coefficient (PCC), Quadratic Weighted Kappa (QWK), and Intra-class Correlation Coefficient (ICC), with details in Appendix~\ref{App:Agreement Metric Details}. Additionally, we set the temperature of all LLMs to 0.2. Appendix~\ref{App:Temperature Settings} shows a pilot study on temperature selection. Appendix~\ref{App:Complete Prompts} presents the full prompts.

\begin{table*}[t]
\centering
\small
\begin{tabular}{l c cccc c}
\toprule
\multirow{2}{*}{\textbf{Method}} & AUT & \multicolumn{5}{c}{TTCW} \\
\cmidrule(lr){2-2} \cmidrule(lr){3-7}
 & Originality & Fluency & Flexibility & Originality & Elaboration & AVG \\
\midrule
Direct Score & 0.2731 & 0.5367 & 0.4234 & 0.2508 & 0.2145 & 0.3558 \\
CoT & 0.219 & 0.3392 & 0.1879 & 0.2227 & 0.2744 & 0.256 \\
SFT & \underline{0.6924} & 0.3888 & 0.2783 & 0.2252 & 0.2095 & 0.2755 \\
Reference-based & -- & \underline{0.6484} & \underline{0.5775} & \underline{0.5802} & \underline{0.5174} & \underline{0.581} \\
\midrule
\textbf{CreaEval} & \textbf{0.7382} & \textbf{0.7458$^{*}$} & \textbf{0.8052$^{*}$} & \textbf{0.6249} & \textbf{0.652$^{*}$} & \textbf{0.707$^{*}$} \\
\bottomrule
\end{tabular}
\caption{Consistency (QWK) results across different methods and dimensions on the AUT and TTCW benchmark. \textbf{Bold} indicates the highest score and \underline{underline} indicates the second highest score. The AVG column summarizes the overall average score. The asterisk (*) marks statistically significant improvements (p < 0.05, t-test) over the second-best method.}
\label{tab:aut_ttcw_results}
\end{table*}

\section{Results}

\subsection{Main Results}

As shown in Table~\ref{Table:QWK_results} (reporting QWK. PCC and ICC show similar trends, see Appendix \ref{App:Complete Results} for full results), CreaEval consistently achieves the best performance (0.64), significantly outperforming all baselines on the CGPST. This indicates that CreaEval better aligns with human judgments with decoupled analysis and judging. In contrast, other training-free methods show relatively weak and similar performance (0.2-0.25), suggesting that simply increasing reasoning steps does not improve accuracy for subjective dimensions evaluation. For training-based methods, SFT achieves the second-best performance (0.47), indicating that supervised data can effectively align LLM judging with human preferences to a certain extent, but still falls short of CreaEval. Furthermore, as presented in Table~\ref{tab:aut_ttcw_results}, CreaEval also achieves superior performance on simple creativity tasks, demonstrating its strong generalizability to simpler creativity tasks beyond complex multi-step evaluations.

Performance varies across dimensions. As shown in Table~\ref{Table:QWK_results}, for dimensions with large range such as Step-2 (\textit{Focus}, \textit{Adequacy}) and Step-4 (\textit{Relevance}), most training-free baselines remain below 0.2, indicating that they struggle to accurately distinguish score differences within wide ranges. In contrast, Step-5 is relatively less subjective, as it mainly involves verifying whether scoring vectors satisfy a non-repetitive ranking structure. Nevertheless, most training-free methods still achieve only weak performance (around 0.08–0.46 v.s. CreaEval 0.94), as they retain redundant information from previous steps without effective evidence extraction, which further disrupts subsequent evaluations. Step-6 is the most challenging step, requiring holistic integration across all previous steps to produce a final action plan. In Step-6, all training-free baselines collapse to very weak performance (<0.2), whereas CreaEval still maintains moderate consistency (around 0.5). This further validates the effectiveness of the decoupled analysis-judging design.

\begin{figure}[!t]
    \centering
    \includegraphics[width=\columnwidth]{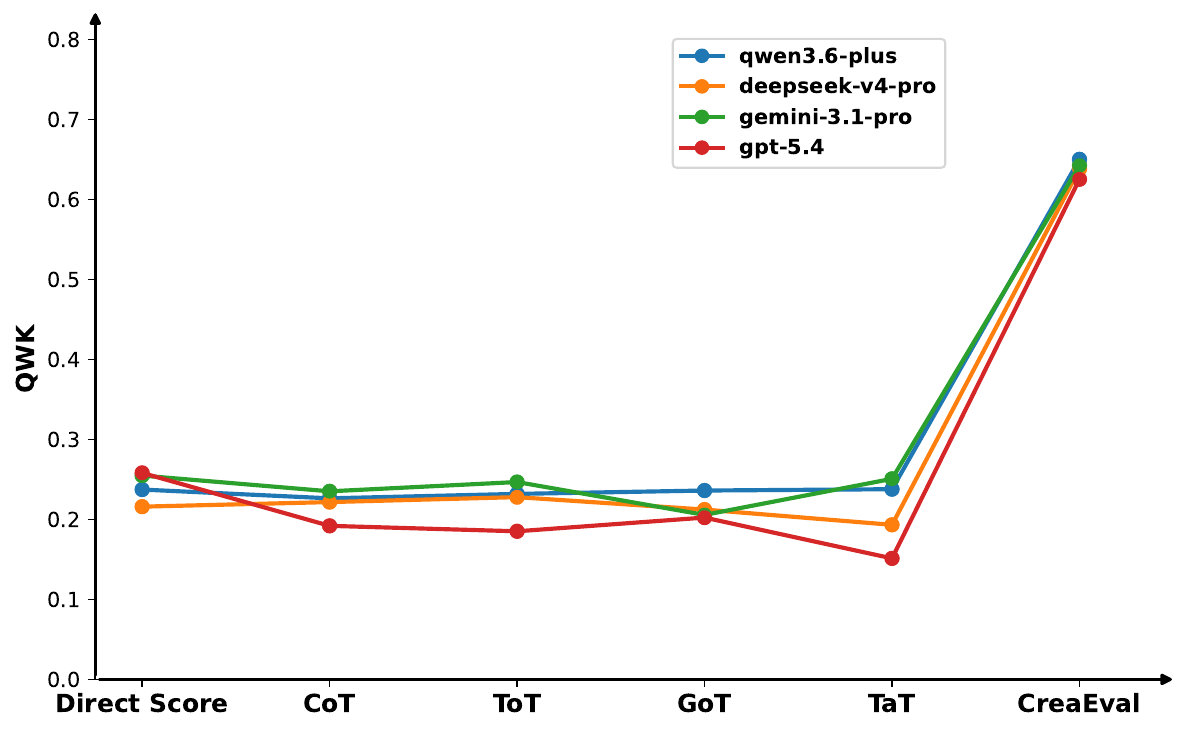} 
    \caption{Judge QWK Consistency Across Methods.}
    \label{Figure:Judge QWK Consistency Across Methods}
\end{figure}

Furthermore, as shown in Figure~\ref{Figure:Judge QWK Consistency Across Methods}, we analyze the consistency of four Judge-LLMs under each method. CreaEval exhibits nearly identical performance across all judges, indicating stable and robust evaluation behavior. This stability stems from replacing raw responses with structured evidence, which reduces the impact of subjective variation in the raw responses in scoring, further demonstrating that CreaEval is a model-agnostic framework that does not rely on the specific Judge-LLM. In contrast, training-free methods show noticeable fluctuations across different judges due to variations in model capability, resulting in more diverse scoring behavior. We further provide an in-depth analysis of scoring stability in Section~\ref{Section:Scoring Stability Analysis}.

\subsection{Ablation Study}
We conduct ablation studies to evaluate the effectiveness of key components in CreaEval. Specifically, we consider four variants: (1) \textbf{w/o Memory}, where the memory mechanism in Phase 1 is removed, and each step is processed independently without cross-step dependency; (2) \textbf{w/o Evidence}, where Judge-LLM directly performs scoring based on the raw responses without extracting structured evidence, corresponding to Direct Score baseline; (3) \textbf{w/o Decoupled}, where analysis and judging are performed jointly within a single LLM without separation, corresponding to TaT baseline; and (4) \textbf{w/o Step-wise}, where evidence extraction is performed in a single pass over all six steps instead of step-by-step iterative extraction.

Table~\ref{Table:Ablations} presents the results. Removing memory leads to performance degradation (QWK = 0.49, 23\%$\downarrow$), demonstrating that the memory mechanism is necessary for maintaining cross-step coherence. Without evidence, the performance decreases (QWK = 0.24, 63\%$\downarrow$), indicating that evidence-based scoring is more accurate than direct scoring. Similarly, when analysis and judging are not decoupled, performance further degrades (QWK = 0.21, 67\%$\downarrow$), suggesting that coupling the two processes introduces analytical bias into scoring. Meanwhile, extracting evidence for all steps at once leads to a collapse in performance (QWK = 0.01, 98\%$\downarrow$), indicating that excessive input information significantly harms SoT-LLM’s evidence extraction accuracy. For example, SoT-LLM provides evidence for some dimensions but only produces binary judgments (e.g., \textit{``Yes''} or \textit{``No''}) for others, which leads to significant degradation in scoring quality.

\begin{table}[!t]
\centering
\begin{tabular}{l|ccc}
\toprule
Method & PCC & ICC & QWK \\
\midrule
\textbf{CreaEval} & \textbf{0.69} & \textbf{0.64} & \textbf{0.64} \\
w/o Memory & 0.57 & 0.49 & 0.49 \\
w/o Evidence & 0.33 & 0.24 & 0.24 \\
w/o Decoupled & 0.29 & 0.2 & 0.21 \\
w/o Step-wise & -0.01 & 0.01 & 0.01 \\
\bottomrule
\end{tabular}
\caption{Ablation results.}
\label{Table:Ablations}
\end{table}

\begin{figure*}
    \centering
    \includegraphics[width=\textwidth]{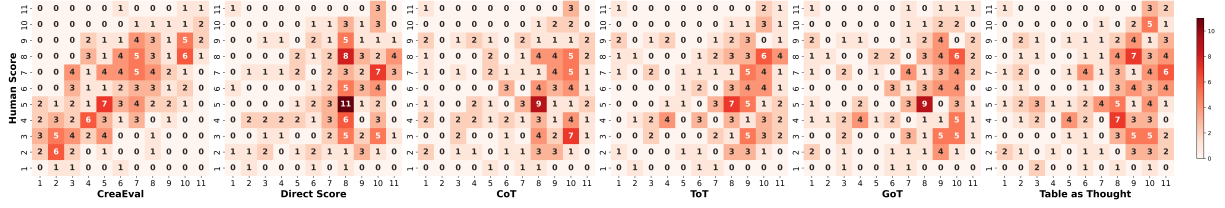} 
    \caption{Headmap between human score and all methods on \textit{Originality} of Step-3.}
    \label{Figure:Headmap on Originality of Step-3}
\end{figure*} 

\section{Discussion}
In this section, we conduct a comprehensive analysis of CreaEval, including robustness, bias mitigation, scoring stability, efficiency, and case study, further validating the effectiveness of CreaEval beyond simple improvements in scoring accuracy.

\subsection{Impact of SoT-LLM in CreaEval}
\label{Section:Impact of SoT-LLM in CreaEval}

\begin{figure}[!t]
    \centering
    \includegraphics[width=\columnwidth]{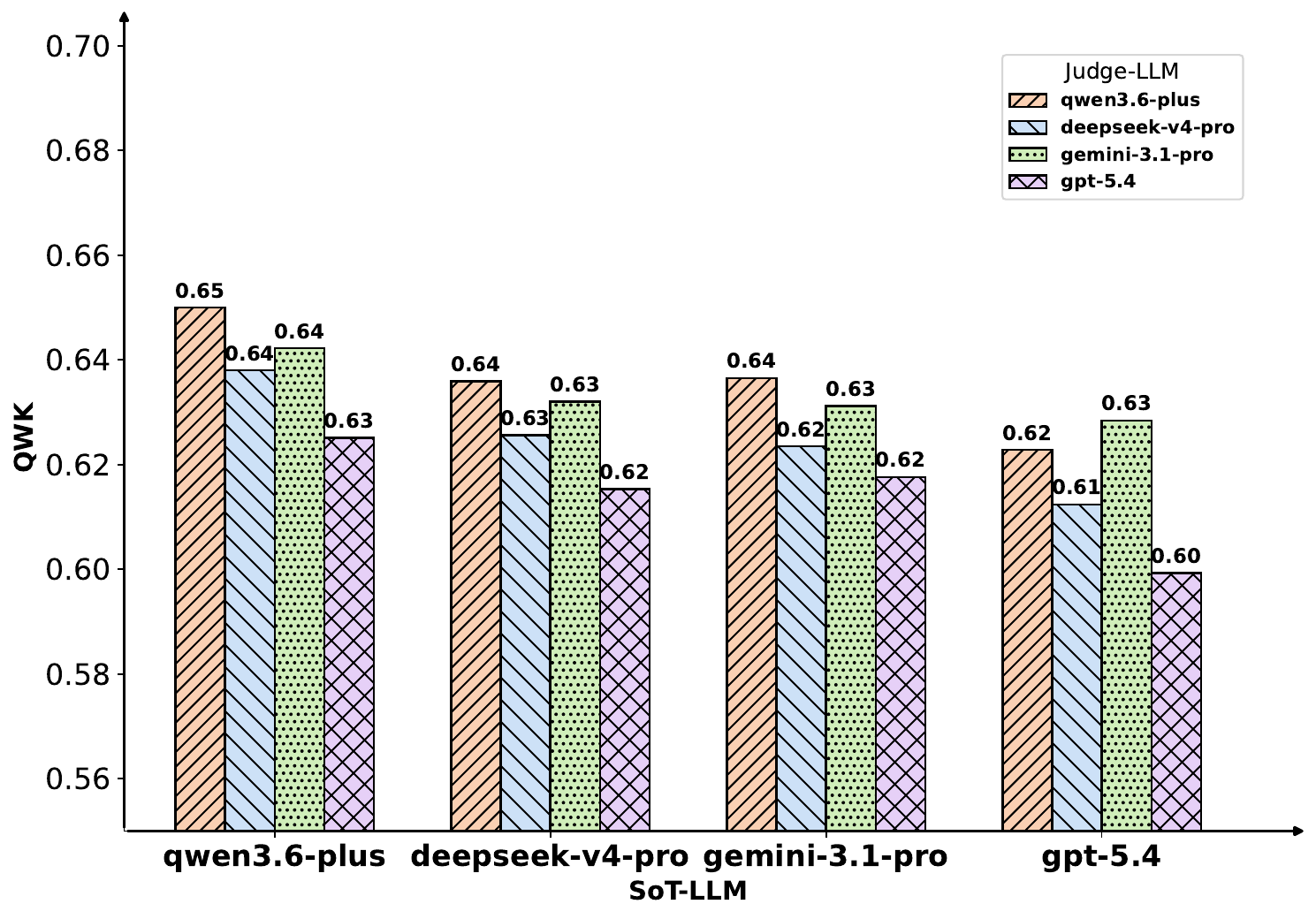} 
    \caption{QWK results across different SoT-LLMs.}
    \label{Figure:QWK across Different SoT-LLMs}
\end{figure}

The quality of extracted evidence directly affects the accuracy of the final scoring, making SoT-LLM a critical component. To evaluate the robustness of CreaEval, we replace SoT-LLM with other LLMs and report QWK across different LLMs combinations. As shown in Figure~\ref{Figure:QWK across Different SoT-LLMs}, the agreement remains consistently above 0.6 across all SoT-LLMs and Judge-LLMs combinations, indicating that CreaEval does not rely on a specific LLM and can serve as a general and robust evaluation framework. PCC and ICC results are provided in Appendix~\ref{App:PCC and ICC Results for CreaEval Robustness Analysis}.

\subsection{Bias Mitigation in CreaEval}
\label{Section:Bias Mitigation in CreaEval}

\begin{figure}
    \centering
    \includegraphics[width=\columnwidth]{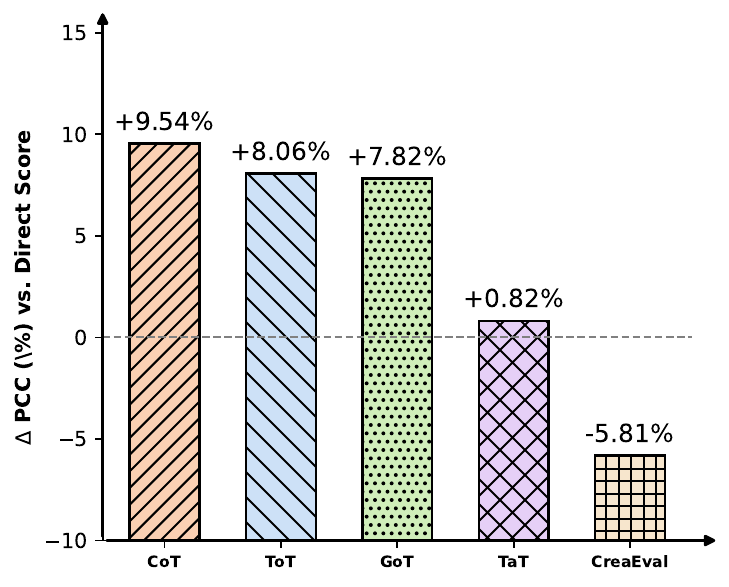} 
    \caption{Relative changes in PCC against Direct Score. PCC measures the linear correlation between Step-6 response length and \textit{Development} scores.}
    \label{Figure:Relative changes in PCC against Direct Score}
\end{figure} 

In LLM-as-a-Judge, LLMs are prone to various evaluation biases during scoring. In automated evaluation of creative tasks, two biases are particularly critical: \textbf{(1) leniency bias}, where LLMs tend to assign overly high scores to subjective dimensions due to their inherent sycophantic tendencies \citep{ye2025justice, gupta2026contextcontentexposingevaluation}; and \textbf{(2) verbosity bias}, where LLM favor longer responses over shorter ones, even when the latter are clearer or of higher quality \citep{NEURIPS2023_91f18a12}.

For leniency bias, we analyze the heatmap between different methods and human scores on \textit{Originality} of Step-3, as shown in Figure~\ref{Figure:Headmap on Originality of Step-3}. The results indicate that all training-free methods exhibit a clear inclination to cluster their scores within the high-value region. In contrast, CreaEval significantly mitigates this tendency, demonstrating a more balanced score distribution that aligns closely with human annotations. Heatmaps for additional dimensions are provided in Appendix~\ref{App:Heatmap Distributions across Dimensions}.

For verbosity bias, we examine PCC between response length of Step-6 and \textit{Development} (the level of detail of action plan) scores to measure their linear correlation. Figure~\ref{Figure:Relative changes in PCC against Direct Score} illustrates the relative changes for training-free methods against Direct Score. Only CreaEval shows a decrease (-5.81\%), indicating that response length is less correlated with scores compared to Direct Score and thus verbosity bias is mitigated. However, CoT, GoT, and ToT even exacerbate the bias despite introducing more complex reasoning paths. TaT shows a slight increase (+0.82\%), suggesting that incorporating analysis can partially reduce verbosity bias. However, due to its coupled analysis-and-judging design, its effectiveness remains inferior to CreaEval.


\begin{figure}[!t]
    \centering
    \includegraphics[width=\columnwidth]{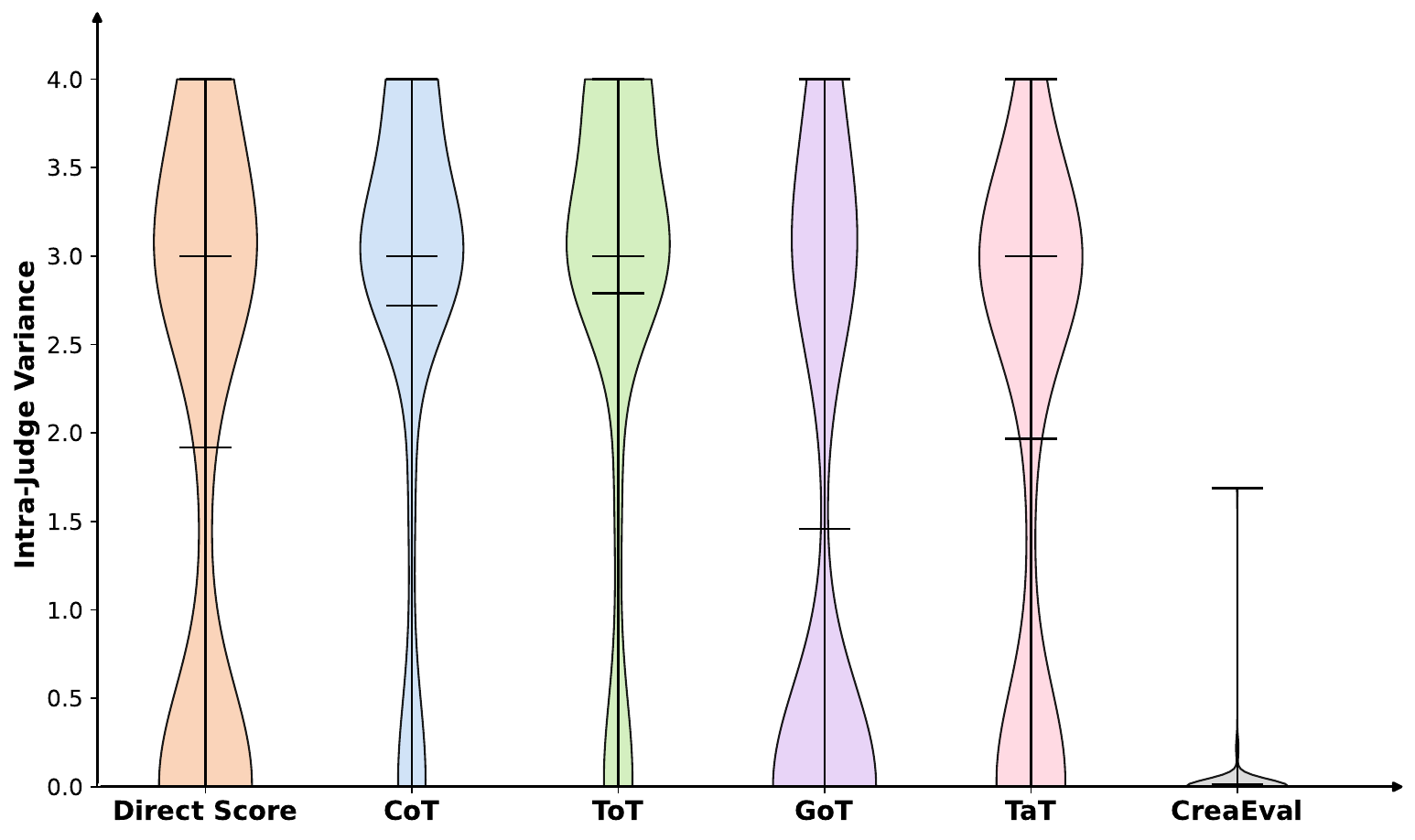} 
    \caption{Distribution of inter-Judge variance across the entire dataset for \textit{Correctly Used} in Step-5.}
    \label{Figure:Distribution of inter-Judge variance for Correctly Used in Step-5}
\end{figure} 

\begin{figure*}
    \centering
    \includegraphics[width=0.95\textwidth]{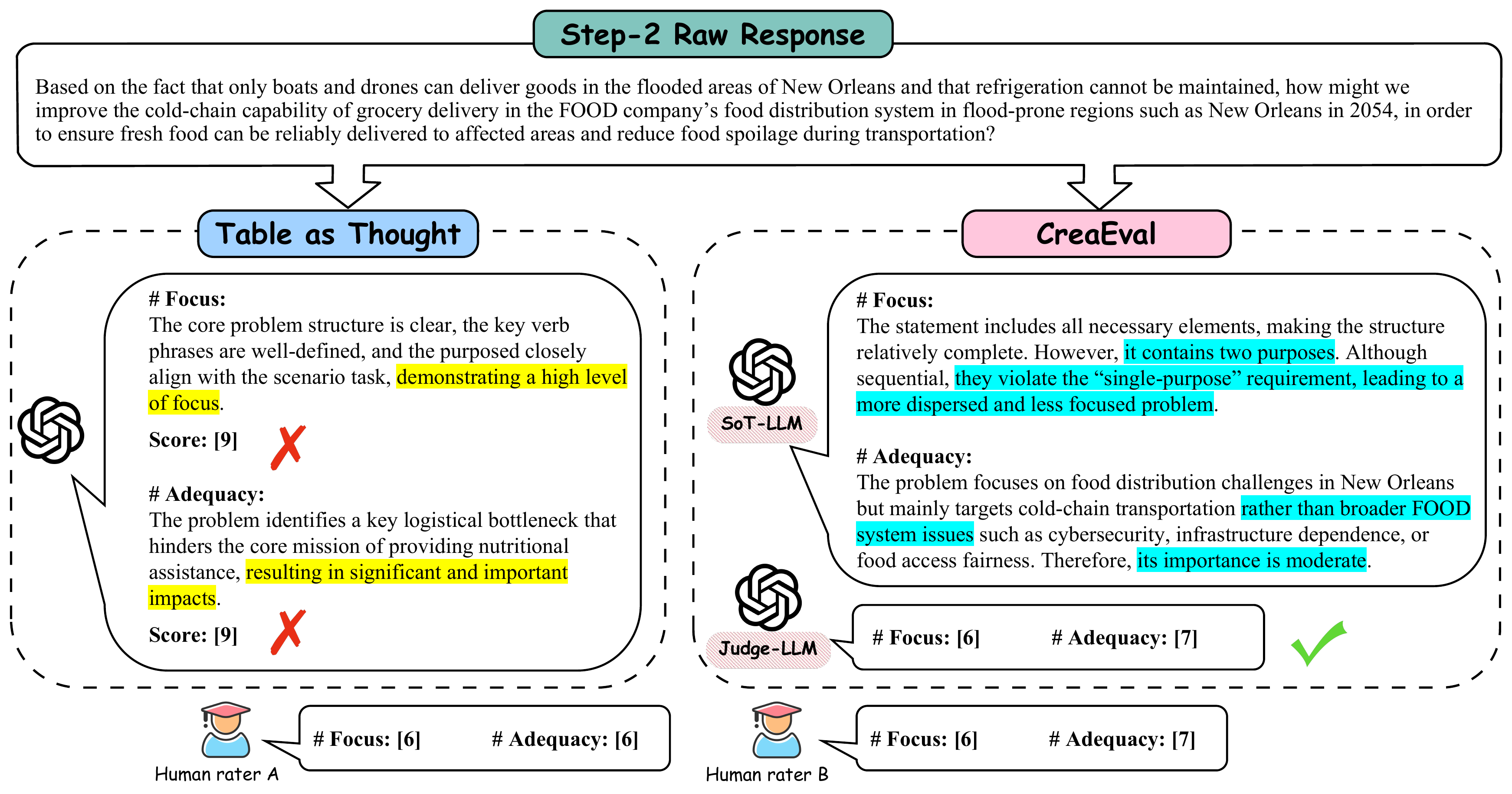} 
    \caption{Case study of Step-2 comparing Table as Thought (TaT) and CreaEval. The example illustrates that TaT tends to produce overly positive and less discriminative judgments (yellow) due to coupled analysis and scoring within a single LLM, leading to leniency bias. In contrast, CreaEval separates evidence extraction and judging into distinct LLMs, enabling more grounded and constraint-aware judgments (blue).}
    \label{Figure:Case Study of Step-2}
\end{figure*} 

\subsection{Scoring Stability Analysis}
\label{Section:Scoring Stability Analysis}
We analyze scoring stability by measuring the distribution of inter-Judge variance for each method. Specifically, for each sample and dimension, we compute the variance of scores assigned by four Judge-LLMs. This yields 200 variance values per method, corresponding to all samples on CGPST dataset, which are then analyzed as a distribution. Figure~\ref{Figure:Distribution of inter-Judge variance for Correctly Used in Step-5} illustrates the variance distribution for \textit{Correctly Used} in Step-5. CreaEval exhibits consistently lower inter-Judge variance, indicating the four Judge-LLMs produce highly similar scores, leading to higher scoring stability. This is attributed to its evidence-based judging design, where all Judges rely on shared extracted evidence rather than raw subjective responses, leading to more consistent evaluations. In contrast, other methods show substantially higher and more dispersed variance, suggesting unstable judgments across different Judges even for the same response. Violin plots for other dimensions are provided in Appendix~\ref{App:Violin Plots of Inter-Judge Variance across Dimensions}.

\subsection{Inference Cost Analysis}

\begin{table}[t]
\centering
\begin{tabular}{l|cc|cc}
\toprule
& \multicolumn{2}{c|}{Time (s)} 
& \multicolumn{2}{c}{Token ($\times 10^3$)} \\
\cmidrule(lr){2-3} \cmidrule(lr){4-5}
Method & p50 & p95 & p50 & p95 \\
\midrule
Direct Score        & 30  & 37  & 13.20  & 15.70 \\
CoT                  & 54  & 158 & 66.09  & 73.24 \\
ToT                  & 200 & 332 & 269.65 & 297.22 \\
GoT                  & 479 & 714 & 567.37 & 620.98 \\
TaT   & 72  & 104 & 16.13  & 18.80 \\
CreaEval        & 173 & 249 & 42.85  & 47.13 \\
\bottomrule
\end{tabular}

\caption{Time and token consumption. Both are reported in terms of the 50th and 95th percentile (p50/p95).}
\label{Table:Time and token consumption}
\end{table}

We further analyze the inference cost of different methods in terms of time and token consumption, as shown in Table~\ref{Table:Time and token consumption}. Overall, methods with more complex reasoning exhibit significantly higher consumption. In particular, GoT and ToT introduce substantial overhead due to expanded reasoning paths. In contrast, CoT and TaT achieves relatively low consumption. CreaEval maintains a better balance, achieving substantially lower cost than ToT and GoT while maintaining strong performance.

\subsection{Case Study}
\label{Section:Case Study}
As shown in Figure~\ref{Figure:Case Study of Step-2}, we present a case study of Step-2. TaT, an undecoupled variant of CreaEval, tends to produce overly positive evaluations (e.g., ``\textit{a high level of focus}''), exhibiting clear leniency bias. In contrast, CreaEval decouples evidence extraction from judging, grounding scoring in structured intermediate evidence rather than raw responses. This design mitigates over-optimistic scoring (e.g., ``\textit{less focused problem}''), leading to more accurate evaluations. Case studies for other steps are provided in Appendix~\ref{App:Case Studies for Remaining Steps}.

\section{Conclusions}
In this work, we propose CreaEval, an automated creativity evaluation framework for CGPST that decouples evaluation into memory-augmented analysis and evidence-based judging. SoT-LLM first converts multi-step responses into structured evidence with cross-step memory, and Judge-LLM performs scoring based solely on this evidence without accessing the raw responses. Experiments across four LLMs show that CreaEval achieves strong alignment with human judgments, consistently outperforming all baselines. Further analysis demonstrates that the decoupled design improves scoring stability and reduces verbosity and leniency biases in LLM-as-a-Judge, offering new insights into evaluating subjective creativity tasks.

\section*{Limitations}
Although our work demonstrates strong effectiveness and achieves promising results, it still has several limitations. In Phase 1, we use a simple rule-based memory mechanism to maintain cross-step coherence due to the fixed step dependencies on CGPST. More advanced memory modules such as hierarchical memory modules could be explored for more general settings. In addition, to the best of our knowledge, CGPST is currently the only publicly available multi-step creativity benchmark dataset, so our experiments are conducted exclusively on this multi-step dataset. Future work will evaluate the effectiveness of CreaEval on additional datasets once they become available.

\bibliography{references}

@inproceedings{NEURIPS2023_91f18a12,
 author = {Zheng, Lianmin and Chiang, Wei-Lin and Sheng, Ying and Zhuang, Siyuan and Wu, Zhanghao and Zhuang, Yonghao and Lin, Zi and Li, Zhuohan and Li, Dacheng and Xing, Eric and Zhang, Hao and Gonzalez, Joseph and Stoica, Ion},
 booktitle = {Advances in Neural Information Processing Systems},
 editor = {A. Oh and T. Naumann and A. Globerson and K. Saenko and M. Hardt and S. Levine},
 pages = {46595--46623},
 publisher = {Curran Associates, Inc.},
 title = {Judging LLM-as-a-Judge with MT-Bench and Chatbot Arena},
 url = {https://proceedings.neurips.cc/paper_files/paper/2023/file/91f18a1287b398d378ef22505bf41832-Paper-Datasets_and_Benchmarks.pdf},
 volume = {36},
 year = {2023}
}

@inproceedings{li-etal-2025-automated-creativity,
    title = "Automated Creativity Evaluation for Large Language Models: A Reference-Based Approach",
    author = "Li, Ruizhe  and
      Zhu, Chiwei  and
      Xu, Benfeng  and
      Wang, Xiaorui  and
      Mao, Zhendong",
    editor = "Christodoulopoulos, Christos  and
      Chakraborty, Tanmoy  and
      Rose, Carolyn  and
      Peng, Violet",
    booktitle = "Findings of the Association for Computational Linguistics: EMNLP 2025",
    month = nov,
    year = "2025",
    address = "Suzhou, China",
    publisher = "Association for Computational Linguistics",
    url = "https://aclanthology.org/2025.findings-emnlp.1171/",
    doi = "10.18653/v1/2025.findings-emnlp.1171",
    pages = "21475--21488",
    ISBN = "979-8-89176-335-7"
}

@inproceedings{liang-etal-2024-encouraging,
    title = "Encouraging Divergent Thinking in Large Language Models through Multi-Agent Debate",
    author = "Liang, Tian  and
      He, Zhiwei  and
      Jiao, Wenxiang  and
      Wang, Xing  and
      Wang, Yan  and
      Wang, Rui  and
      Yang, Yujiu  and
      Shi, Shuming  and
      Tu, Zhaopeng",
    editor = "Al-Onaizan, Yaser  and
      Bansal, Mohit  and
      Chen, Yun-Nung",
    booktitle = "Proceedings of the 2024 Conference on Empirical Methods in Natural Language Processing",
    month = nov,
    year = "2024",
    address = "Miami, Florida, USA",
    publisher = "Association for Computational Linguistics",
    url = "https://aclanthology.org/2024.emnlp-main.992/",
    doi = "10.18653/v1/2024.emnlp-main.992",
    pages = "17889--17904"
}

@article{torrance1966torrance,
  title={Torrance tests of creative thinking},
  author={Torrance, E Paul},
  journal={Educational and psychological measurement},
  year={1966}
}

@inproceedings{10.1145/3706598.3714198,
author = {Kumar, Harsh and Vincentius, Jonathan and Jordan, Ewan and Anderson, Ashton},
title = {Human Creativity in the Age of LLMs: Randomized Experiments on Divergent and Convergent Thinking},
year = {2025},
isbn = {9798400713941},
publisher = {Association for Computing Machinery},
address = {New York, NY, USA},
url = {https://doi.org/10.1145/3706598.3714198},
doi = {10.1145/3706598.3714198},
booktitle = {Proceedings of the 2025 CHI Conference on Human Factors in Computing Systems},
articleno = {23},
numpages = {18},
location = {
},
series = {CHI '25}
}

@article{Zhao2025,
  title = {Assessing and Understanding Creativity in Large Language Models},
  volume = {22},
  ISSN = {2731-5398},
  url = {http://dx.doi.org/10.1007/s11633-025-1546-4},
  DOI = {10.1007/s11633-025-1546-4},
  number = {3},
  journal = {Machine Intelligence Research},
  publisher = {Springer Science and Business Media LLC},
  author = {Zhao,  Yunpu and Zhang,  Rui and Li,  Wenyi and Li,  Ling},
  year = {2025},
  month = apr,
  pages = {417–436}
}

@misc{lu2024llmdiscussionenhancingcreativity,
      title={LLM Discussion: Enhancing the Creativity of Large Language Models via Discussion Framework and Role-Play}, 
      author={Li-Chun Lu and Shou-Jen Chen and Tsung-Min Pai and Chan-Hung Yu and Hung-yi Lee and Shao-Hua Sun},
      year={2024},
      eprint={2405.06373},
      archivePrefix={arXiv},
      primaryClass={cs.CL},
      url={https://arxiv.org/abs/2405.06373}, 
}

@inproceedings{10.1145/3613904.3642731,
author = {Chakrabarty, Tuhin and Laban, Philippe and Agarwal, Divyansh and Muresan, Smaranda and Wu, Chien-Sheng},
title = {Art or Artifice? Large Language Models and the False Promise of Creativity},
year = {2024},
isbn = {9798400703300},
publisher = {Association for Computing Machinery},
address = {New York, NY, USA},
url = {https://doi.org/10.1145/3613904.3642731},
doi = {10.1145/3613904.3642731},
booktitle = {Proceedings of the 2024 CHI Conference on Human Factors in Computing Systems},
articleno = {30},
numpages = {34},
location = {Honolulu, HI, USA},
series = {CHI '24}
}

@inproceedings{lin-etal-2025-creativity,
    title = "Creativity in {LLM}-based Multi-Agent Systems: A Survey",
    author = "Lin, Yi-Cheng  and
      Chen, Kang-Chieh  and
      Li, Zhe-Yan  and
      Wu, Tzu-Heng  and
      Wu, Tzu-Hsuan  and
      Chen, Kuan-Yu  and
      Lee, Hung-yi  and
      Chen, Yun-Nung",
    editor = "Christodoulopoulos, Christos  and
      Chakraborty, Tanmoy  and
      Rose, Carolyn  and
      Peng, Violet",
    booktitle = "Proceedings of the 2025 Conference on Empirical Methods in Natural Language Processing",
    month = nov,
    year = "2025",
    address = "Suzhou, China",
    publisher = "Association for Computational Linguistics",
    url = "https://aclanthology.org/2025.emnlp-main.1403/",
    doi = "10.18653/v1/2025.emnlp-main.1403",
    pages = "27584--27607",
    ISBN = "979-8-89176-332-6"
}

@article{Ye_Gu_Zhao_Yin_Wang_2025, title={Assessing the Creativity of LLMs in Proposing Novel Solutions to Mathematical Problems}, volume={39}, url={https://ojs.aaai.org/index.php/AAAI/article/view/34760}, DOI={10.1609/aaai.v39i24.34760}, abstractNote={The mathematical capabilities of AI systems are complex and multifaceted. Most existing research has predominantly focused on the correctness of AI-generated solutions to mathematical problems. In this work, we argue that beyond producing correct answers, AI systems should also be capable of, or assist humans in, developing novel solutions to mathematical challenges. This study explores the creative potential of Large Language Models (LLMs) in mathematical reasoning, an aspect that has received limited attention in prior research. We introduce a novel framework and benchmark, CreativeMath, which encompasses problems ranging from middle school curricula to Olympic-level competitions, designed to assess LLMs’ ability to propose innovative solutions after some known solutions have been provided. Our experiments demonstrate that, while LLMs perform well on standard mathematical tasks, their capacity for creative problem-solving varies considerably. Notably, the Gemini-1.5-Pro model outperformed other LLMs in generating novel solutions. This research opens a new frontier in evaluating AI creativity, shedding light on both the strengths and limitations of LLMs in fostering mathematical innovation, and setting the stage for future developments in AI-assisted mathematical discovery.}, number={24}, journal={Proceedings of the AAAI Conference on Artificial Intelligence}, author={Ye, Junyi and Gu, Jingyi and Zhao, Xinyun and Yin, Wenpeng and Wang, Guiling}, year={2025}, month={Apr.}, pages={25687-25696} }

@misc{wang2026teamllmhumanliketeamorientedcollaboration,
      title={TeamLLM: A Human-Like Team-Oriented Collaboration Framework for Multi-Step Contextualized Tasks}, 
      author={Xiangyu Wang and Jin Wu and Haoran Shi and Wei Xia and Jiarui Yu and Chanjin Zheng},
      year={2026},
      eprint={2604.06765},
      archivePrefix={arXiv},
      primaryClass={cs.CL},
      url={https://arxiv.org/abs/2604.06765}, 
}

@inproceedings{NEURIPS2022_9d560961,
 author = {Wei, Jason and Wang, Xuezhi and Schuurmans, Dale and Bosma, Maarten and ichter, brian and Xia, Fei and Chi, Ed and Le, Quoc V and Zhou, Denny},
 booktitle = {Advances in Neural Information Processing Systems},
 editor = {S. Koyejo and S. Mohamed and A. Agarwal and D. Belgrave and K. Cho and A. Oh},
 pages = {24824--24837},
 publisher = {Curran Associates, Inc.},
 title = {Chain-of-Thought Prompting Elicits Reasoning in Large Language Models},
 url = {https://proceedings.neurips.cc/paper_files/paper/2022/file/9d5609613524ecf4f15af0f7b31abca4-Paper-Conference.pdf},
 volume = {35},
 year = {2022}
}

@inproceedings{NEURIPS2023_271db992,
 author = {Yao, Shunyu and Yu, Dian and Zhao, Jeffrey and Shafran, Izhak and Griffiths, Tom and Cao, Yuan and Narasimhan, Karthik},
 booktitle = {Advances in Neural Information Processing Systems},
 editor = {A. Oh and T. Naumann and A. Globerson and K. Saenko and M. Hardt and S. Levine},
 pages = {11809--11822},
 publisher = {Curran Associates, Inc.},
 title = {Tree of Thoughts: Deliberate Problem Solving with Large Language Models},
 url = {https://proceedings.neurips.cc/paper_files/paper/2023/file/271db9922b8d1f4dd7aaef84ed5ac703-Paper-Conference.pdf},
 volume = {36},
 year = {2023}
}

@article{Besta_2024,
   title={Graph of Thoughts: Solving Elaborate Problems with Large Language Models},
   volume={38},
   ISSN={2159-5399},
   url={http://dx.doi.org/10.1609/aaai.v38i16.29720},
   DOI={10.1609/aaai.v38i16.29720},
   number={16},
   journal={Proceedings of the AAAI Conference on Artificial Intelligence},
   publisher={Association for the Advancement of Artificial Intelligence (AAAI)},
   author={Besta, Maciej and Blach, Nils and Kubicek, Ales and Gerstenberger, Robert and Podstawski, Michal and Gianinazzi, Lukas and Gajda, Joanna and Lehmann, Tomasz and Niewiadomski, Hubert and Nyczyk, Piotr and Hoefler, Torsten},
   year={2024},
   month=Mar, pages={17682–17690} }

@inproceedings{sun-etal-2025-tables,
    title = "Tables as Thought: Exploring Structured Thoughts in {LLM} Reasoning",
    author = "Sun, Zhenjie  and
      Deng, Naihao  and
      Yu, Haofei  and
      You, Jiaxuan",
    editor = "Chang, Shuaichen  and
      Hulsebos, Madelon  and
      Liu, Qian  and
      Chen, Wenhu  and
      Sun, Huan",
    booktitle = "Proceedings of the 4th Table Representation Learning Workshop",
    month = jul,
    year = "2025",
    address = "Vienna, Austria",
    publisher = "Association for Computational Linguistics",
    url = "https://aclanthology.org/2025.trl-1.3/",
    doi = "10.18653/v1/2025.trl-1.3",
    pages = "19--33",
    ISBN = "979-8-89176-268-8"
}

@misc{wang2026t2sbenchstructureofthoughtbenchmarking,
      title={T2S-Bench \& Structure-of-Thought: Benchmarking and Prompting Comprehensive Text-to-Structure Reasoning}, 
      author={Qinsi Wang and Hancheng Ye and Jinhee Kim and Jinghan Ke and Yifei Wang and Martin Kuo and Zishan Shao and Dongting Li and Yueqian Lin and Ting Jiang and Chiyue Wei and Qi Qian and Wei Wen and Helen Li and Yiran Chen},
      year={2026},
      eprint={2603.03790},
      archivePrefix={arXiv},
      primaryClass={cs.CL},
      url={https://arxiv.org/abs/2603.03790}, 
}

@inproceedings{qi-etal-2025-sot,
    title = "{S}o{T}: Structured-of-Thought Prompting Guides Multilingual Reasoning in Large Language Models",
    author = "Qi, Rui  and
      Man, Zhibo  and
      Chen, Yufeng  and
      Mo, Fengran  and
      Xu, Jinan  and
      Huang, Kaiyu",
    editor = "Christodoulopoulos, Christos  and
      Chakraborty, Tanmoy  and
      Rose, Carolyn  and
      Peng, Violet",
    booktitle = "Findings of the Association for Computational Linguistics: EMNLP 2025",
    month = nov,
    year = "2025",
    address = "Suzhou, China",
    publisher = "Association for Computational Linguistics",
    url = "https://aclanthology.org/2025.findings-emnlp.586/",
    doi = "10.18653/v1/2025.findings-emnlp.586",
    pages = "11024--11039",
    ISBN = "979-8-89176-335-7"
}

@article{treffinger2012four,
  title={Four decades of creative vision: Insights from an evaluation of the Future Problem Solving Program International (FPSPI)},
  author={Treffinger, Donald J and Solomon, Marianne and Woythal, Deb},
  journal={The Journal of Creative Behavior},
  volume={46},
  number={3},
  pages={209--219},
  year={2012},
  publisher={Wiley Online Library}
}

@article{Treffinger1995,
  title = {Creative Problem Solving: Overview and educational implications},
  volume = {7},
  ISSN = {1573-336X},
  url = {http://dx.doi.org/10.1007/BF02213375},
  DOI = {10.1007/bf02213375},
  number = {3},
  journal = {Educational Psychology Review},
  publisher = {Springer Science and Business Media LLC},
  author = {Treffinger,  Donald J.},
  year = {1995},
  month = sep,
  pages = {301–312}
}

@inproceedings{do-etal-2024-autoregressive,
    title = "Autoregressive Score Generation for Multi-trait Essay Scoring",
    author = "Do, Heejin  and
      Kim, Yunsu  and
      Lee, Gary",
    editor = "Graham, Yvette  and
      Purver, Matthew",
    booktitle = "Findings of the Association for Computational Linguistics: EACL 2024",
    month = mar,
    year = "2024",
    address = "St. Julian{'}s, Malta",
    publisher = "Association for Computational Linguistics",
    url = "https://aclanthology.org/2024.findings-eacl.115/",
    doi = "10.18653/v1/2024.findings-eacl.115",
    pages = "1659--1666"
}

@inproceedings{wang-liu-2025-mes,
    title = "{T}-{MES}: Trait-Aware Mix-of-Experts Representation Learning for Multi-trait Essay Scoring",
    author = "Wang, Jiong  and
      Liu, Jie",
    editor = "Rambow, Owen  and
      Wanner, Leo  and
      Apidianaki, Marianna  and
      Al-Khalifa, Hend  and
      Eugenio, Barbara Di  and
      Schockaert, Steven",
    booktitle = "Proceedings of the 31st International Conference on Computational Linguistics",
    month = jan,
    year = "2025",
    address = "Abu Dhabi, UAE",
    publisher = "Association for Computational Linguistics",
    url = "https://aclanthology.org/2025.coling-main.81/",
    pages = "1224--1236"
}

@article{Li_Pan_2025, title={KAES: Multi-aspect Shared Knowledge Finding and Aligning for Cross-prompt Automated Scoring of Essay Traits}, volume={39}, url={https://ojs.aaai.org/index.php/AAAI/article/view/34626}, DOI={10.1609/aaai.v39i23.34626}, abstractNote={Cross-prompt automated essay scoring (AES) aims to train models using essays from different source prompts and test them on new target prompt essays. A core challenge of the task is to learn as much shared knowledge as possible between essays from different prompts in order to better represent new prompt essays. Previous studies primarily focus on learning this knowledge on a general, coarse-grained level, ignoring that the shared knowledge among prompts is highly detailed and contains a more comprehensive range of information that is not fully investigated. In this paper, we propose a novel multi-aspect knowledge finding and aligning optimization strategy to better acquire this detailed various shared knowledge. We also introduce LLM to extract explicit, interpretable knowledge from implicit, multi-aspect shared knowledge and use this knowledge to improve the representation and evaluation performance of new prompt essays. We conduct extensive experiments on public datasets. The results show that our approach outperforms current state-of-the-art models and is effective on cross-prompt AES.}, number={23}, journal={Proceedings of the AAAI Conference on Artificial Intelligence}, author={Li, Xia and Pan, Wenjing}, year={2025}, month={Apr.}, pages={24476–24484} }

@misc{gupta2026contextcontentexposingevaluation,
      title={Context Over Content: Exposing Evaluation Faking in Automated Judges}, 
      author={Manan Gupta and Inderjeet Nair and Lu Wang and Dhruv Kumar},
      year={2026},
      eprint={2604.15224},
      archivePrefix={arXiv},
      primaryClass={cs.AI},
      url={https://arxiv.org/abs/2604.15224}, 
}

@inproceedings{ye2025justice,
 author = {Ye, Jiayi and Wang, Yanbo and Huang, Yue and Chen, Dongping and Zhang, Qihui and Moniz, Nuno and Gao, Tian and Geyer, Werner and Huang, Chao and Chen, Pin-Yu and Chawla, Nitesh and Zhang, Xiangliang},
 booktitle = {International Conference on Learning Representations},
 editor = {Y. Yue and A. Garg and N. Peng and F. Sha and R. Yu},
 pages = {102351--102390},
 title = {Justice or Prejudice? Quantifying Biases in LLM-as-a-Judge},
 url = {https://proceedings.iclr.cc/paper_files/paper/2025/file/fdca08d371e4b6c031397909e20043bd-Paper-Conference.pdf},
 volume = {2025},
 year = {2025}
}

@article{harsch2013comparing,
  title={Comparing holistic and analytic scoring methods: Issues of validity and reliability},
  author={Harsch, Claudia and Martin, Guido},
  journal={Assessment in Education: Principles, Policy \& Practice},
  volume={20},
  number={3},
  pages={281--307},
  year={2013},
  publisher={Taylor \& Francis}
}

@article{klein1998analytic,
  title={Analytic versus holistic scoring of science performance tasks},
  author={Klein, Stephen P and Stecher, Brian M and Shavelson, Richard J and McCaffrey, Daniel and Ormseth, Tor and Bell, Robert M and Comfort, Kathy and Othman, Abdul R},
  journal={Applied Measurement in Education},
  volume={11},
  number={2},
  pages={121--137},
  year={1998},
  publisher={Taylor \& Francis}
}

@inproceedings{ICLR2025_646ca7b9,
 author = {Feng, Kehua and Ding, Keyan and Yu, Jing and Qu, Yiwen and Chen, Zhiwen and lv, chengfei and Yu, Gang and Zhang, Qiang and Chen, Huajun},
 booktitle = {International Conference on Learning Representations},
 editor = {Y. Yue and A. Garg and N. Peng and F. Sha and R. Yu},
 pages = {40346--40367},
 title = {SaMer: A Scenario-aware Multi-dimensional Evaluator for Large Language Models},
 url = {https://proceedings.iclr.cc/paper_files/paper/2025/file/646ca7b994bc46afe33d680dbe7ed67a-Paper-Conference.pdf},
 volume = {2025},
 year = {2025}
}

@article{Ridley_He_Dai_Huang_Chen_2021, title={Automated Cross-prompt Scoring of Essay Traits}, volume={35}, url={https://ojs.aaai.org/index.php/AAAI/article/view/17620}, DOI={10.1609/aaai.v35i15.17620}, abstractNote={The majority of current research in Automated Essay Scoring (AES) focuses on prompt-specific scoring of either the overall quality of an essay or the quality with regards to certain traits. In real-world applications obtaining labelled data for a target essay prompt is often expensive or unfeasible, requiring the AES system to be able to perform well when predicting scores for essays from unseen prompts. As a result, some recent research has been dedicated to cross-prompt AES. However, this line of research has thus far only been concerned with holistic, overall scoring, with no exploration into the scoring of different traits. As users of AES systems often require feedback with regards to different aspects of their writing, trait scoring is a necessary component of an effective AES system. Therefore, to address this need, we introduce a new task named Automated Cross-prompt Scoring of Essay Traits, which requires the model to be trained solely on non-target-prompt essays and to predict the holistic, overall score as well as scores for a number of specific traits for target-prompt essays. This task challenges the model’s ability to generalize in order to score essays from a novel domain as well as its ability to represent the quality of essays from multiple different aspects. In addition, we introduce a new, innovative approach which builds on top of a state-of-the-art method for cross-prompt AES. Our method utilizes a trait-attention mechanism and a multi-task architecture that leverages the relationships between each trait to simultaneously predict the overall score and the score of each individual trait. We conduct extensive experiments on the widely used ASAP and ASAP++ datasets and demonstrate that our approach is able to outperform leading prompt-specific trait scoring and cross-prompt AES methods.}, number={15}, journal={Proceedings of the AAAI Conference on Artificial Intelligence}, author={Ridley, Robert and He, Liang and Dai, Xin-yu and Huang, Shujian and Chen, Jiajun}, year={2021}, month={May}, pages={13745–13753} }

@inproceedings{papineni-etal-2002-bleu,
    title = "{B}leu: a Method for Automatic Evaluation of Machine Translation",
    author = "Papineni, Kishore  and
      Roukos, Salim  and
      Ward, Todd  and
      Zhu, Wei-Jing",
    editor = "Isabelle, Pierre  and
      Charniak, Eugene  and
      Lin, Dekang",
    booktitle = "Proceedings of the 40th Annual Meeting of the Association for Computational Linguistics",
    month = jul,
    year = "2002",
    address = "Philadelphia, Pennsylvania, USA",
    publisher = "Association for Computational Linguistics",
    url = "https://aclanthology.org/P02-1040/",
    doi = "10.3115/1073083.1073135",
    pages = "311--318"
}

@misc{zhang2020bertscoreevaluatingtextgeneration,
      title={BERTScore: Evaluating Text Generation with BERT}, 
      author={Tianyi Zhang and Varsha Kishore and Felix Wu and Kilian Q. Weinberger and Yoav Artzi},
      year={2020},
      eprint={1904.09675},
      archivePrefix={arXiv},
      primaryClass={cs.CL},
      url={https://arxiv.org/abs/1904.09675}, 
}

@article{AUT_1,
title = {Beyond semantic distance: Automated scoring of divergent thinking greatly improves with large language models},
journal = {Thinking Skills and Creativity},
volume = {49},
pages = {101356},
year = {2023},
issn = {1871-1871},
doi = {https://doi.org/10.1016/j.tsc.2023.101356},
url = {https://www.sciencedirect.com/science/article/pii/S1871187123001256},
author = {Peter Organisciak and Selcuk Acar and Denis Dumas and Kelly Berthiaume}
}

@article{AUT_2,
title = {Using large language models to evaluate alternative uses task flexibility score},
journal = {Thinking Skills and Creativity},
volume = {52},
pages = {101549},
year = {2024},
issn = {1871-1871},
doi = {https://doi.org/10.1016/j.tsc.2024.101549},
url = {https://www.sciencedirect.com/science/article/pii/S1871187124000877},
author = {Eran Hadas and Arnon Hershkovitz}
}

@article{FPSPI_1,
author = {Treffinger, Donald and Solomon, Marianne and Woythal, Deb},
year = {2012},
month = {09},
pages = {},
title = {Four Decades of Creative Vision: Insights from an Evaluation of the Future Problem Solving Program International (FPSPI)},
volume = {46},
journal = {The Journal of Creative Behavior},
doi = {10.1002/jocb.14}
}

@article{FPSPI_2,
author = {Alt, Dorit and Kapshuk, Yoav and Dekel, Heli},
year = {2022},
month = {11},
pages = {101201},
title = {Promoting Perceived Creativity and Innovative Behavior: Benefits of Future Problem-solving Programs for Higher Education Students},
volume = {47},
journal = {Thinking Skills and Creativity},
doi = {10.1016/j.tsc.2022.101201}
}

\clearpage
\appendix

\section{CGPST Benchmark Details}
\label{App:CGPST Benchmark Details}

Contextually-Grounded and Procedurally-Structured tasks (CGPST) \citep{wang2026teamllmhumanliketeamorientedcollaboration} is a benchmark designed to evaluate the comprehensive problem-solving abilities of LLMs through contextualized multi-step tasks. Compared with traditional creativity benchmarks such as AUT and TTCT, CGPST is substantially more complex and therefore more challenging for automated evaluation. Specifically, CGPST exhibits four key characteristics: \textbf{Contextual Grounding}, where each task is based on a complete future scenario with rich contextual information and detailed problem settings; \textbf{Procedural Structure}, where each task consists of six interdependent and sequential steps progressively leading to the resolution of a real-world problem; \textbf{Process-Oriented Evaluation}, where all intermediate steps are systematically evaluated rather than focusing only on the final response; and \textbf{Multi-Dimensional Assessment}, where each step is evaluated across multiple comprehensive dimensions \citep{Treffinger1995, treffinger2012four, wang2026teamllmhumanliketeamorientedcollaboration}.

The CGPST dataset contains 10 different scenarios, each consisting of 20 complete six-step response samples, resulting in a total of 200 samples. Each sample is annotated with validated scores from two human experts. To facilitate understanding, we provide a complete future scenario in \textit{Ocean Soup Future Scenario} box (Page~\pageref{box:fs}) and an example sample in \textit{A Complete CGPST Sample} box (Page~\pageref{box:sample}). Table~\ref{Table:cgpst_scenarios} presents the themes of all scenarios.

As shown in Table~\ref{Table:Step-wise Information on CGPST}, CGPST consists of six sequential steps. Given a future scenario, LLMs are first required to identify up to eight challenges (\textbf{Step-1}), then select the most promising challenge as an underlying problem (\textbf{Step-2}), and generate up to eight solutions for this problem (\textbf{Step-3}). Subsequently, LLMs create five evaluation criteria for the problem and proposed solutions (\textbf{Step-4}), which are then used to rank the solutions and select the best one (\textbf{Step-5}). Finally, the selected solution is further developed into a comprehensive action plan addressing the underlying problem and generating positive impacts on the future scenario (\textbf{Step-6}). The strong interdependency across all six steps further increases the difficulty of automated evaluation. We provide detailed descriptions of  dimensions for each step in Table~\ref{Table:Step-1 Rubrics} to \ref{Table:Step-6 Rubrics}.

\begin{table}[!t]
\centering
\begin{tabular}{cl}
\toprule
\textbf{Scenario ID} & \textbf{Theme} \\
\midrule

FS1  & Autonomous Transportation \\
FS2  & Terraforming \\
FS3  & Sustainable Development \\
FS4  & Throw Away Society \\
FS5  & Antibiotic Resistance \\
FS6  & Neurotechnology \\
FS7  & Criminal Justice Systems \\
FS8  & Biosecurity \\
FS9  & Food Safety \\
FS10 & Ocean Soup \\
\bottomrule

\end{tabular}

\caption{Overview of the 10 future scenarios on the CGPST dataset. Each scenario contains 20 six-step response samples, resulting in a total of 200 samples annotated with two validated human expert scores.}
\label{Table:cgpst_scenarios}
\end{table}

\section{Baseline Implementation Details}
\label{App:Baseline Implementation Details}

We provide the detailed implementation of the baseline methods as follows.
\paragraph{Direct Score} LLM directly assigns scores to all steps of a response in a single pass. The input includes the future scenario, the full response, dimension descriptions, and corresponding rubrics.
\paragraph{Chain-of-Thought (CoT)} Given the inherent multi-step structure of CGPST, the evaluation is performed step by step, where each step is scored sequentially. Specifically, a single response requires six sequential passes corresponding to the six steps \citep{NEURIPS2022_9d560961}.
\paragraph{Tree-of-Thought (ToT)} ToT follows a tree-structured evaluation process, with each layer aligned to a step on CGPST. At each layer, three independent Score LLMs generate three candidate scores, forming three branches of this layer. A separate Judge LLM then selects the most appropriate score among these candidates as the final output for this step \citep{NEURIPS2023_271db992}.

\paragraph{Graph-of-Thought (GoT)} Built upon ToT, GoT introduces an additional refinement operation for thought transformation. Specifically, for each step, three Score LLMs first generate candidate scores, which are then refined through a self-refinement process before final selection. The Judge LLM then selects the final score from the refined candidates for each step. This refinement mechanism allows iterative improvement of intermediate scoring nodes, forming a graph-structured reasoning process \citep{Besta_2024}.

\paragraph{Table as Thought} This method organizes reasoning within a tabular schema. In the original paper, the table is stored in JSON format, which is consistent with the representation format used in CreaEval. However, unlike CreaEval, both analysis and judging are performed by a single LLM within this framework, making it an \textbf{undecoupled variant of CreaEval}. We therefore regard it as a baseline that shares the same structured representation but does not decouple the evaluation process.

\paragraph{Supervised Fine-Tuning (SFT)} We split the CGPST dataset into training and test sets with a ratio of 7:3. Subsequently, we perform SFT on \texttt{Qwen3.5-9B} using the instruction-response pairs constructed from the training set. After training, the model’s capability is evaluated on the test set.

\paragraph{Scenario-aware Multi-dimensional Evaluator (SaMer)} SaMer introduces a three-branch scoring head atop a frozen \texttt{Qwen3.5-9B} to enable scenario-aware, preference-driven evaluation. During training, the base LLM is frozen and only the head is optimized through a multi-objective loss combining three signals: a dimension prediction loss that learns step-to-dimension relevance, a dimension-level ranking loss that aligns pairwise preferences on individual dimensions, and an overall preference loss that learns the final winner from pairwise comparisons. At inference, only dimensions predicted as relevant receive non-zero weights via softmax, and the overall score is computed as the weighted sum of dimension scores.

\section{Agreement Metric Details}
\label{App:Agreement Metric Details}
On the CGPST dataset, each sample is annotated by two human raters. Therefore, when computing agreement between LLM scores and human annotations, \textbf{we calculate the consistency separately with Human A and Human B for each dimension, and then average the two results as the final agreement score}. Importantly, we do not compute agreement against the averaged human score. This is to avoid misleading cases where Human A gives a score of 4 and Human B gives 8; an LLM score of 6 would appear perfectly accurate if compared to the mean of two humans (6), even though it does not truly match either annotator.

We present three metrics below to facilitate a better understanding of the experimental results.

\paragraph{Pearson correlation coefficient (PCC)}

PCC measures the \textbf{linear correlation} between predicted scores and human annotations. It evaluates how well the model predictions align with human judgments in terms of overall trend consistency.
\begin{equation}
PCC = \frac{\sum_{i=1}^{n}(x_i - \bar{x})(y_i - \bar{y})}
{\sqrt{\sum_{i=1}^{n}(x_i - \bar{x})^2} \sqrt{\sum_{i=1}^{n}(y_i - \bar{y})^2}}
\end{equation}

where $x_i$ and $y_i$ denote the predicted and human scores for sample $i$, and $\bar{x}$ and $\bar{y}$ denote their respective means.

\paragraph{Quadratic weighted kappa (QWK)}

QWK measures the agreement between two raters while taking into account the \textbf{ordinal nature} of the ratings and the degree of disagreement. It is particularly suitable for discrete or ordinal scoring tasks and is one of the most commonly used metrics in automated scoring~\citep{Ridley_He_Dai_Huang_Chen_2021, do-etal-2024-autoregressive}.
\begin{equation}
QWK = 1 - \frac{\sum_{i,j} w_{ij} O_{ij}}
{\sum_{i,j} w_{ij} E_{ij}}
\end{equation}
\begin{equation}
w_{ij} = \frac{(i - j)^2}{(N - 1)^2}
\end{equation}

where $O_{ij}$ and $E_{ij}$ are the observed and expected agreement matrices, respectively, and $w_{ij}$ is the quadratic weight.

\paragraph{Intra-class correlation coefficient (ICC)}

ICC is widely used to assess inter-rater reliability by measuring the proportion of variance attributable to differences between subjects relative to total variance. It reflects the consistency of quantitative measurements across different raters.
\begin{equation}
ICC = \frac{\sigma^2_{\text{between}}}{\sigma^2_{\text{between}} + \sigma^2_{\text{within}}}
\end{equation}

where $\sigma^2_{\text{between}}$ and $\sigma^2_{\text{within}}$ denote the between-subject and within-subject variance, respectively.

\onecolumn
\begin{tcolorbox}[
    colback=gray!5,
    colframe=black,
    title=Ocean Soup Future Scenario,
    breakable
]
\label{box:fs}
\small
\ttfamily

\hspace{2em}As Jobie Sakai leans on the railing of the Ola Kai, she admires the beauty of her ocean paradise. Jobie, a fifth generation Hawaiian devoted to the future of her homeland, is dedicated to her job as an environmental chemist aboard the Ola Kai \#6, one of Hawai'i's floating science laboratories. The Ola Kai (meaning "healthy ocean") Project is a combined effort of the Hawaiian Environmental Council and the University of Hawai'i.

\vspace{1em}
\hspace{2em}Now, in 2035 after 15 active years, the project is struggling to live up to its nickname: the OK Project. Originally, the OK Project focused on the waters affected by Hawai'i's island-generated pollutants. Public interest in the project led to a resurgence in eco-education; recycling and the reduced use of plastics became an accepted part of island life for Hawai'i residents. The "adopt a beach" clean-up program became a popular draw for eco-tourists. However, researchers like Jobie became increasingly aware that their efforts were not enough.

\vspace{1em}
\hspace{2em}The world's largest manufacturers of plastic products border both sides of the Pacific. A ten million square mile system of rotating currents called the North Pacific Gyre has its axis near the 137 islands of the Hawaiian chain. Pacific environmental regulations have historically been weak or disregarded by heavy industrial nations who continue to use these waters as a dumping ground. Consequently, the 1500 mile-long archipelago paradise has been attacked by ocean soup for many years.

\vspace{1em}
\hspace{2em}The soup surrounds Hawai'i, placing the islands and their resources at risk of permanent damage. Especially vulnerable are the sparsely-inhabited northwest islands, the world's largest protected marine sanctuary that is home to many endangered fish, birds, seals, and Hawai'i's beleaguered fishing industry. Eco-tourism routes have been altered to reduce impact on indigenous species and circumnavigated due to the location of floating laboratories.

\vspace{1em}
\hspace{2em}With the increasing damaging effects from ocean soup on the island chain, Jobie and her coworkers realized that the Ola Kai Project's numerous floating labs had to broaden their territory while narrowing their focus. Due to the scope of the damage, the project directors reached out to other groups working in the Pacific and consulted with the National Oceanic and Atmospheric Administration (NOAA). It was determined that the best approach would be to divide up the responsibilities among agencies. Now the OK Project labs focus solely on the battle against microplastics, leaving the collection of larger trash to other organizations.

\vspace{1em}
\hspace{2em}Ola Kai laboratory crews record various data for analysis, keeping track of multiple fish species and beneficial organisms like plankton. Specified lab crews weigh the microplastic debris collected on a weekly basis and compile that data while Jobie and other chemists continue their in-depth examination of the plastic degradation and its effect on the waters surrounding the islands.

\vspace{1em}
\hspace{2em}Collection of debris that is smaller than a pencil eraser has often done more harm than good to sea life. After experimenting with several collection methods, the OK Project currently uses below-surface robotic collectors that randomly collect plastic particles being carried by the currents. Project teams are also experimenting with alternative collection methods, including new nanofiber sieves and use of pollution-dissolving lasers.

\vspace{1em}
\hspace{2em}Ola Kai's floating labs have plastic-to-fuel conversion systems capable of harvesting tons of plastic pollution and converting it into diesel fuel for the labs, thus eliminating the need to return to shore for disposal of the waste in a landfill. In spite of their progress, water samples still show an alarming amount of plastic particles.

\end{tcolorbox}
\twocolumn

\onecolumn
\begin{tcolorbox}[
    colback=gray!5,
    colframe=black,
    title=A Complete CGPST Sample,
    breakable
]
\label{box:sample}
\small
\ttfamily

\#\#\# \textbf{Step-1 Identify Challenges}

1. The concentration of microplastics may have exceeded the density of plankton by tenfold, disrupting the energy input of the base food chain. This challenge arises from the warning in the scenario that "water samples still show an alarming amount of plastic particles."  

2. Subsurface robotic collectors may miss low-velocity eddy zones, creating data gaps and masking local ecological collapse points. This is directly related to the scenario's mention that "after experimenting with several collection methods," weekly weighing is still required, implying sampling limitations.  

3. The plastic-to-fuel conversion system may emit nanoscale black carbon particles, which could exacerbate imbalances in ocean surface heat absorption. This challenge stems from the scenario emphasizing "eliminating the need to return to shore for disposal" without addressing potential secondary emissions. 

4. Endangered species in the northwest islands may face unknown toxicological effects from ingesting micro-fragments of plastics broken down by lasers. The scenario mentions that laser technology "can dissolve pollutants" but does not evaluate the byproducts of fragmentation.  

5. Adjustments to eco-tourism routes around floating laboratories may transfer visitor pressure to other more fragile reefs. This challenge is directly related to the scenario's note that routes were "altered to reduce impact" but without ensuring overall pressure reduction.  

6. Legal exemptions for manufacturers on both sides of the Pacific may cause Hawaiian regional governance to operate in isolation. This challenge is closely related to the scenario's statement that "environmental regulations have historically been weak or disregarded."  

7. The efficiency of the plastic-to-diesel system may suddenly decline due to sea spray corrosion, forcing laboratories to rely on land-based resupply. This challenge is implied in the scenario mentioning "harvesting tons of plastic" without considering long-term durability.  

8. Data protocols among the network of floating labs may be incompatible, hindering multinational collaboration in compiling a comprehensive microplastic hotspot map. This challenge is directly related to the scenario emphasizing "dividing responsibilities among agencies" without a unified standard.

\vspace{2em}

\#\#\# \textbf{Step-2 Select an Underlying Problem}

Challenge ID: 1. In 2035, in the waters surrounding the Hawaiian Islands, when microplastic concentrations may have exceeded the density of plankton tenfold, threatening the energy input at the base of the food chain, how can we reduce the overwhelming proportion of microplastics at the bottom of the food chain to restore energy flow in the Pacific ecosystem and protect the future of endangered species?  

Time: 2035

Location: Waters surrounding the Hawaiian Islands

Theme: Microplastic Management and Ecological Restoration

\vspace{2em}

\#\#\# \textbf{Step-3 Produce Solutions}

1. The Ola Kai project chemistry team will deploy glycosylated nanosponges, dispersing 2 tons within a 20 km radius of the Ola Kai mooring point by August 2035. Subsurface robots will recover the flocs and recycle them through the onboard plastic-to-diesel system, directly reducing microplastic ingestion by plankton, lowering the proportion of plastics at the base of the food chain, restoring energy flow, and protecting endangered species. 

2. Google X Lab and Hawaiian community divers will run a crowdsourced "photoacoustic unmanned vessel + AR snorkeling goggles" collection program across the northwest islands by December 2035. Unmanned vessels will map microplastic clouds in real-time using laser sonar, while AR glasses guide divers to precise retrieval points, clearing high-density fragments weekly to reduce microplastic accumulation at the base of the food chain and preserve Pacific ecosystem energy balance.  

3. NOAA and Hawaiian Electric will pilot a 2-nautical-mile-diameter "bubble curtain + photocatalytic net" system north of Kaua'i by October 2035. Wave-driven pumps will concentrate microplastics, which are then broken down by photocatalytic nets into short-chain acids absorbable by phytoplankton, reducing microplastic dominance on plankton and restoring baseline energy input.  

4. SpaceX and a local high school team will launch the CubeSat constellation "KiloEye" by July 2035. Weekly scans of the 137 Hawaiian Islands will use hyperspectral data to direct Ola Kai drones for targeted microplastic removal, lowering the risk of plankton mis-ingestion at the source and ensuring Pacific ecosystem energy flow is rebalanced.  

5. Japan's SpiraNova and the University of Hawai'i will plant 300 "biopolymer-coated kelp ropes" off the west coast of the Big Island in Q3 2035. Kelp leaves will adsorb microplastics, and harvested ropes will be processed into high-value composites, directly removing plastics at the base of the food chain and generating revenue while protecting endangered species’ habitats.  
6. The Ola Kai project biology team will release living blue-green algae “Plastic Sentinel” strains in a 500-hectare demonstration area off the northwest reef of the main Hawaiian Island by November 2035. These algae continuously secrete degrading enzymes to break down 0.1–1 mm microplastics, reducing plankton ingestion, restoring baseline energy input, and protecting the Pacific ecosystem.  

7. MantaSync and the University of Hawai'i will deploy five “body-mounted” manta ray filtration units along the Maui–Ni’ihau route by September 2035. These units capture microplastics in real-time during swimming and ferment them into manta ray body oils. Due to their large feeding area, they significantly dilute plastics at the base of the food chain, reducing pollution at entry points. 

8. A local cruise company and the state government will retrofit the first ferry deck in Honolulu Harbor into a “container-scale algae farm” by October 2035. Chlorella algae will consume microplastics from tourist wastewater and convert them into aviation-grade biodiesel, reducing overall plastic input from tourism, lowering plankton ingestion, and maintaining Pacific energy flow.

\vspace{2em}

\#\#\# \textbf{Step-4 Select Criteria}

1. Which approach can most rapidly reduce the net microplastic content inside plankton?  

2. Which pathway requires the lowest one-time capital investment?  
3. Which model has the highest potential for rapid global replication and scaling?  

4. Which technology poses the least risk of secondary ecological disturbance to endangered species?

5. Which solution maintains the lowest full-life-cycle carbon footprint over 15 years?

\vspace{2em}

\#\#\# \textbf{Step-5 Apply Criteria to Top Solution}

Solution ID | Criterion 1 | Criterion 2 | Criterion 3 | Criterion 4 | Criterion 5 | Total Score  

1 | 8 | 6 | 7 | 7 | 6 | 34  

2 | 6 | 8 | 5 | 6 | 7 | 32  

3 | 7 | 3 | 6 | 3 | 5 | 24  

4 | 5 | 5 | 8 | 5 | 4 | 27  

5 | 4 | 4 | 4 | 4 | 3 | 19  

6 | 3 | 7 | 3 | 5 | 8 | 26  

7 | 2 | 2 | 2 | 2 | 2 | 10  

8 | 1 | 1 | 1 | 1 | 1 | 5  

The solution with the highest total score is: 1. 

The solution is: The Ola Kai project chemistry team will deploy glycosylated nanosponges, dispersing 2 tons within a 20 km radius of the Ola Kai mooring point by August 2035. Subsurface robots will recover the flocs and recycle them through the onboard plastic-to-diesel system, directly reducing microplastic ingestion by plankton, lowering the proportion of plastics at the base of the food chain, restoring energy flow, and protecting endangered species. 

\vspace{2em}

\#\#\# \textbf{Step-6 Develop an Action Plan}

The solution with the highest total score is: 1.

\vspace{1em}
The solution is: The Ola Kai project chemistry team will deploy glycosylated nanosponges, dispersing 2 tons within a 20 km radius of the Ola Kai mooring point by August 2035. Subsurface robots will recover the flocs and recycle them through the onboard plastic-to-diesel system, directly reducing microplastic ingestion by plankton, lowering the proportion of plastics at the base of the food chain, restoring energy flow, and protecting endangered species.

\vspace{1em}  
Action Plan:

\vspace{1em}
1. Problem Closure  
Nanosponges use a "molecular magnet" mechanism to selectively adsorb 0.1–1 mm microplastics, aggregating them into millimeter-scale flocs that cannot be ingested by plankton. Recovered flocs are immediately converted into diesel, achieving a “collect-convert-use” zero-waste cycle, directly reducing the overwhelming proportion of microplastics at the base of the food chain.

\vspace{1em}
2. Implementation Steps and Timeline

    \vspace{1em}
   • Phase A – R\&D and Validation (Now–Oct 2025): Ola Kai Chemistry Team × MIT Materials Department iterate the third-generation biodegradable nanosponges and complete biotoxicity-degradation tests.

   \vspace{1em}
   • Phase B – Pilot Demonstration (Nov 2025–Apr 2026): Deploy 100 kg in South Bay, Oahu; 30-day monitoring shows $\geq 70\%$ reduction of microplastics in plankton.

   \vspace{1em}
   • Phase C – Scale-up Production (May 2026–Feb 2027): Hawaii biorefinery produces 2 tons per month and establishes freeze-dry packaging chain.

   \vspace{1em}
   • Phase D – Full Deployment (Mar 2027–Jul 2035): Deploy 2 tons in March, June, September each year, covering a 20 km radius grid; six “Kokua-$\gamma$” subsurface robots operate in shifts, producing 400 L/day diesel for self-use.

   \vspace{1em}
   • Phase E – Monitoring and Iteration (Parallel): Weekly underwater imaging + Raman verification to maintain microplastic concentration <50\% of plankton; formula updated every six months.

\vspace{1em}
3. Resources and Responsibilities  
   • Funding: Ola Kai Research \$300k + NOAA Innovation Fund \$400k + State Green Bonds \$300k + Carbon Credit Pre-sale; total $\leq$ \$1M.

   \vspace{1em}
   • Team: Chemistry team handles materials, MIT provides R\&D, NOAA provides monitoring platform, State Environmental Department supervises approvals.

\vspace{1em}
4. Risks and Contingency

    \vspace{1em}
   • Nanomaterial leakage: Three passive samplers monitor in real-time; $>10\,\mu\text{g}/\text{L}$ triggers magnetic recovery nets.

   \vspace{1em}
   • Robot malfunction: 1:1 spare parts + 48-hour offshore repair; if failure rate >15\%, NOAA backup ROVs are deployed.

   \vspace{1em}
   • Regulatory delays: Suspension during typhoon season; stock maintained at 1.5× safety level.

\vspace{1em}
5. Impacts and Scaling

    \vspace{1em}
   • Local: By 2028, microplastic content in plankton decreases by 80\%, coral spawning rates increase by 30\%.

   \vspace{1em}
   • Regional: By 2030, open “Nanosponges Sharing Depot” allows replication in Guam, Palau, Tuvalu.

   \vspace{1em}
   • Global: By 2032, included in IMO Green Shipping Guidelines; long-haul fleets can treat plastics in-transit, establishing a Pacific-wide “food chain firewall.”

\end{tcolorbox}
\twocolumn

\section{Complete Prompts}
\label{App:Complete Prompts}

\subsection{SoT-LLM System Prompts}
\begin{tcolorbox}[
    colback=gray!5,
    colframe=black,
    title=SoT-LLM System Prompts,
    breakable
]
\ttfamily

You are an objective structured information extraction model (SoT (Structure-of-Thought)-LLM). Your task is to transform the input textual response into a structured representation (SoT) that strictly conforms to a predefined schema.

\vspace{1em}

The input textual response belongs to one step of CGPST (Contextualized-Grounding and Procedurally-Structured Task). In this task, the respondent is required to complete the following six steps sequentially based on a given future scenario:

\vspace{1em}

1.Identify Challenges: Identify up to 8 reasonable challenges based on the future scenario. 

2.Select an Underlying Problem: Select the most promising and meaningful challenge from Step-1 as the underlying problem. 

3.Produce Solutions: Generate up to 8 solutions for the underlying problem from Step-2. 

4.Select Criteria: Generate 5 evaluation criteria for the solutions from Step-3. 

5.Apply Criteria to Top Solution: Rank the solutions from Step-3 using the criteria from Step-4 and select the highest-scoring solution. 

6.Develop an Action Plan: Develop the top solution from Step-5 into a comprehensive action plan to address the underlying problem from Step-2. 
\end{tcolorbox}

\subsection{SoT-LLM User Prompts}
\label{App(sub):SoT-LLM User Prompts}
In the following SoT-LLM user prompts, \{step\_num\} refers to the current step, \{future\_scenario\} to the corresponding future scenario, \{step\_schema\} to the predefined SoT extraction format (see Appendix~\ref{App(sub):SoT Schema for SoT-LLM}), \{score\_dimensions\_wo\_rubrics\} to the Description fields of each evaluation dimension for that step (see the Description columns in Tables~\ref{Table:Step-1 Rubrics}–~\ref{Table:Step-6 Rubrics}), and \{raw\_text\} to the original response text of the step.
\begin{tcolorbox}[
    colback=gray!5,
    colframe=black,
    title=SoT-LLM User Prompts,
    breakable
]
\ttfamily

\#\#\# Your Task
Please perform structured extraction on the response text for \{step\_num\} of CGPST according to the given schema, and provide the results in the required format.

\vspace{2em}

\#\#\# Future Scenario
The CGPST future scenario corresponding to the response text is as follows:

\vspace{1em}

\{future\_scenario\}

\vspace{2em}

\#\#\# Schema

\{step\_schema\}

\vspace{2em}

\#\#\# Description of Schema Fields

1. sot\_evidence:

- Refers to fine-grained and verifiable observational information extracted from the original response text, rather than directly copying the response content. 

- Extraction must be strictly grounded in the original content and performed honestly, without forcing interpretations or artificially exaggerating content merely to match certain dimensions. 

- Analyze whether the response satisfies the requirements of each dimension, providing evidence for subsequent scoring.

- If there is no relevant evidence for a certain dimension in the response text, the corresponding dimension\_evidence should be filled with None. Do not fabricate information.

\vspace{1em}

2. task\_status\_memory:

- Summarize the response for use as contextual reference in subsequent steps. The summary must remain complete while excluding irrelevant or redundant information.

- If multiple responses are provided, summarize each one separately.

\vspace{2em}

\#\#\# Introduction to Step Dimensions

\{score\_dimensions\_wo\_rubrics\}

\vspace{2em}

\#\#\# Response Text for This Step

\{raw\_text\}

\vspace{2em}

\#\#\# Task Requirements

1. Perform structured information extraction and classification solely based on the original response text. Be factual and do not introduce external knowledge or new information.

2. Do not make unsupported subjective inferences or extensions.

3. Strictly follow the provided schema fields when filling in the output. All fields must strictly conform to the specified data types. Do not add any fields that are not defined in the schema, and do not modify any existing schema fields.

4. Each dimension includes corresponding examples. Follow the analytical style of these examples and conduct similar dimension-by-dimension analysis.

5. If multiple responses exist, analyze them separately and identify them using labels such as "item\_1", "item\_2", etc.

6. The output must be in strict JSON format and must not contain any additional explanations.

\end{tcolorbox}

\subsection{Judge-LLM System Prompts}
\begin{tcolorbox}[
    colback=gray!5,
    colframe=black,
    title=Judge-LLM System Prompts,
    breakable
]
\ttfamily

You are an objective and impartial evaluation model responsible for assessing the response based on the provided structured result (SoT (Structure-of-Thought) output).

\vspace{1em}

The input textual response belongs to CGPST (Contextualized-Grounding and Procedurally-Structured Task). In this task, the respondent is required to complete the following six steps sequentially based on a given future scenario:

\vspace{1em}

1.Identify Challenges: Identify up to 8 reasonable challenges based on the future scenario. 

2.Select an Underlying Problem: Select the most promising and meaningful challenge from Step-1 as the underlying problem. 

3.Produce Solutions: Generate up to 8 solutions for the underlying problem from Step-2. 

4.Select Criteria: Generate 5 evaluation criteria for the solutions from Step-3. 

5.Apply Criteria to Top Solution: Rank the solutions from Step-3 using the criteria from Step-4 and select the highest-scoring solution. 

6.Develop an Action Plan: Develop the top solution from Step-5 into a comprehensive action plan to address the underlying problem from Step-2. 
\end{tcolorbox}

\subsection{Judge-LLM User Prompts}
\label{App(sub):Judge-LLM User Prompts}
In the following Judge-LLM user prompts, \{future\_scenario\} refers to the corresponding future scenario, \{sot\_output\} to the concatenated SoT results of all six steps extracted by SoT-LLM, \{score\_dimensions\} to the Description and Rubrics fields of all evaluation dimensions across all steps (see the Description and Rubrics columns in Tables~\ref{Table:Step-1 Rubrics}–~\ref{Table:Step-6 Rubrics}), and \{output\_template\} to the output scoring JSON format (see Appendix~\ref{App(sub):Score Template for Judge-LLM}).
\begin{tcolorbox}[
    colback=gray!5,
    colframe=black,
    title=Judge-LLM User Prompts,
    breakable
]
\ttfamily

\#\#\# Your Task
Please evaluate a response text from CGPST (covering all six steps) based on the provided structured result (SoT output). Score the response according to the given dimensions and score ranges, and provide the results in the required format.

\vspace{2em}

\#\#\# Future Scenario
The CGPST future scenario corresponding to the response text is as follows:

\vspace{1em}

\{future\_scenario\}

\vspace{2em}

\#\#\# SoT Structured Result

\{sot\_output\}

\vspace{2em}

\#\#\# Introduction to the Scoring Dimensions for Each Step

\{score\_dimensions\}

\vspace{2em}

\#\#\# Scoring Output Format

\{output\_template\}

\vspace{2em}

\#\#\# Task Requirements

1. Each dimension must be evaluated independently and strictly according to the provided scoring dimension descriptions. Do not use relative or vague standards; scores must follow the specified scoring ranges exactly.

2. All scoring must be based on the provided SoT structured result.

3. The output must be in strict JSON format. Field names must not be modified, and no fields may be added, removed, or omitted. Do not include any additional explanations.

4. Only output numeric scores. Do not include words such as “points” or “score”.

\end{tcolorbox}

\subsection{SoT Schema for SoT-LLM}
\label{App(sub):SoT Schema for SoT-LLM}
Phase 1 performs Evidence extraction in the form of Structure-of-Thought (SoT) as follows, which is used as the \{step\_schema\} field in Appendix~\ref{App(sub):SoT-LLM User Prompts}. Steps-1, 3, and 4 are composed of multiple response items: Step-1 involves up to 8 challenges, Step-3 up to 8 proposed solutions, and Step-4 5 criteria. Therefore, evidence extraction is performed at the item level for these steps. For the remaining steps, which consist of a single response item, evidence is extracted directly at each dimension. After evidence extraction is performed for each step by SoT-LLM, the extracted evidence from all steps are concatenated and passed to Judge-LLM as the \{sot\_output\} field in Appendix~\ref{App(sub):SoT-LLM User Prompts}.

\begin{tcolorbox}[
    colback=gray!5,
    colframe=black,
    title=Step-1 and Step-3 SoT Schema,
    breakable
]
\ttfamily
\begin{verbatim}
{
  "sot_evidence": {
    "item_1": {
      "Fluency": "...",
      "Flexibility": "...",
      "Elaboration": "...",
      "Originality": "..."
    },
    "item_2": {
      "Fluency": "...",
      "Flexibility": "...",
      "Elaboration": "...",
      "Originality": "..."
    },
    ...
  },
  "task_status_memory": "..."
}
\end{verbatim}

\end{tcolorbox}

\begin{tcolorbox}[
    colback=gray!5,
    colframe=black,
    title=Step-2 SoT Schema,
    breakable
]
\ttfamily
\begin{verbatim}
{
  "sot_evidence": {
    "Condition Phrase": "...",
    "Stem & KVP": "...",
    "Purpose": "...",
    "FS Parameters": "...",
    "Focus": "...",
    "Adequacy": "..."
  },
  "task_status_memory": "..."
}
\end{verbatim}

\end{tcolorbox}

\begin{tcolorbox}[
    colback=gray!5,
    colframe=black,
    title=Step-4 SoT Schema,
    breakable
]
\ttfamily
\begin{verbatim}
{
  "sot_evidence": {
    "item_1": {
      "Correctly Written": "...",
      "Relevance": "..."
    }
  },
  "task_status_memory": "..."
}
\end{verbatim}

\end{tcolorbox}

\begin{tcolorbox}[
    colback=gray!5,
    colframe=black,
    title=Step-5 SoT Schema,
    breakable
]
\ttfamily
\begin{verbatim}
{
  "sot_evidence": {
    "item_1": {
      "Correctly Used": "..."
    }
  },
  "task_status_memory": "..."
}
\end{verbatim}

\end{tcolorbox}

\begin{tcolorbox}[
    colback=gray!5,
    colframe=black,
    title=Step-6 SoT Schema,
    breakable
]
\ttfamily
\begin{verbatim}
{
  "sot_evidence": {
    "Relevance": "...",
    "Effectiveness": "...",
    "Criteria": "...",
    "Impact": "...",
    "Humaneness": "...",
    "Development": "..."
  },
  "task_status_memory": "..."
}
\end{verbatim}

\end{tcolorbox}

\subsection{Score Template for Judge-LLM}
\label{App(sub):Score Template for Judge-LLM}
The following JSON template is used for SoT-based scoring by Judge-LLM, and corresponds to the \{output\_template\} field in Appendix~\ref{App(sub):Judge-LLM User Prompts}. All baselines adopt the same scoring template. For step-wise methods such as CoT, ToT, and GoT, only the corresponding step-specific sub-template is extracted and used for evaluation at each step. 

In addition, Steps 1, 3, and 4 require item-level scoring, and the final step-level scores for each dimension are aggregated and reported in the "summary" field. Steps-1 and 3 compute \textit{Flexibility} as the number of distinct categories covered across all challenges or solutions. Therefore, item-level scoring in this dimension is not required, i.e., \textit{Flexibility} appears only in the "summary" field.

\begin{tcolorbox}[
    colback=gray!5,
    colframe=black,
    title=Score Template,
    breakable
]
\ttfamily
\begin{verbatim}
{
  "Step-1": {
    "each_challenge": [
      {
        "id": "...",
        "Fluency": "...",
        "Elaboration": "...",
        "Originality": "..."
      },
      {
        "id": "...",
        "Fluency": "...",
        "Elaboration": "...",
        "Originality": "..."
      },
      ...
    ],
    "summary": {
      "Fluency": "...",
      "Flexibility": "...",
      "Elaboration": "...",
      "Originality": "...",
      "Overall": "..."
    }
  },
  "Step-2": {
    "Condition Phrase": "...",
    "Stem & KVP": "...",
    "Purpose": "...",
    "FS Parameters": "...",
    "Focus": "...",
    "Adequacy": "...",
    "Overall": "..."
  },
  "Step-3": {
    "each_solution": [
      {
        "id": "...",
        "Fluency": "...",
        "Elaboration": "...",
        "Originality": "..."
      },
      {
        "id": "...",
        "Fluency": "...",
        "Elaboration": "...",
        "Originality": "..."
      },
      ...
    ],
    "summary": {
      "Fluency": "...",
      "Flexibility": "...",
      "Elaboration": "...",
      "Originality": "...",
      "Overall": "..."
    }
  },
  "Step-4": {
    "each_criteria": [
      {
        "id": "...",
        "Correctly Written": "...",
        "Relevance": "..."
      },
      {
        "id": "...",
        "Correctly Written": "...",
        "Relevance": "..."
      },
      ...
    ],
    "summary": {
      "Correctly Written": "...",
      "Relevance": "...",
      "Overall": "..."
    }
  },
  "Step-5": {
    "Correctly Used": "...",
    "Overall": "..."
  },
  "Step-6": {
    "Relevance": "...",
    "Effectiveness": "...",
    "Criteria": "...",
    "Impact": "...",
    "Humaneness": "...",
    "Development": "...",
    "Overall": "..."
  }
}
\end{verbatim}

\end{tcolorbox}

\begin{table*}[t]
\centering
\small

\begin{tabular}{l|llcll}
\toprule
\textbf{Model} & \textbf{Abbreviations} & \textbf{Type} & \textbf{Parameters} & \textbf{Access} & \textbf{URL} \\ 
\midrule
qwen3.6-plus-2026-04-02 & \texttt{qwen3.6-plus} & Closed-source & - & API & \href{https://bailian.console.aliyun.com/cn-beijing?spm=5176.29597918.J_SEsSjsNv72yRuRFS2VknO.2.34407b08ZxRs7B&tab=model#/model-market/detail/qwen3.6-plus-2026-04-02?serviceSite=asia-pacific-china}{\color{blue}{Qwen Link}} \\

deepseek-v4-pro & \texttt{deepseek-v4-pro} & Open-source & 862B & API & \href{https://api-docs.deepseek.com/}{\color{blue}{Deepseek Link}} \\

gemini-3.1-pro-preview & \texttt{gemini-3.1-pro} & Closed-source & - & API & \href{https://ai.google.dev/gemini-api/docs/models/gemini-3.1-pro-preview}{\color{blue}{Google Link}} \\

gpt-5.4 & \texttt{gpt-5.4} & Closed-source & - & API & \href{https://developers.openai.com/api/docs/models/gpt-5.4}{\color{blue}{OpenAI Link}} \\
\bottomrule
\end{tabular}

\caption{Overview of LLMs used in the experiments, including model type, parameter size, access mode, and links.}
\label{Table:LLMs_list}
\end{table*}

\section{LLM Details}
\label{App:LLM Details}
Table~\ref{Table:LLMs_list} summarizes the four LLMs used in our experiments. All experiments were conducted via official APIs.

\section{Temperature Settings}
\label{App:Temperature Settings}
The temperature parameter affects the performance of LLMs: lower values yield more consistent outputs, while higher values promote diversity. As a result, we conduct a small-scale pilot study to examine the effect of temperature on our proposed CreaEval. Specifically, we randomly sample \textbf{10 responses} from the CGPST dataset and select \textbf{Step-1 and Step-2} for analysis. Three temperature settings (\textbf{0.2, 0.5, and 0.8}) are evaluated.

In Memory-augmented Extraction phase, evidence extraction is performed \textbf{five times} for each response and dimension across all temperature settings. Since the evidence corresponding to the same response and dimension should remain consistent across repeated runs, we evaluate extraction stability using \textbf{Self-BLEU} \citep{papineni-etal-2002-bleu} and \textbf{BERTScore} \citep{zhang2020bertscoreevaluatingtextgeneration}. Self-BLEU measures \textbf{lexical similarity} by calculating token overlap between generated texts, while BERTScore leverages pretrained contextual embeddings to evaluate \textbf{semantic similarity} between texts. Higher scores on these metrics indicate greater similarity among the extracted evidence across repeated runs, reflecting more stable extraction performance under the corresponding temperature setting. Specifically, for a given dimension, five repeated runs produce five pieces of evidence $S = \{ s_1, s_2, s_3, s_4, s_5 \}$, and Self-BLEU is computed as follows:
\begin{equation}
BLEU_{r}=BLEU(s_{r}, \; S \setminus s_{r})
\end{equation}
\begin{equation}
Self-BLEU=\frac{1}{5} \sum_{r=1}^{5} BLEU_{r}
\end{equation}

BERTScore is computed as follows:
\begin{equation}
BERT_{r}=\frac{1}{4}\sum_{k \neq r}bertscore\big(s_{r}, s_{k}\big)
\end{equation}
\begin{equation}
BERTScore=\frac{1}{5} \sum_{r=1}^{5} BERT_{r}
\end{equation}

The similarity results of the four SoT-LLMs under different temperature settings across all dimensions are reported in Tables~\ref{Table:pre_experiment_bertscore} and ~\ref{Table:pre_experiment_selfbleu}. When the temperature is set to 0.2, all four LLMs exhibit the highest levels of both semantic and lexical similarity, indicating that lower temperature leads to more stable evidence extraction. \textbf{Therefore, we set the temperature of SoT-LLM to 0.2.}

In Evidence-based Judging phase, for each Judge-LLM, we use the evidence generated at a temperature of 0.2 as the basis for scoring. Each response is evaluated \textbf{five times}, and we compute the \textbf{variance} of the five scores for each dimension. Since the same evidence should theoretically lead to identical scores, a lower variance indicates more stable and consistent scoring. Table~\ref{Table:pre_experiment_variance} presents the variance results of the four LLMs under three temperature settings. The variance remains below 0.2 across all temperatures, indicating that the evidence-based scoring process is highly stable. Moreover, all four LLMs achieve the lowest variance when the temperature is set to 0.2. \textbf{Therefore, we set the temperature of Judge-LLM to 0.2.}

\section{Exploration on Evidence-Based Supervised Fine-Tuning}
\label{App:SFT_Evidence}

To evaluate the effectiveness of evidence-based scoring, we perform an additional experiment named SFT\_Evidence. 
Specifically, instead of mapping raw multi-step responses directly to rubric scores (Vanilla SFT), SFT\_Evidence fine-tunes the model on the second phase of our CreaEval using extracted \textbf{evidence-score} pairs under the same 7:3 train-test split.

As presented in Table~\ref{Table:SFT_Evidence}, SFT\_Evidence achieves a substantial performance improvement over Vanilla SFT (0.632 vs. 0.4665 in Average QWK) and closely approaches the performance of CreaEval (0.6388). This result highlights the critical role of intermediate evidence in our decoupled design for accurate creativity evaluation. More importantly, while SFT\_Evidence relies on additional supervised fine-tuning with evidence-score pairs, CreaEval achieves superior accuracy in a completely train-free manner without requiring extra training overhead, making it a more practical and preferable solution.

\begin{table*}[t]
\centering

\begin{subtable}{\textwidth}
\centering
\resizebox{\textwidth}{!}{
\begin{tabular}{l|
cccc
ccc
cccc}
\toprule

\multirow{2}{*}{\textbf{Method}}
& \multicolumn{4}{c}{Step-1}
& \multicolumn{3}{c}{Step-2}
& \multicolumn{4}{c}{Step-3} \\

\cmidrule(lr){2-5}
\cmidrule(lr){6-8}
\cmidrule(lr){9-12}

& Fluency & Flexibility & Elaboration & Originality
& Integrity & Focus & Adequacy
& Fluency & Flexibility & Elaboration & Originality \\

\midrule

SFT
& 0.5027 & 0.4324 & 0.4765 & 0.5353
& 0.476 & 0.4869 & 0.4194
& 0.4723 & 0.4662 & 0.4756 & 0.4541 \\

SFT\_Evidence & \underline{0.7486} & \underline{0.6509} & \textbf{0.6573} & \underline{0.6417} & \underline{0.7712} & \underline{0.5499} & \textbf{0.4864} & \underline{0.8267} & \underline{0.6905} & \textbf{0.6979} & \underline{0.6585} \\

CreaEval
& \textbf{0.7505} & \textbf{0.6665} & \underline{0.656} & \textbf{0.665}
& \textbf{0.7756} & \textbf{0.554} & \underline{0.4831}
& \textbf{0.8315} & \textbf{0.6975} & \underline{0.6969} & \textbf{0.6965} \\

\bottomrule
\end{tabular}
}
\end{subtable}

\vspace{1mm}

\begin{subtable}{\textwidth}
\centering
\resizebox{\textwidth}{!}{
\begin{tabular}{l|
cc
c
cccccc
c}
\toprule

\multirow{2}{*}{\textbf{Method}}
& \multicolumn{2}{c}{Step-4}
& \multicolumn{1}{c}{Step-5}
& \multicolumn{6}{c}{Step-6}
& \multirow{2}{*}{AVG} \\

\cmidrule(lr){2-3}
\cmidrule(lr){4-4}
\cmidrule(lr){5-10}

& Correctly Written & Relevance
& Correctly Used
& Relevance & Effectiveness & Criteria & Impact & Humaneness & Development
& \\

\midrule

SFT & 0.4599 & 0.4572 & 0.5317 & 0.4408
& 0.4469 & 0.4582 & 0.4694
& 0.4065 & 0.4617 & 0.4665 \\

SFT\_Evidence & \underline{0.7888} & \underline{0.5349} & \textbf{0.948} & \textbf{0.5449} & \underline{0.5089} & \textbf{0.5315} & \underline{0.4691} & \underline{0.4365} & \underline{0.4975} & \underline{0.632} \\

CreaEval
& \textbf{0.7933} & \textbf{0.535}
& \underline{0.9439}
& \underline{0.5366} & \textbf{0.5214} & \underline{0.5276} & \textbf{0.4952} & \textbf{0.4388} & \textbf{0.5119}
& \textbf{0.6388} \\

\bottomrule
\end{tabular}
}
\end{subtable}

\caption{Consistency (QWK) comparison among Vanilla SFT, SFT\_Evidence, and CreaEval on the CGPST benchmark. \textbf{Bold} indicates the highest score and \underline{underline} indicates the second highest score. The AVG column summarizes the overall average score.}
\label{Table:SFT_Evidence}
\end{table*}

\section{Complete Results}
\label{App:Complete Results}
We report QWK results across all methods and dimensions in the main paper (Table~\ref{Table:QWK_results}), and present the corresponding PCC and ICC results in Tables~\ref{Table:PCC_results} and~\ref{Table:ICC_results}, respectively.

\section{PCC and ICC Results for CreaEval Robustness Analysis}
\label{App:PCC and ICC Results for CreaEval Robustness Analysis}
\begin{figure}[!t]
    \centering
    \includegraphics[width=\columnwidth]{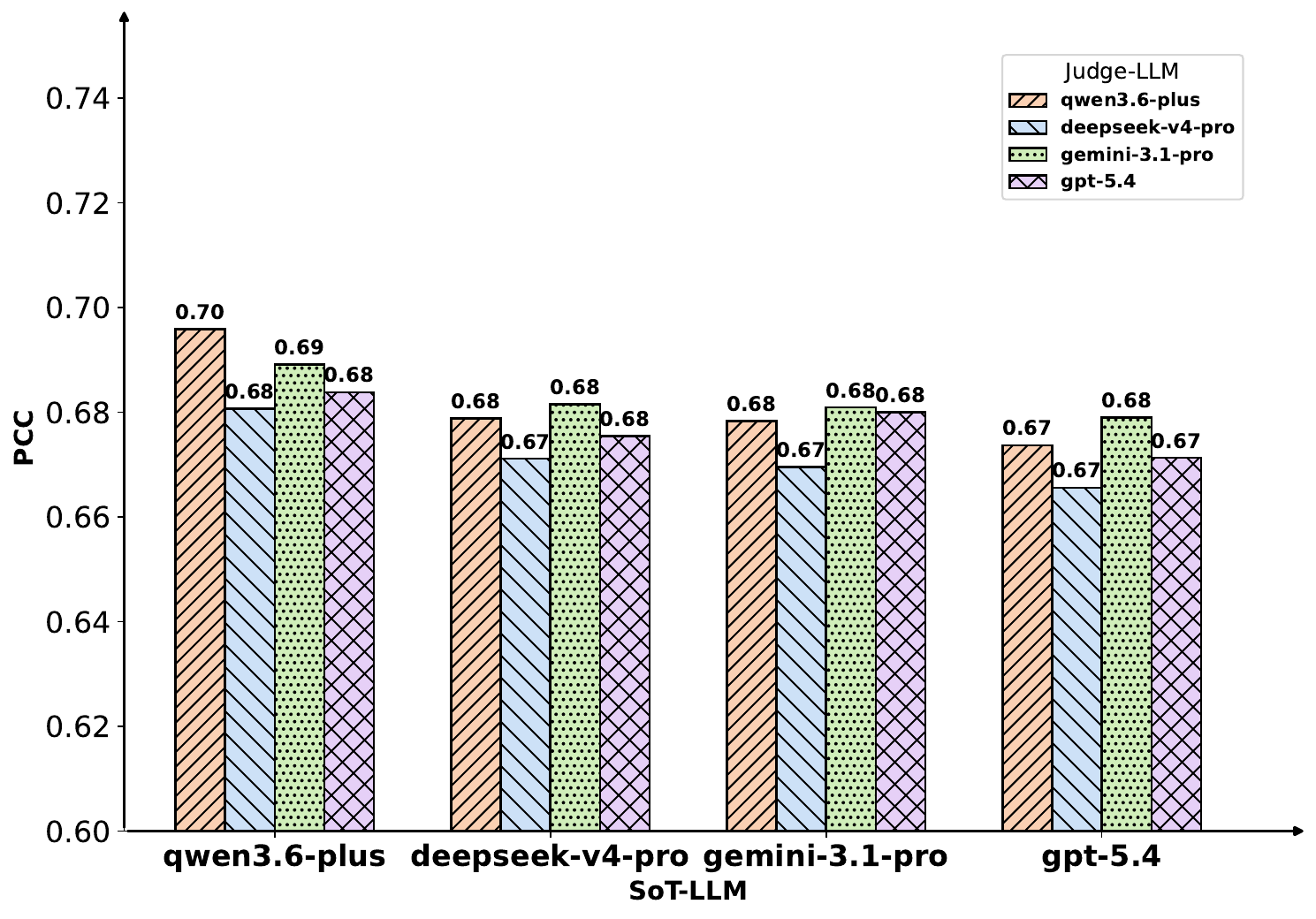} 
    \caption{PCC Results across Different SoT-LLMs.}
    \label{Figure:PCC across Different SoT-LLMs}
\end{figure}

\begin{figure}[!t]
    \centering
    \includegraphics[width=\columnwidth]{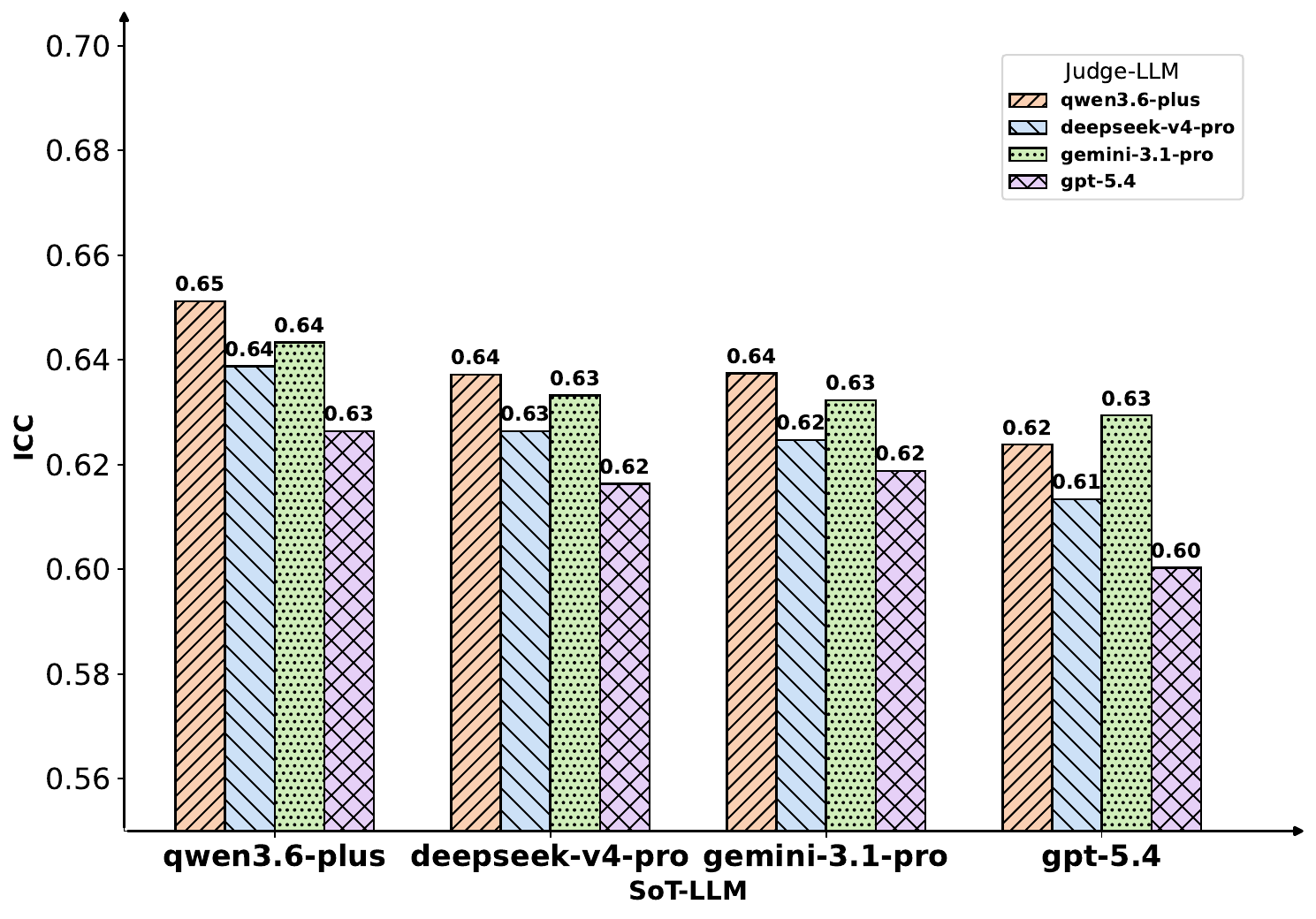} 
    \caption{ICC Results across Different SoT-LLMs.}
    \label{Figure:ICC across Different SoT-LLMs}
\end{figure}

In Section~\ref{Section:Impact of SoT-LLM in CreaEval}, we examine the robustness of the CreaEval framework by replacing the SoT-LLM with different LLMs and report the QWK results. The PCC and ICC results are shown in Figures~\ref{Figure:PCC across Different SoT-LLMs} and~\ref{Figure:ICC across Different SoT-LLMs}, respectively. The PCC values are consistently above 0.65, while the ICC values exceed 0.6 across all settings, demonstrating that CreaEval does not rely on any specific LLM and can serve as a general and robust evaluation framework.

\section{Heatmap Distributions across Dimensions}
\label{App:Heatmap Distributions across Dimensions}
\begin{figure*}[!t]
    \centering
    \includegraphics[width=\textwidth]{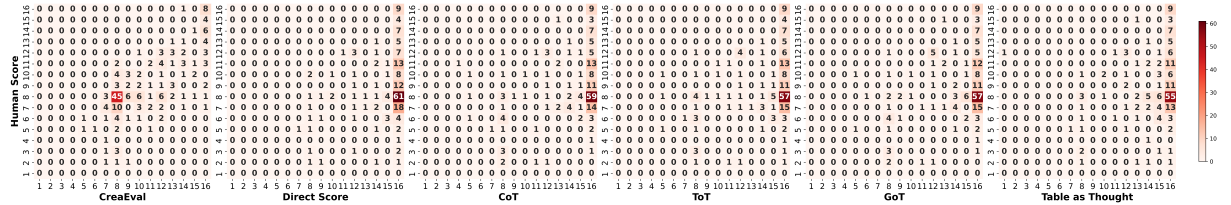} 
    \caption{Headmap between human score and all methods on \textit{Elaboration} of Step-1.}
    \label{Figure:Headmap on Elaboration of Step-1}
\end{figure*}

\begin{figure*}[!t]
    \centering
    \includegraphics[width=\textwidth]{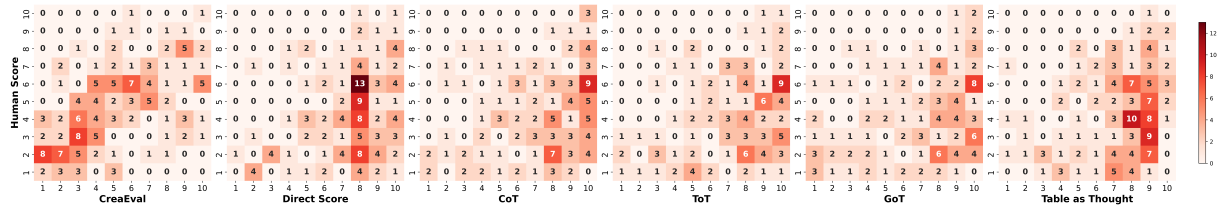} 
    \caption{Headmap between human score and all methods on \textit{Originality} of Step-1.}
    \label{Figure:Headmap on Originality of Step-1}
\end{figure*}

\begin{figure*}[!t]
    \centering
    \includegraphics[width=\textwidth]{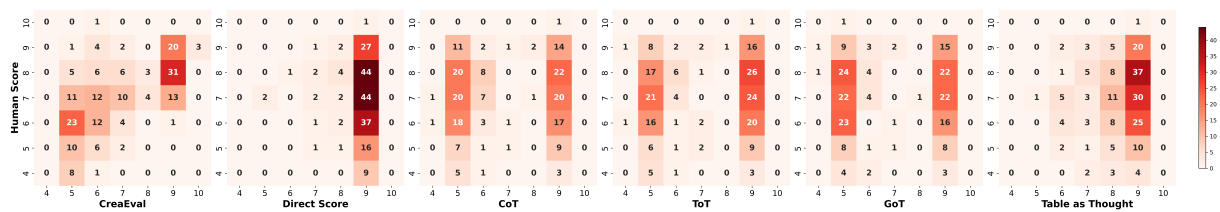} 
    \caption{Headmap between human score and all methods on \textit{Focus} of Step-2.}
    \label{Figure:Headmap on Focus of Step-2}
\end{figure*}

\begin{figure*}[!t]
    \centering
    \includegraphics[width=\textwidth]{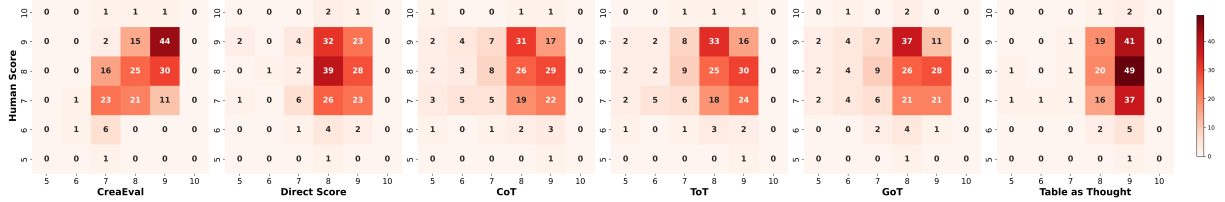} 
    \caption{Headmap between human score and all methods on \textit{Adequacy} of Step-2.}
    \label{Figure:Headmap on Adequacy of Step-2}
\end{figure*}

\begin{figure*}[!t]
    \centering
    \includegraphics[width=\textwidth]{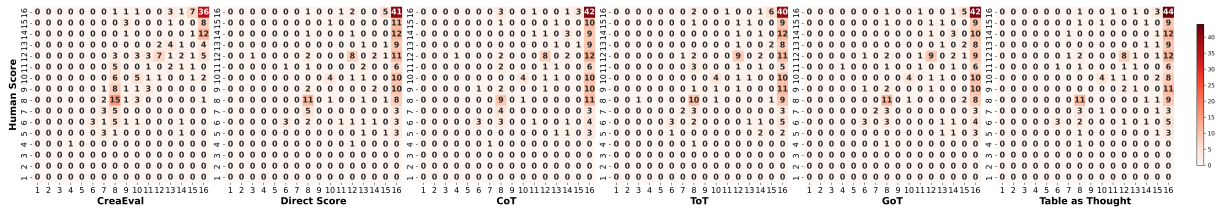} 
    \caption{Headmap between human score and all methods on \textit{Elaboration} of Step-3.}
    \label{Figure:Headmap on Elaboration of Step-3}
\end{figure*}

\begin{figure*}[!t]
    \centering
    \includegraphics[width=\textwidth]{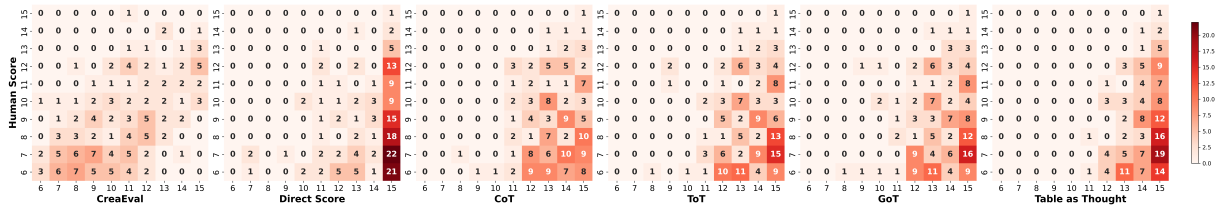} 
    \caption{Headmap between human score and all methods on \textit{Relevance} of Step-4.}
    \label{Figure:Headmap on Relevance of Step-4}
\end{figure*}

\begin{figure*}[!t]
    \centering
    \includegraphics[width=\textwidth]{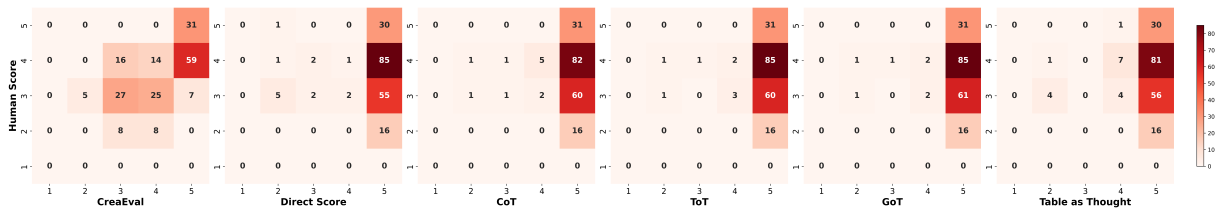} 
    \caption{Headmap between human score and all methods on \textit{Relevance} of Step-6.}
    \label{Figure:Headmap on Relevance of Step-6}
\end{figure*}

\begin{figure*}[!t]
    \centering
    \includegraphics[width=\textwidth]{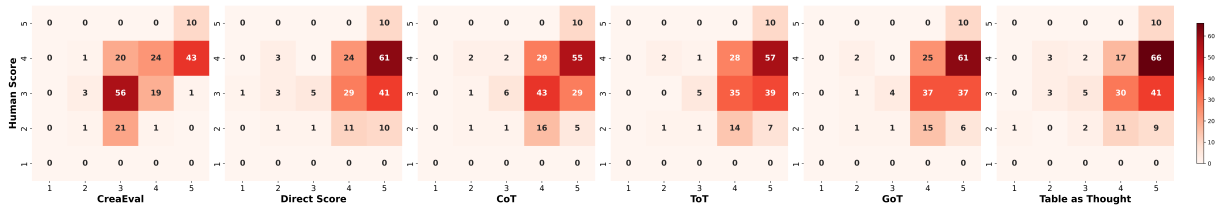} 
    \caption{Headmap between human score and all methods on \textit{Effectiveness} of Step-6.}
    \label{Figure:Headmap on Effectiveness of Step-6}
\end{figure*}

\begin{figure*}[!t]
    \centering
    \includegraphics[width=\textwidth]{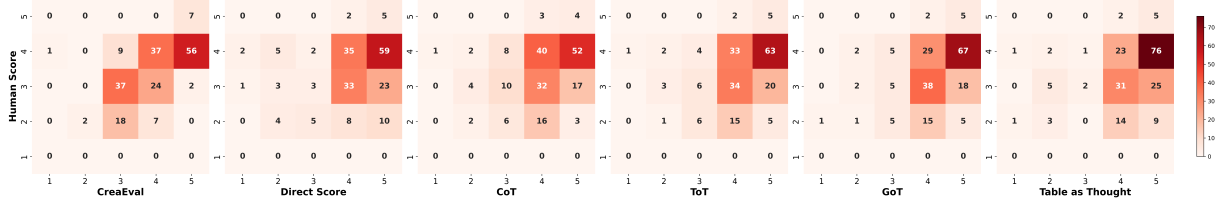} 
    \caption{Headmap between human score and all methods on \textit{Criteria} of Step-6.}
    \label{Figure:Headmap on Criteria of Step-6}
\end{figure*}

\begin{figure*}[!t]
    \centering
    \includegraphics[width=\textwidth]{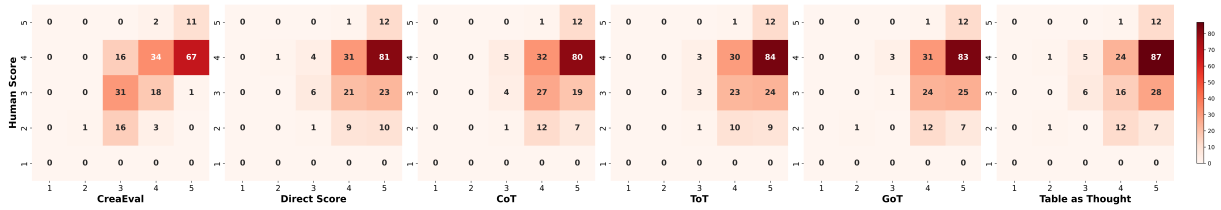} 
    \caption{Headmap between human score and all methods on \textit{Impact} of Step-6.}
    \label{Figure:Headmap on Impact of Step-6}
\end{figure*}

\begin{figure*}[!t]
    \centering
    \includegraphics[width=\textwidth]{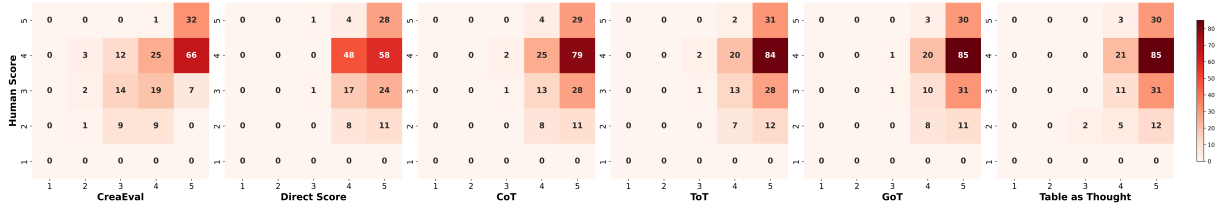} 
    \caption{Headmap between human score and all methods on \textit{Humaneness} of Step-6.}
    \label{Figure:Headmap on Humaneness of Step-6}
\end{figure*}

\begin{figure*}[!t]
    \centering
    \includegraphics[width=\textwidth]{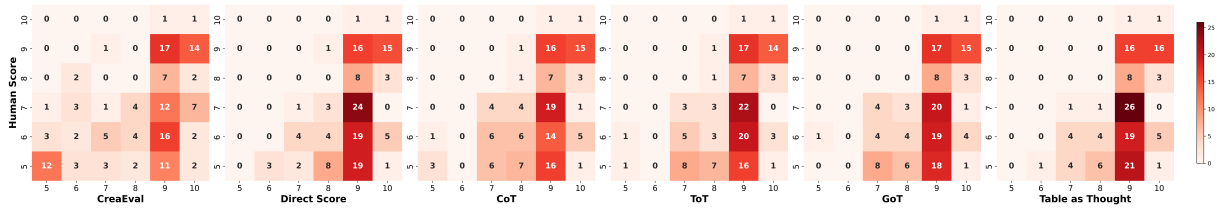} 
    \caption{Headmap between human score and all methods on \textit{Development} of Step-6.}
    \label{Figure:Headmap on Development of Step-6}
\end{figure*}

In Section~\ref{Section:Bias Mitigation in CreaEval}, we analyze the mitigation of leniency bias in LLM-as-a-judge by the heatmaps of human scores against all training-free methods on \textit{Originality} of Step-3. Figures~\ref{Figure:Headmap on Elaboration of Step-1} to~\ref{Figure:Headmap on Development of Step-6} present the heatmaps for several representative dimensions. Across these dimensions, CreaEval consistently alleviates leniency bias compared to other methods. This advantage is particularly evident in Step-6, the most comprehensive step, which requires holistic integration of all previous steps to generate a final action plan. In all dimensions of Step-6 (Figure~\ref{Figure:Headmap on Relevance of Step-6} to ~\ref{Figure:Headmap on Development of Step-6}), CreaEval facilitates a more balanced score distribution, which further demonstrates that the decoupled analysis-and-judging design achieves superior alignment with human annotations.

\section{Violin Plots of Inter-Judge Variance across Dimensions}
\label{App:Violin Plots of Inter-Judge Variance across Dimensions}
\begin{figure*}[!t]
  \includegraphics[width=0.48\linewidth]{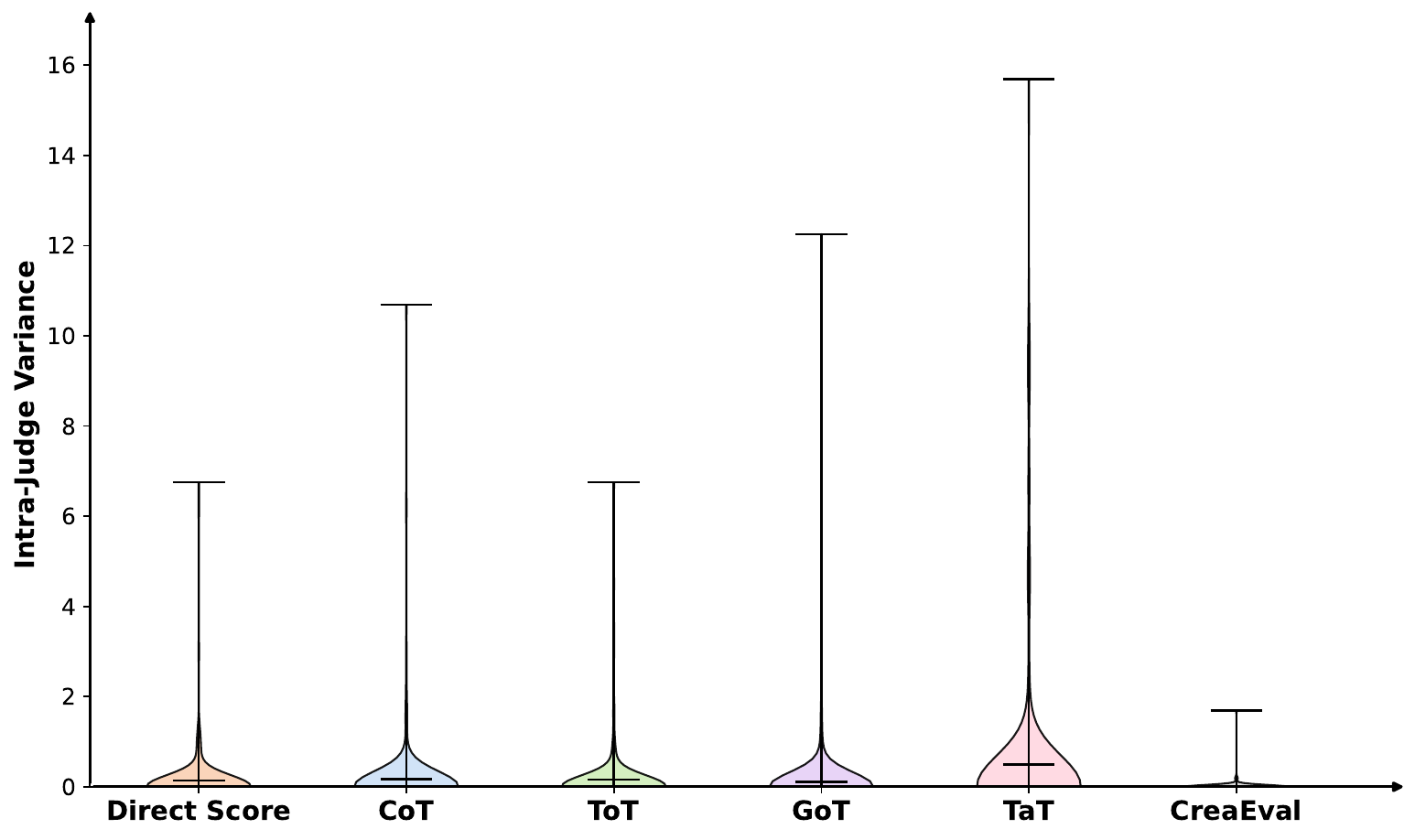} \hfill
  \includegraphics[width=0.48\linewidth]{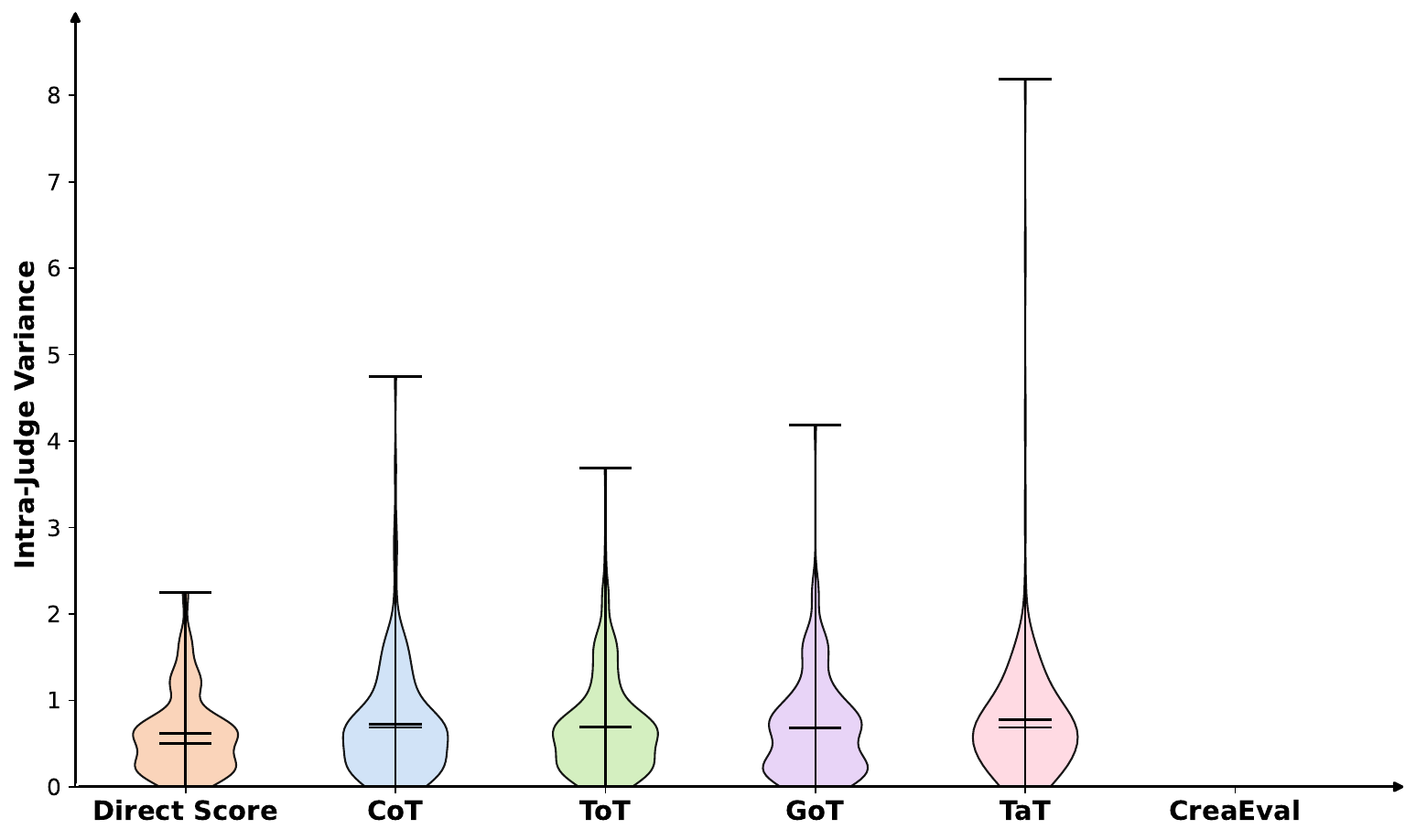} 
  \caption {Distributions of inter-Judge variance for \textit{Fluency} (left) and \textit{Flexibility} (right) in Step-1.}
  \label{Figure:Distributions for Fluency and Flexibility in Step-1}
\end{figure*}

\begin{figure*}[!t]
  \includegraphics[width=0.48\linewidth]{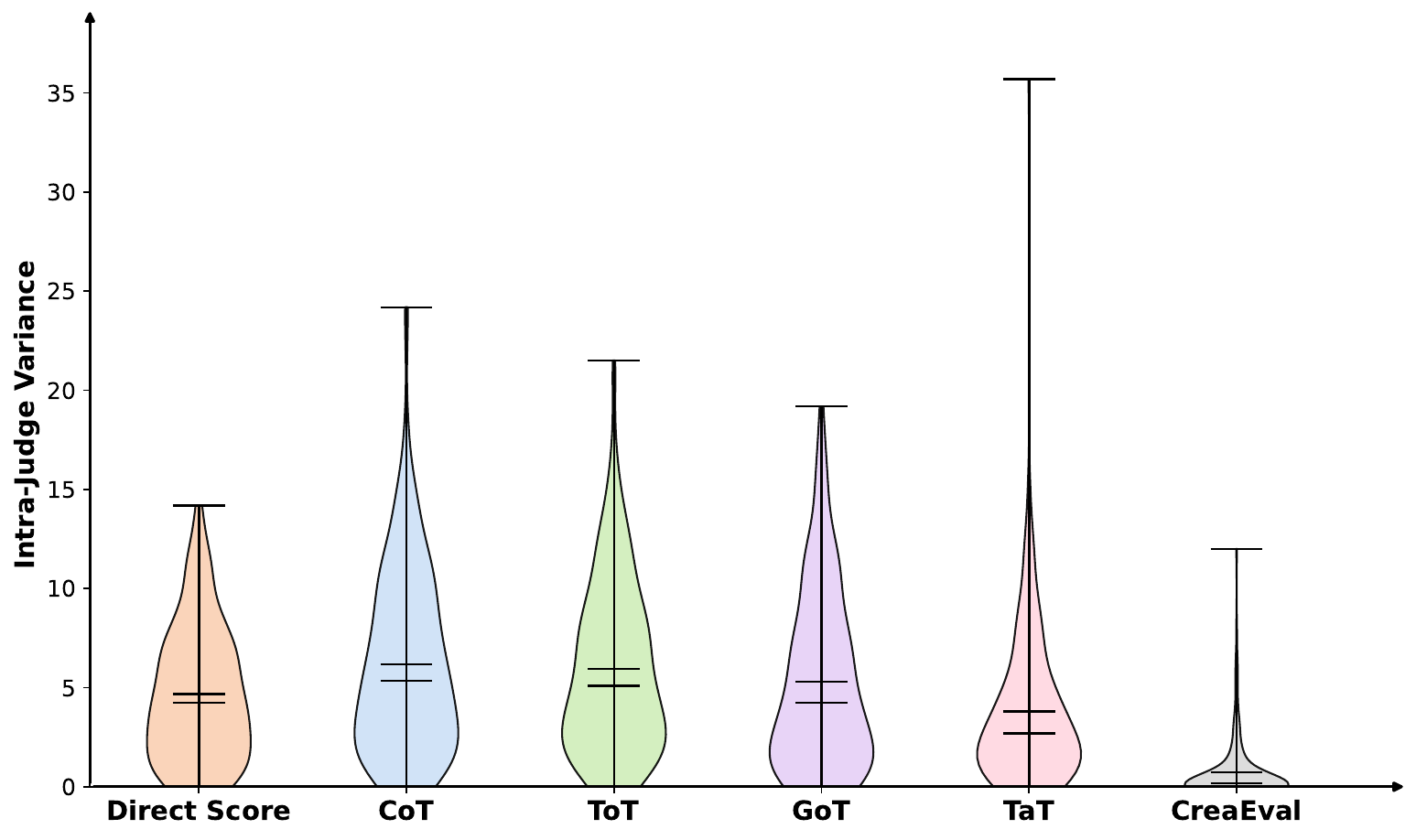} \hfill
  \includegraphics[width=0.48\linewidth]{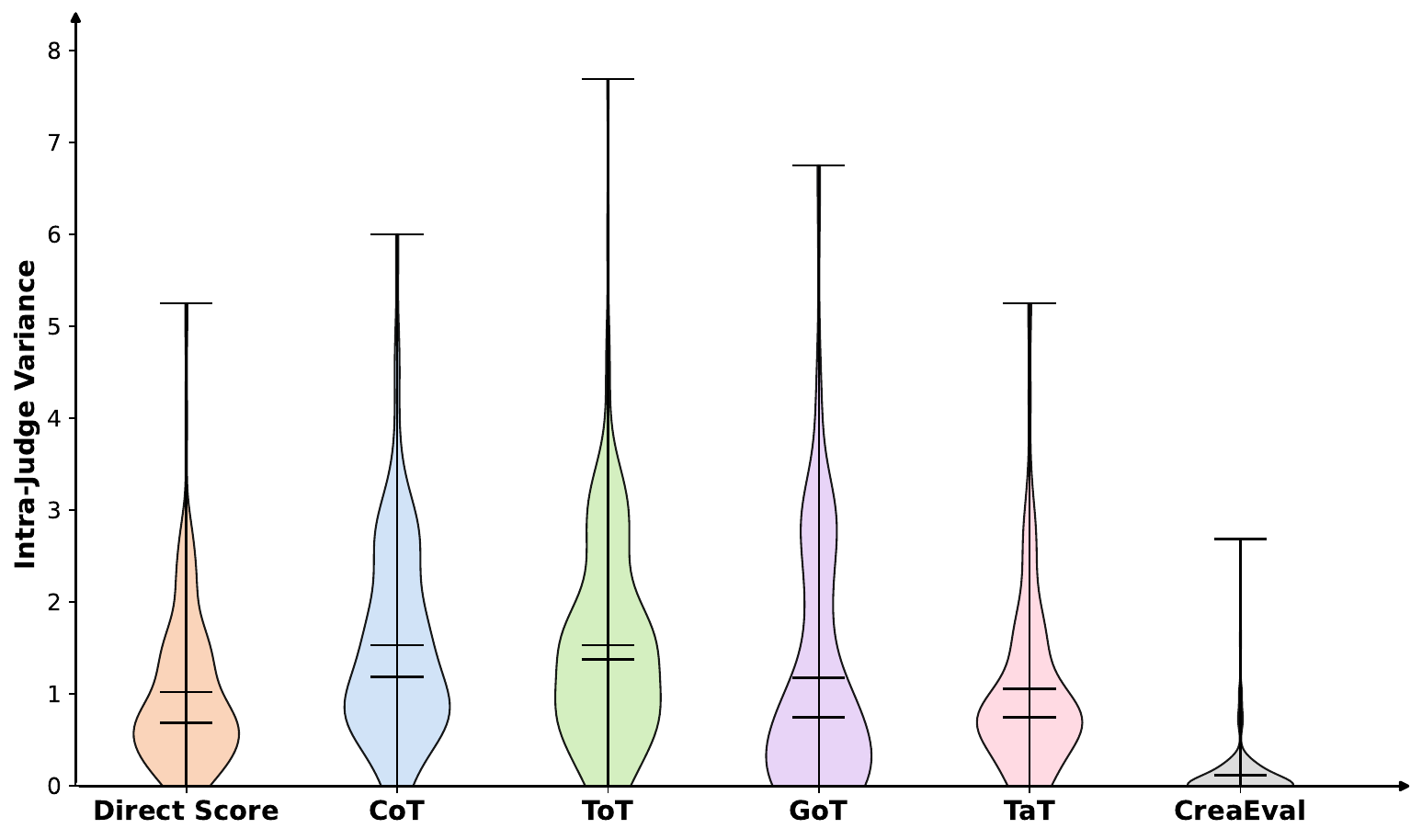} 
  \caption {Distributions of inter-Judge variance for \textit{Originality} in Step-1 (left) and \textit{Integrity} in Step-2 (right).}
  \label{Figure:Distributions for Originality in Step-1 and Integrity in Step-2}
\end{figure*}

\begin{figure*}[!t]
  \includegraphics[width=0.48\linewidth]{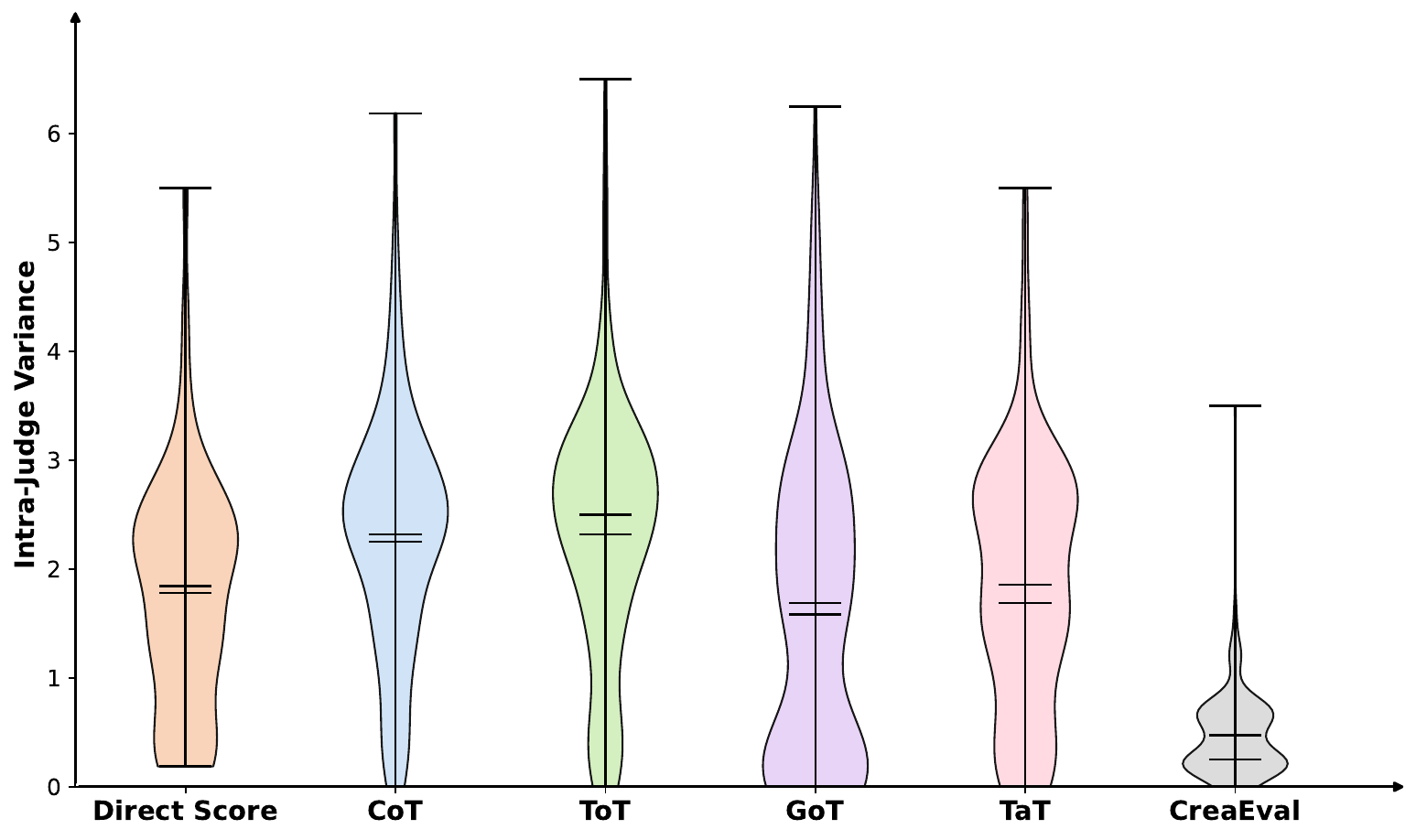} \hfill
  \includegraphics[width=0.48\linewidth]{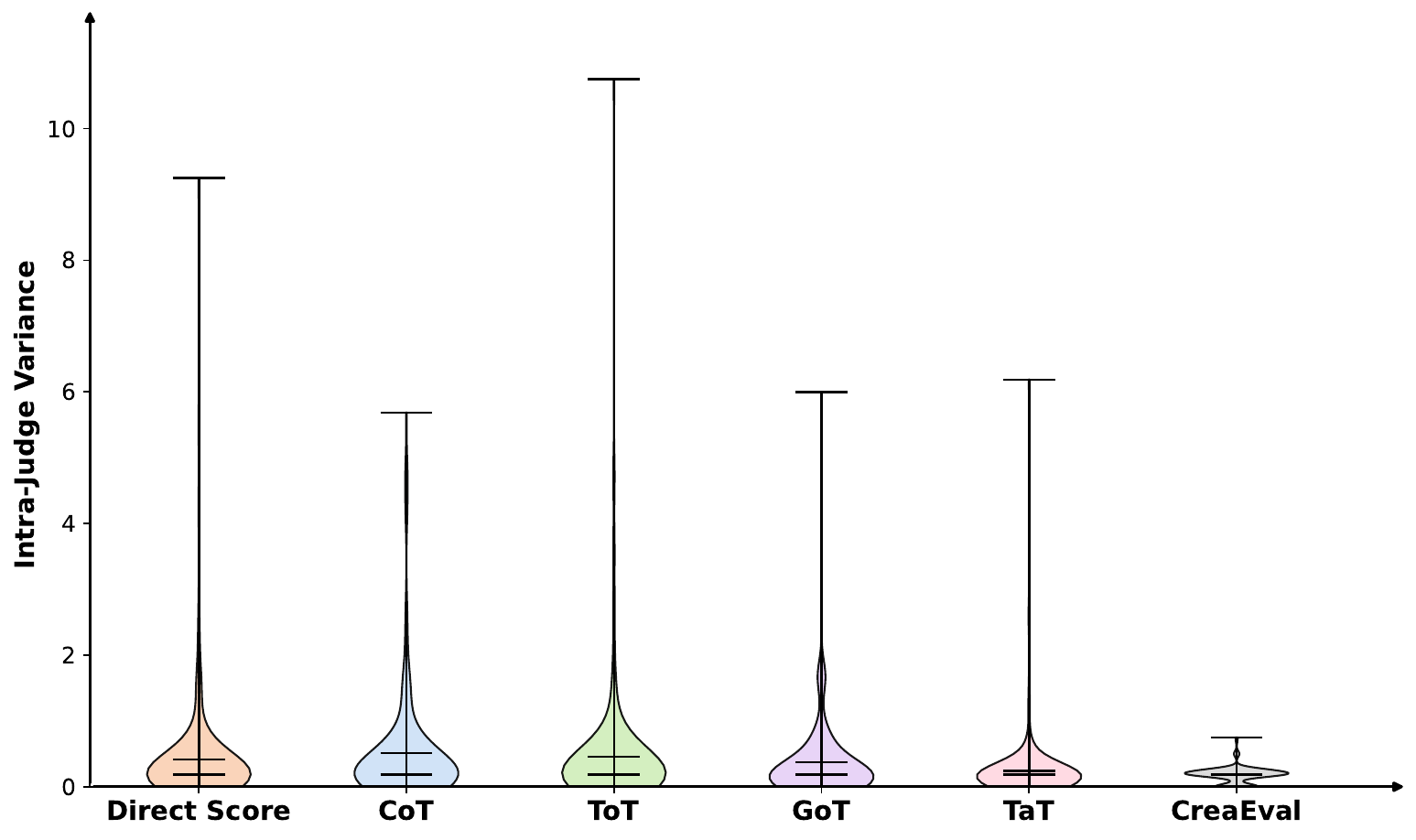} 
  \caption {Distributions of inter-Judge variance for \textit{Focus} (left) and \textit{Adequacy} (right) in Step-2.}
  \label{Figure:Distributions for Focus and Adequacy in Step-2}
\end{figure*}

\begin{figure*}[!t]
  \includegraphics[width=0.48\linewidth]{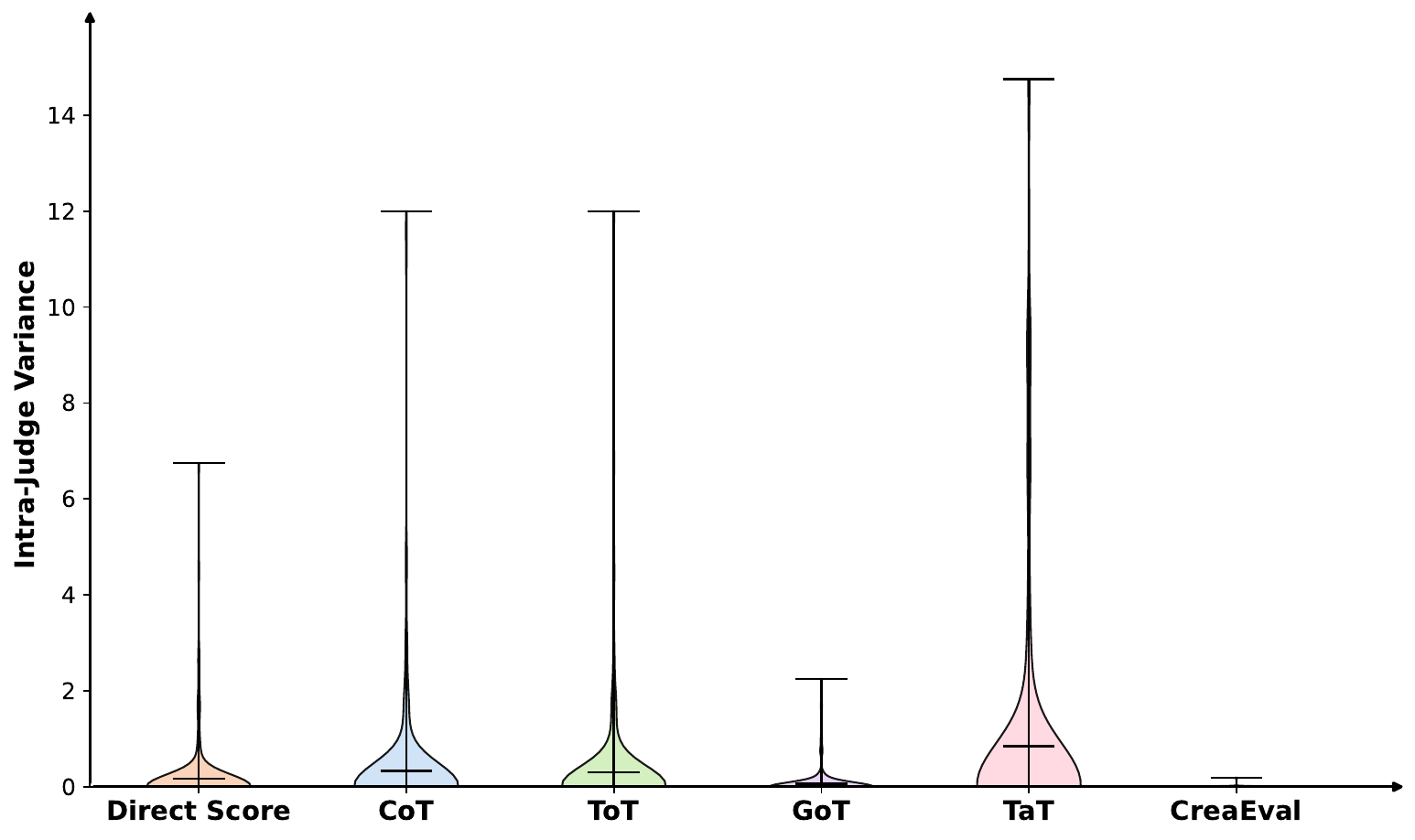} \hfill
  \includegraphics[width=0.48\linewidth]{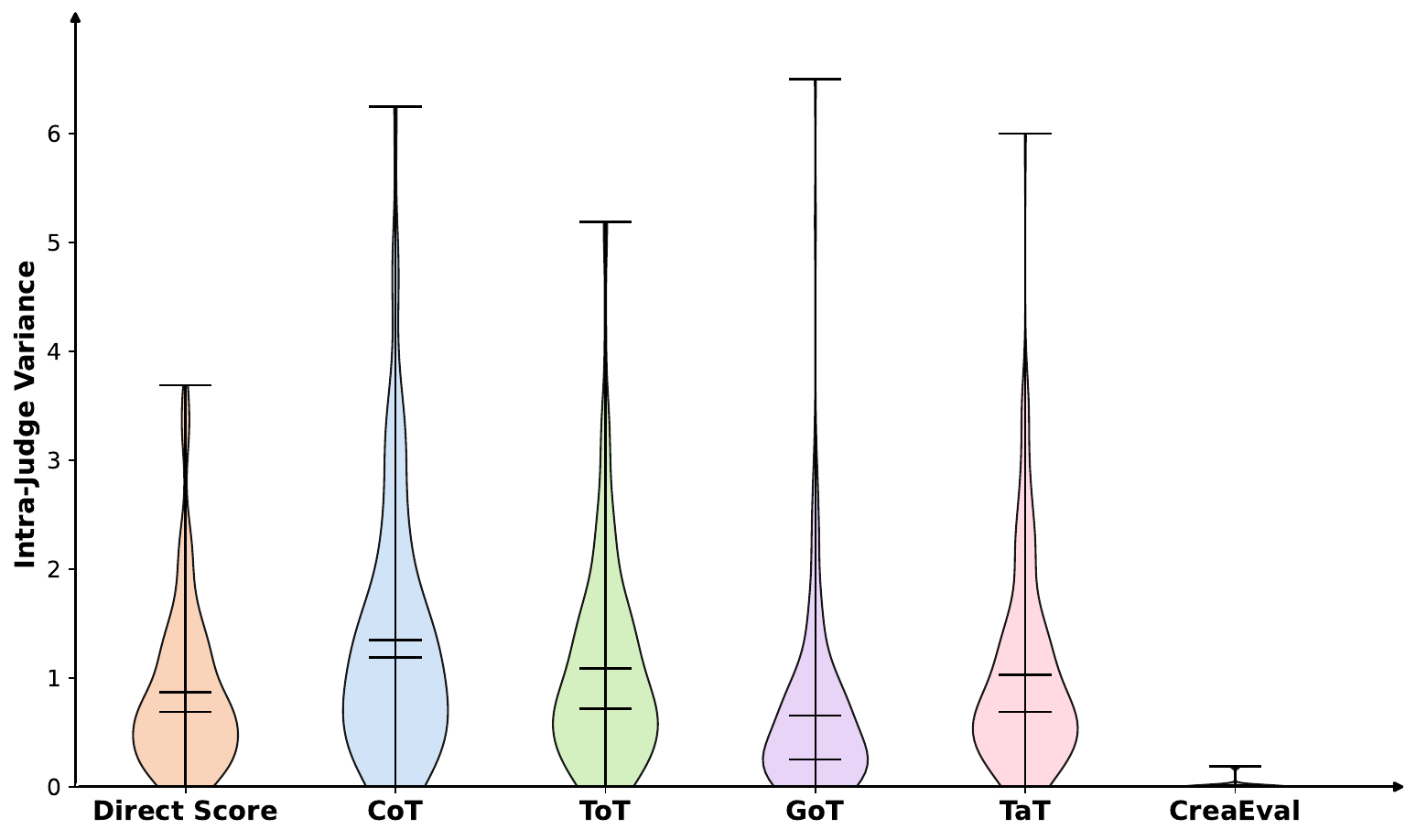} 
  \caption {Distributions of inter-Judge variance for \textit{Fluency} (left) and \textit{Flexibility} (right) in Step-3.}
  \label{Figure:Distributions for Fluency and Flexibility in Step-3}
\end{figure*}

\begin{figure*}[!t]
  \includegraphics[width=0.48\linewidth]{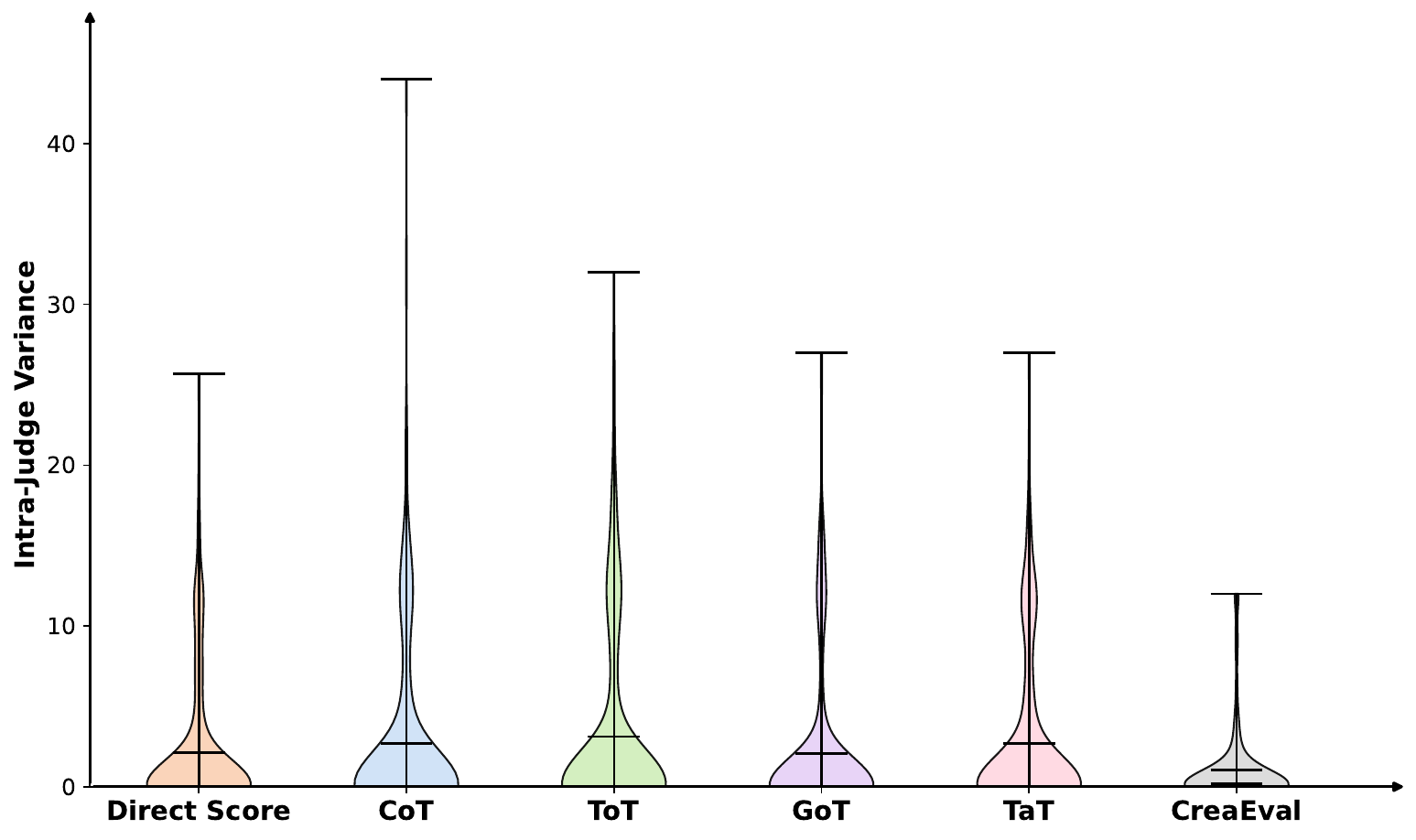} \hfill
  \includegraphics[width=0.48\linewidth]{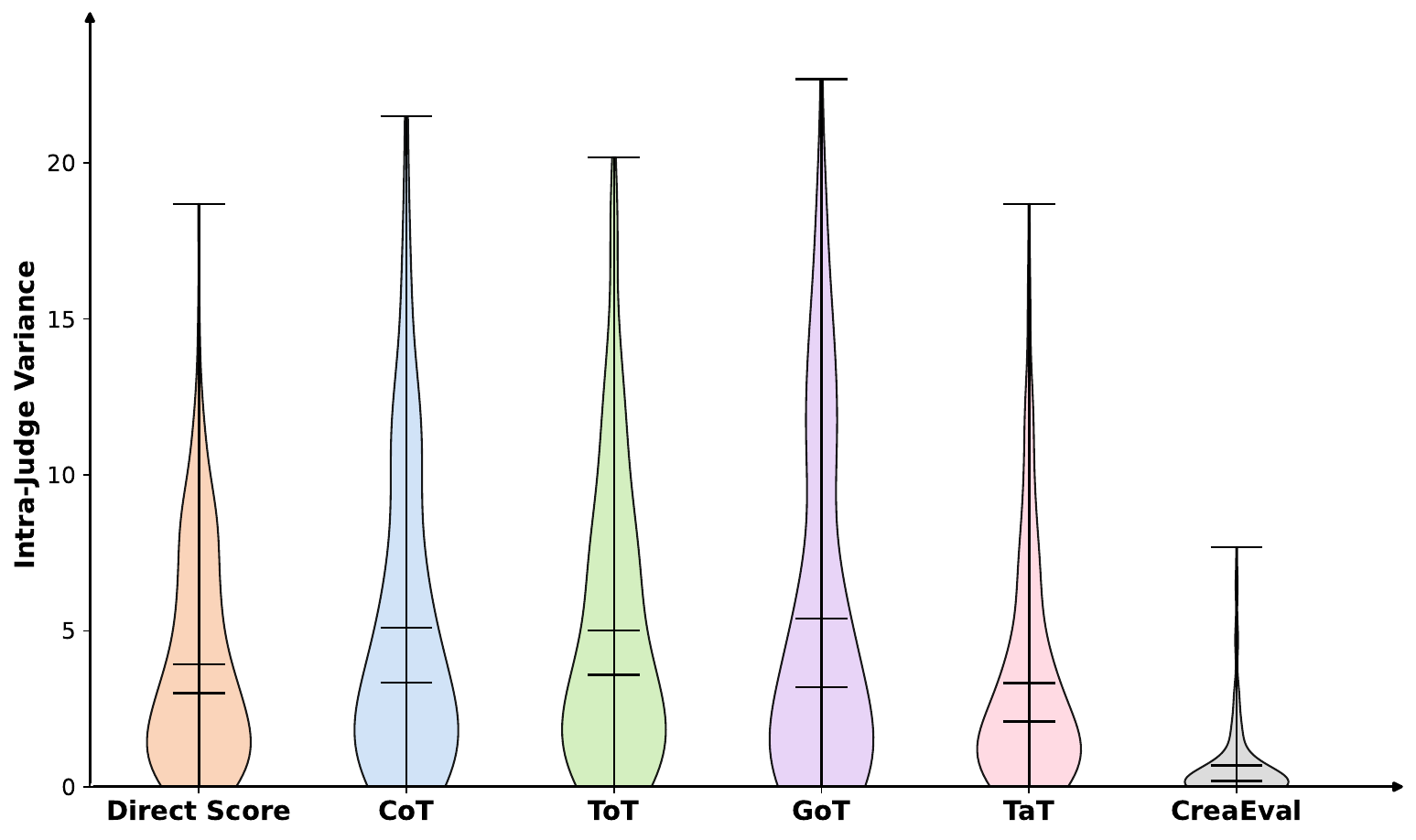} 
  \caption {Distributions of inter-Judge variance for \textit{Elaboration} (left) and \textit{Originality} (right) in Step-3.}
  \label{Figure:Distributions for Elaboration and Originality in Step-3}
\end{figure*}

\begin{figure*}[!t]
  \includegraphics[width=0.48\linewidth]{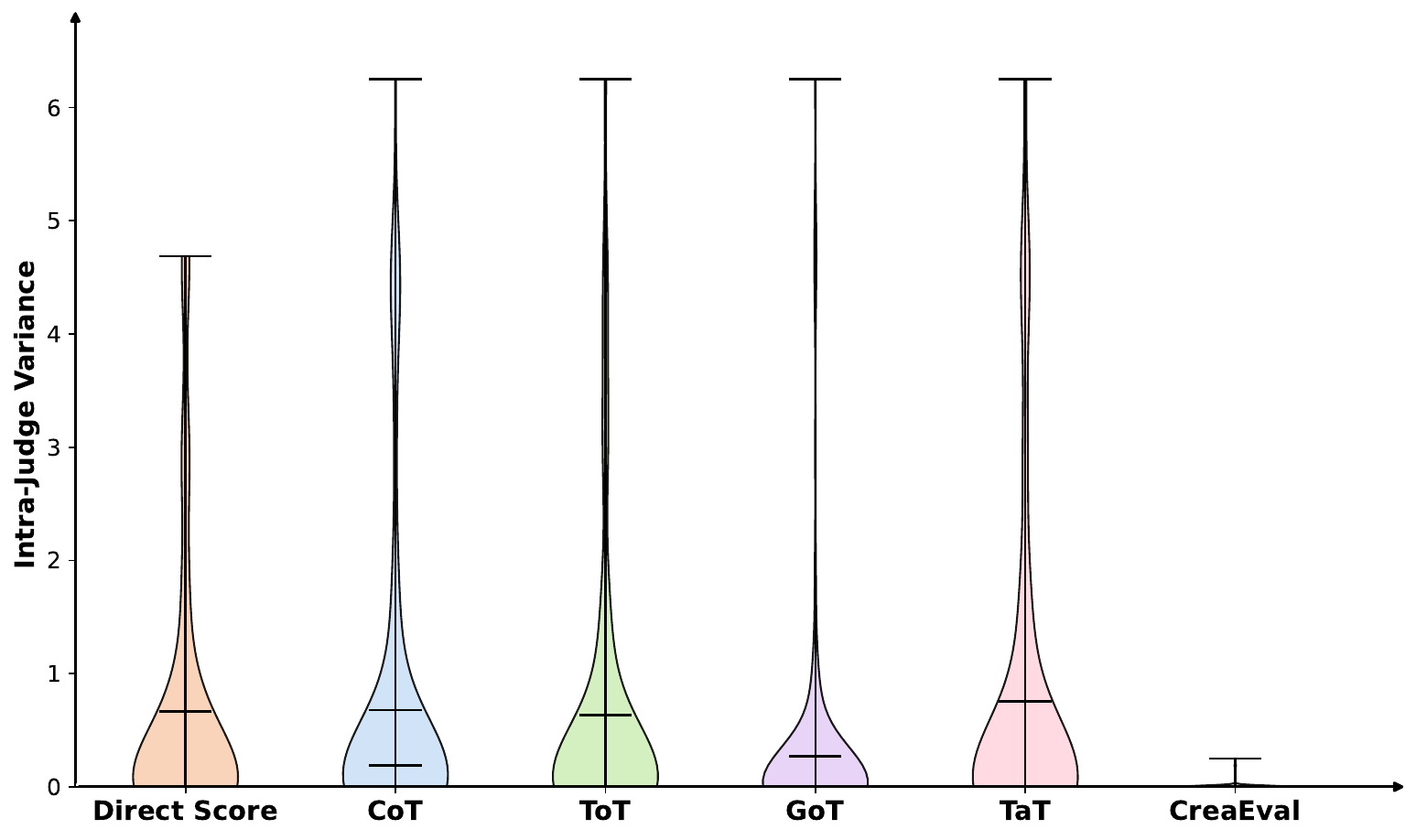} \hfill
  \includegraphics[width=0.48\linewidth]{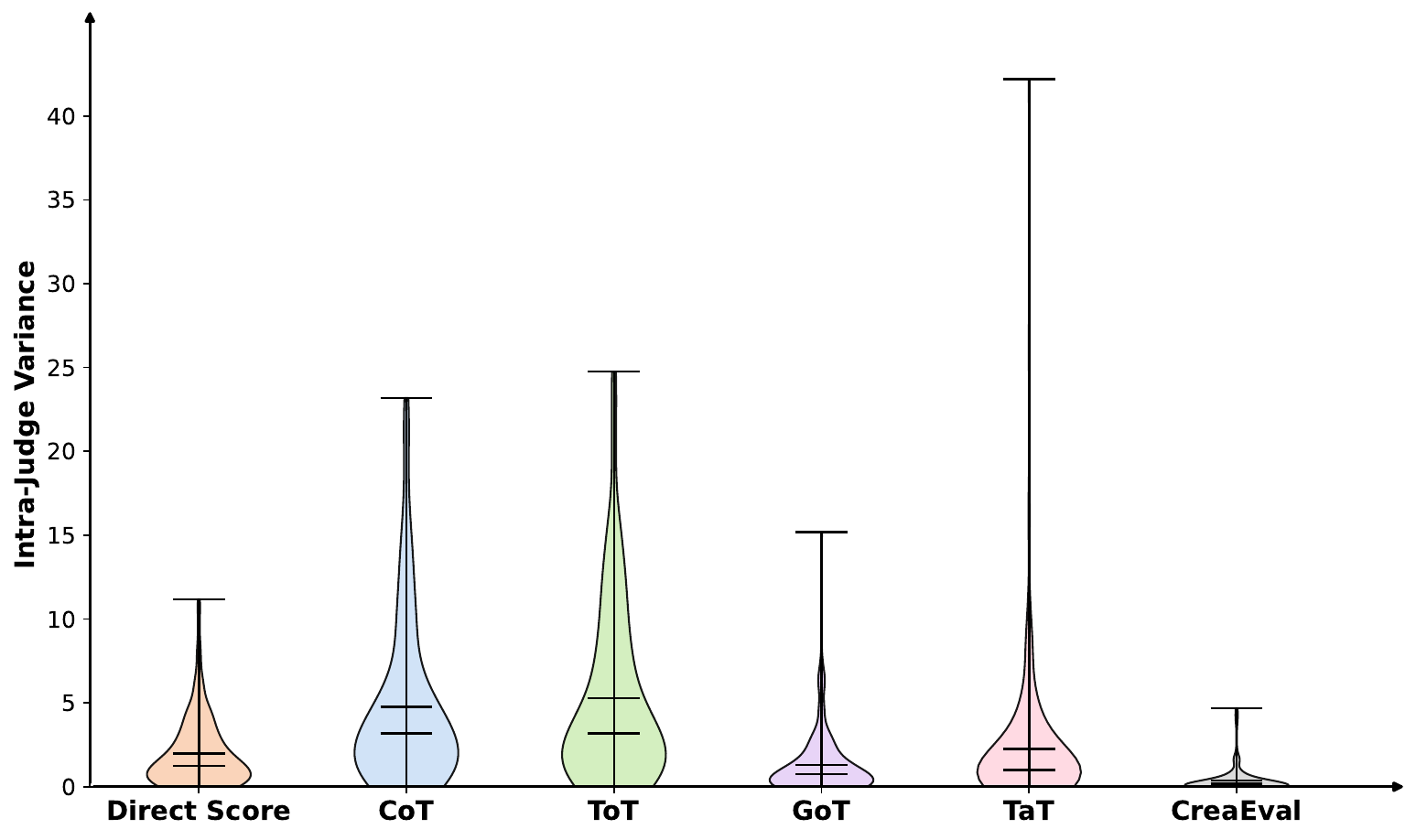} 
  \caption {Distributions of inter-Judge variance for \textit{Correctly Written} (left) and \textit{Relevance} (right) in Step-4.}
  \label{Figure:Distributions for Correctly Written and Relevance in Step-4}
\end{figure*}

\begin{figure*}[!t]
  \includegraphics[width=0.48\linewidth]{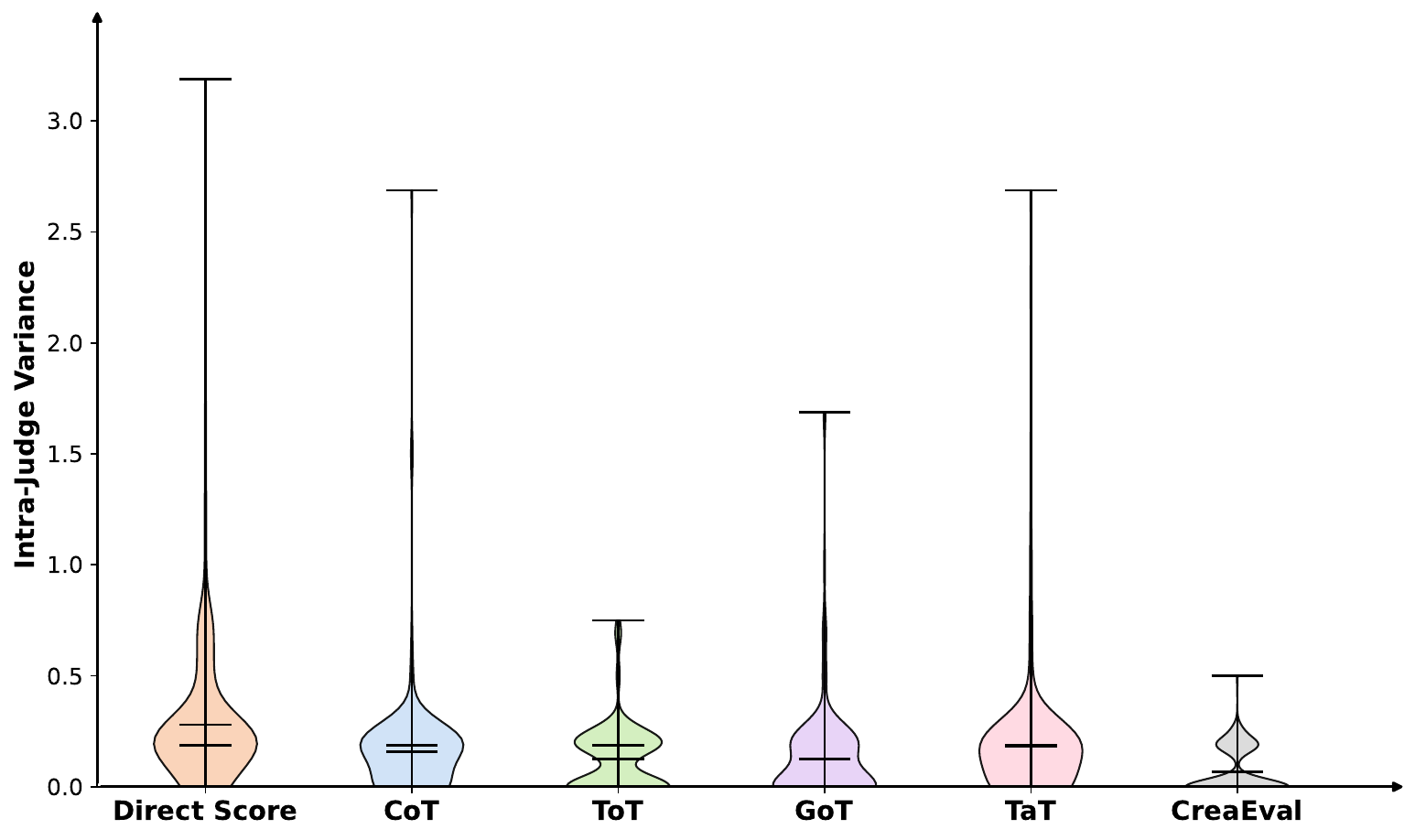} \hfill
  \includegraphics[width=0.48\linewidth]{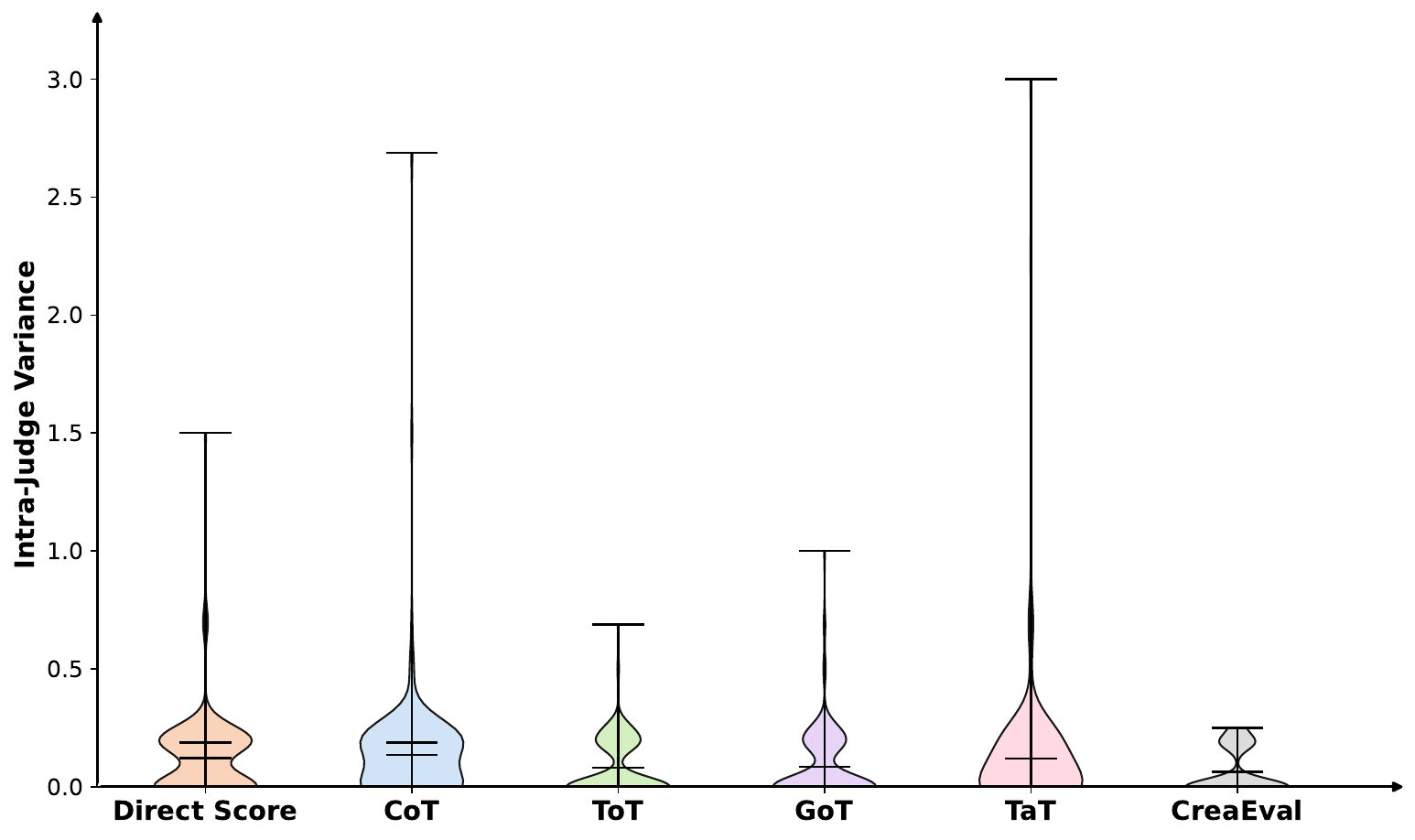} 
  \caption {Distributions of inter-Judge variance for \textit{Effectiveness} (left) and \textit{Impact} (right) in Step-6.}
  \label{Figure:Distributions for Effectiveness and Impact in Step-6}
\end{figure*}

In Section~\ref{Section:Scoring Stability Analysis}, we analyze scoring stability by measuring the distribution of inter-Judge variance on \textit{Correctly Used} of Step-5 for each method. Figures~\ref{Figure:Distributions for Fluency and Flexibility in Step-1} to~\ref{Figure:Distributions for Effectiveness and Impact in Step-6} present violin plots for additional representative dimensions. Across these dimensions, CreaEval consistently exhibits lower inter-Judge variance, indicating that the four Judge-LLMs within CreaEval produce more similar scores. This improvement can be attributed to the two-phase design that decouples analysis from judging, enabling evidence-grounded scoring that constrains the plausible scoring range and thereby enhances scoring stability.

Interestingly, on \textit{Flexibility} of Step-1, CreaEval shows no variance (the violin plot is empty at the corresponding position). This is because the evidence already provides the category for each challenge, so each Judge-LLM only needs to count the number of different categories under this dimension. As a result, all four Judge-LLMs produced identical scores.

\section{Case Studies for Remaining Steps}
\label{App:Case Studies for Remaining Steps}
\begin{figure*}
    \centering
    \includegraphics[width=\textwidth]{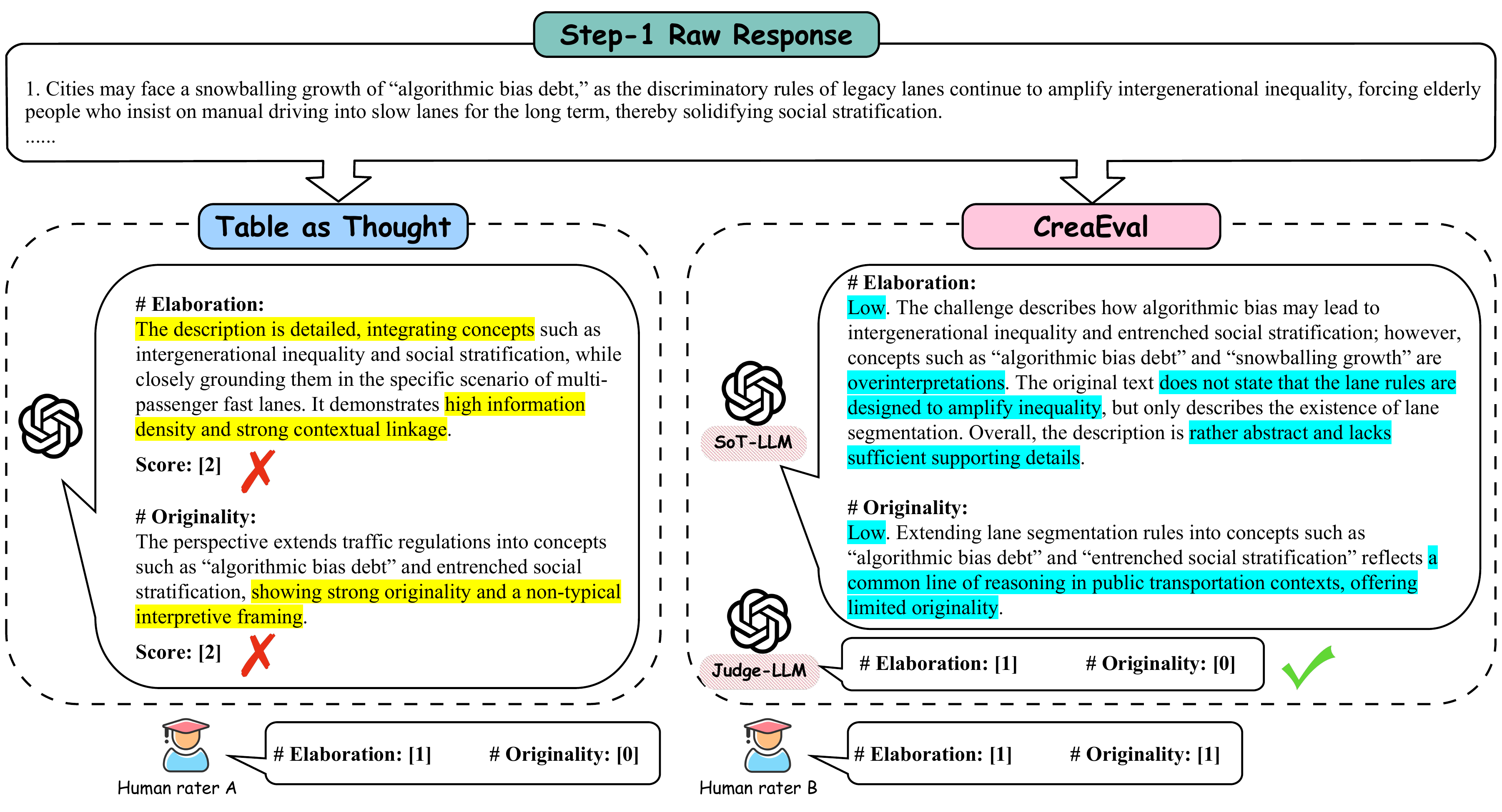} 
    \caption{Case study of Step-1 comparing Table as Thought (TaT) and CreaEval. TaT assigns high scores by emphasizing the richness and novelty of abstract concepts such as “algorithmic bias debt” and “social stratification” (yellow), while CreaEval provides a more grounded analysis by distinguishing contextual relevance from unsupported overinterpretation (blue). Consequently, CreaEval aligns more closely with human judges in identifying limited elaboration and only modest originality.}
    \label{Figure:Case Study of Step-1}
\end{figure*}

\begin{figure*}
    \centering
    \includegraphics[width=\textwidth]{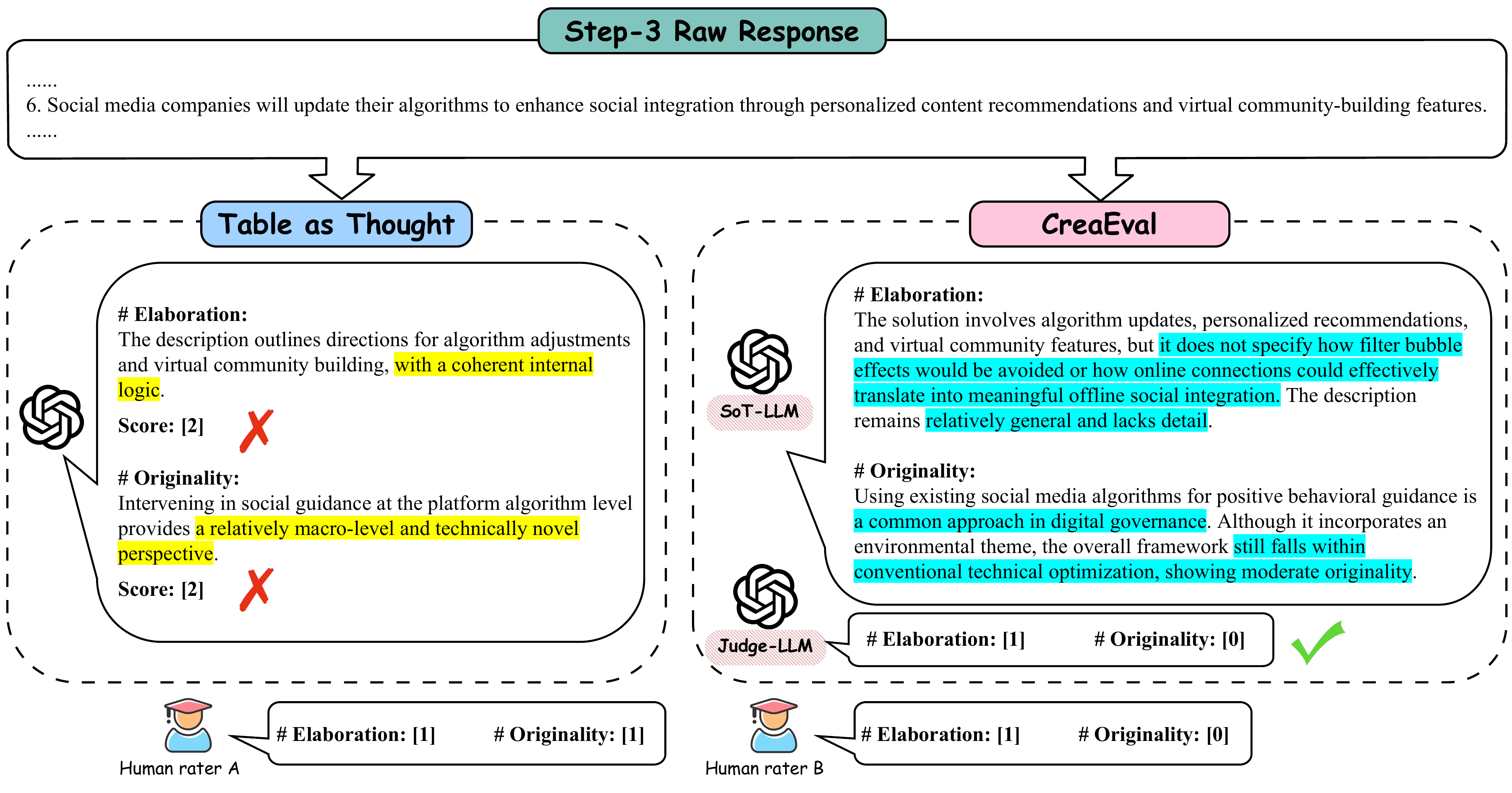} 
    \caption{Case study of Step-3 comparing Table as Thought (TaT) and CreaEval. TaT assigns high scores by emphasizing the coherence and technical framing of algorithm-driven social guidance (yellow), whereas CreaEval further examines the lack of concrete mechanisms, such as preventing filter bubbles and ensuring the transition from online interaction to meaningful offline social integration (blue). As a result, CreaEval produces more conservative scores that are better aligned with human judgments on both elaboration and originality.}
    \label{Figure:Case Study of Step-3}
\end{figure*}

\begin{figure*}
    \centering
    \includegraphics[width=\textwidth]{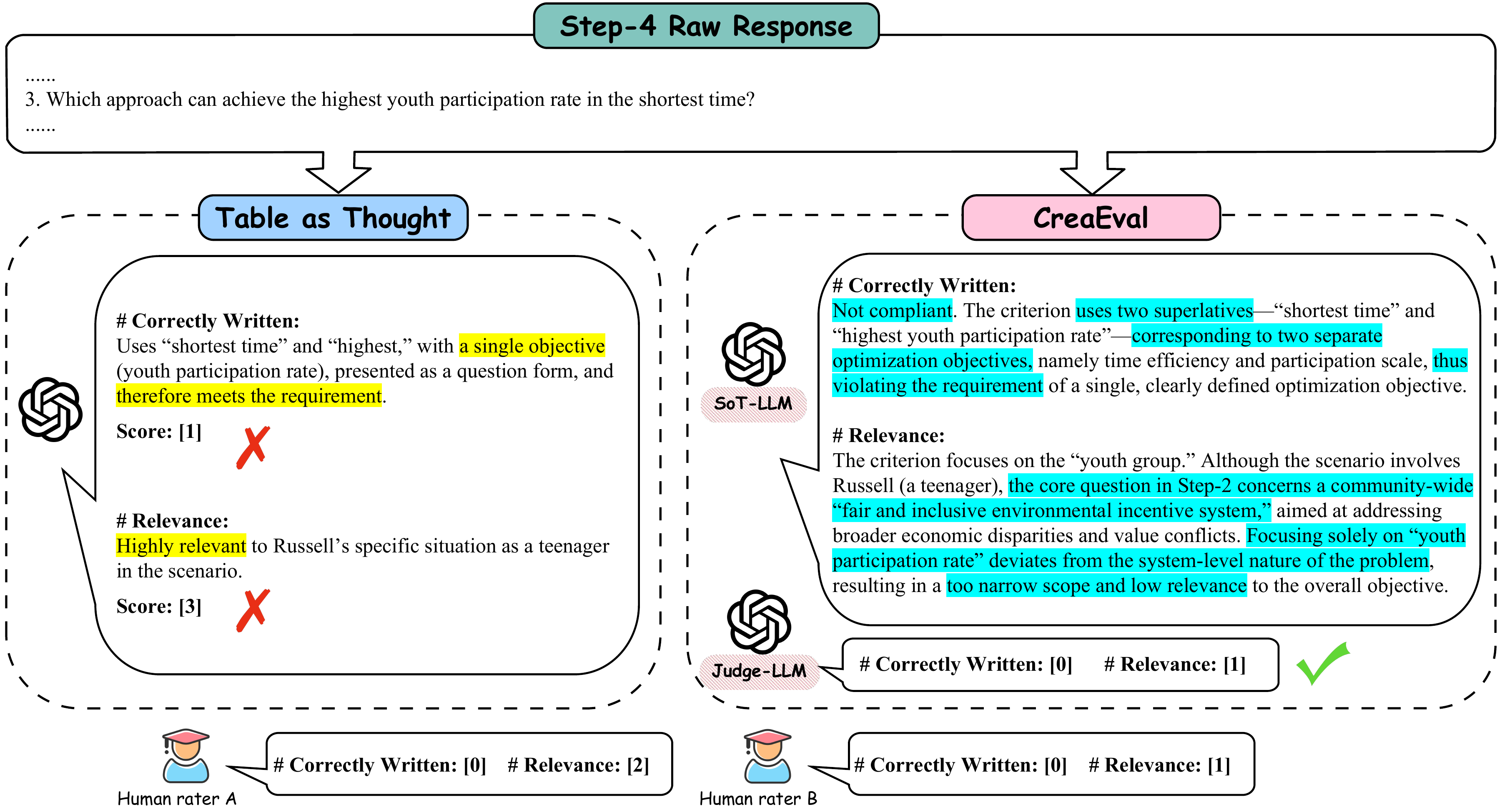} 
    \caption{Case study of Step-4 comparing Table as Thought (TaT) and CreaEval. TaT assigns high scores by focusing on youth participation as a relevant target (yellow), whereas CreaEval identifies that the criterion contains multiple optimization objectives (“shortest time” and “highest participation rate”) and is overly narrow compared with the community-wide fairness goal of the task (blue). As a result, CreaEval aligns more closely with human judgments.}
    \label{Figure:Case Study of Step-4}
\end{figure*}

\begin{figure*}
    \centering
    \includegraphics[width=\textwidth]{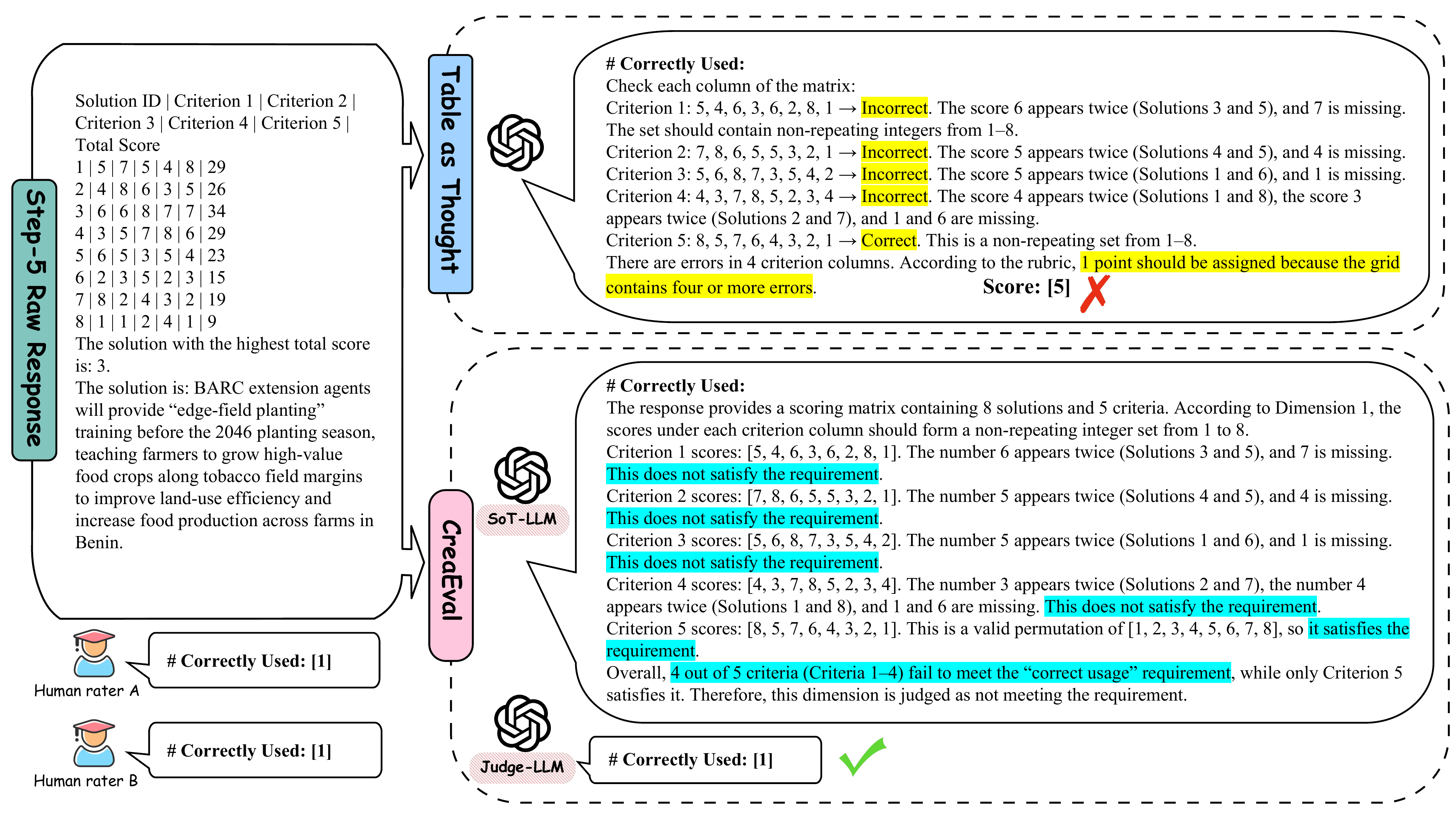} 
    \caption{Case study of Step-5 comparing Table as Thought (TaT) and CreaEval. Both methods identify repeated and missing scores in multiple criterion columns of the ranking matrix. However, while TaT incorrectly assigns a high score despite detecting four column errors, CreaEval consistently maps the detected violations to the rubric requirement, producing a judgment aligned with human evaluators.}
    \label{Figure:Case Study of Step-5}
\end{figure*}

\begin{figure*}
    \centering
    \includegraphics[width=\textwidth]{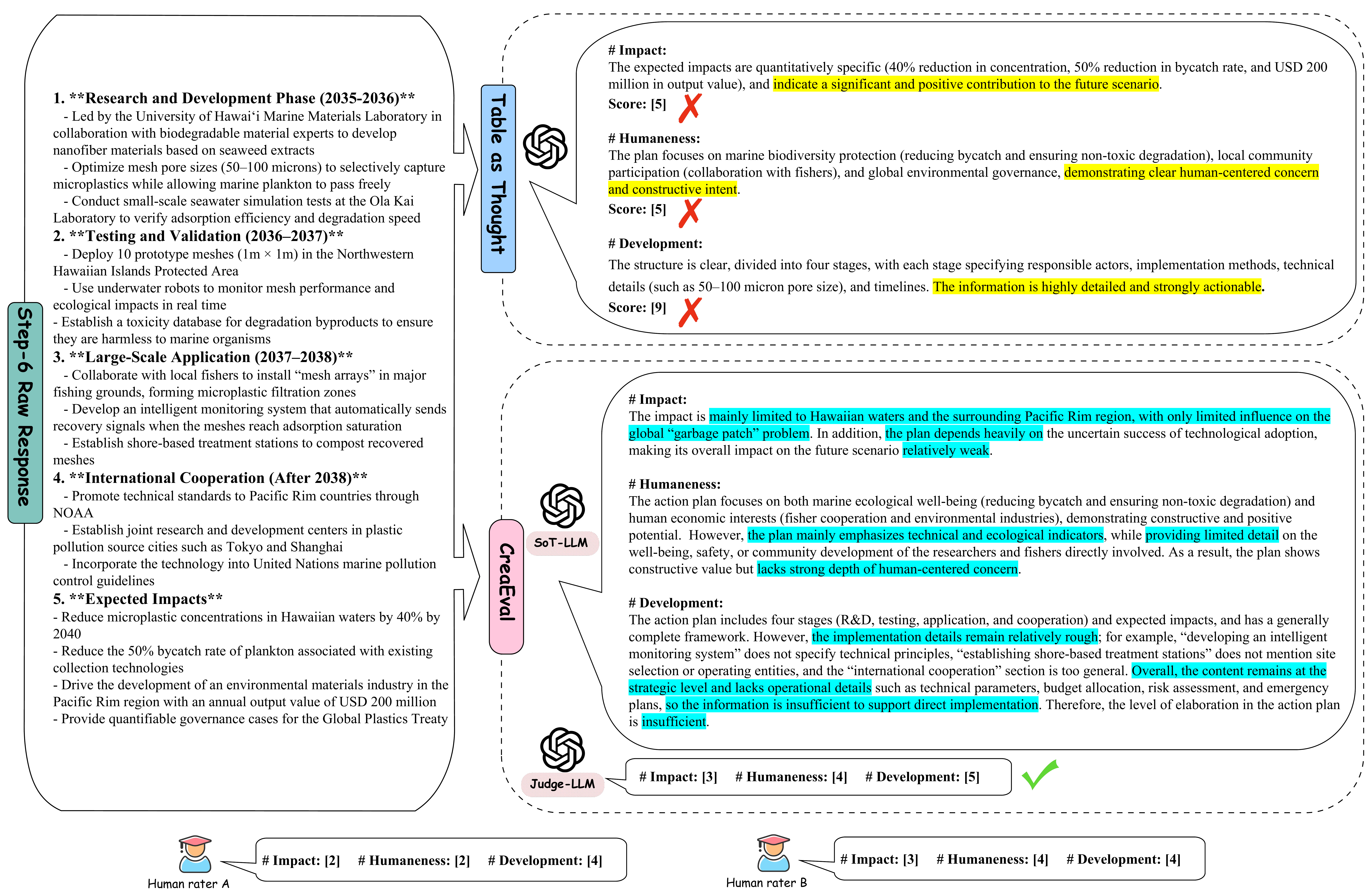} 
    \caption{Case study of Step-6 comparing Table as Thought (TaT) and CreaEval. TaT assigns high scores by emphasizing the detailed structure, quantified impacts, and constructive environmental vision of the action plan (yellow), whereas CreaEval further examines the limited global scalability, insufficient human-centered details, and lack of operational implementation specifics (blue). Consequently, CreaEval produces more conservative evaluations that align more closely with human judgments across the three dimensions.}
    \label{Figure:Case Study of Step-6}
\end{figure*}

In Section~\ref{Section:Case Study}, we present a case study for Step-2. Figures~\ref{Figure:Case Study of Step-1} to~\ref{Figure:Case Study of Step-6} provide case studies for the remaining five steps, further illustrating how CreaEval improves the accuracy of subjective dimensions and mitigates biases through its decoupled analysis-and-judging design.

Notably, in Step-5, although both TaT and CreaEval correctly identified the analysis results (i.e., four errors in the scoring matrix), TaT still assigned an incorrect score of 5, while CreaEval produced the correct score of 1. This shows that, in a coupled analysis-and-judging setting, even when the analysis is correct, mixing analysis and scoring can still affect the final result, thereby reducing the accuracy of the final score.

\begin{table*}[!t]
\centering
\small
\setlength{\tabcolsep}{4pt}
\renewcommand{\arraystretch}{1.15}
\begin{tabular}{l |c |cccc ccc c}
\toprule
\multirow{2}{*}{Model} & \multirow{2}{*}{Temperature} & \multicolumn{4}{c}{Step-1} & \multicolumn{3}{c}{Step-2} & \multirow{2}{*}{AVG} \\
\cmidrule(lr){3-6} \cmidrule(lr){7-9}
& & Fluency & Flexibility & Elaboration & Originality & Integrity & Focus & Adequacy & \\
\midrule

\multirow{3}{*}{qwen3.6-plus}
& 0.2 & \textbf{0.9074} & \textbf{0.982} & \textbf{0.9173} & \textbf{0.9086} & \textbf{0.9754} & \textbf{0.9437} & \textbf{0.9383} & \textbf{0.939} \\
& 0.5 & 0.896 & 0.9752 & 0.9139 & 0.9069 & 0.9684 & 0.9397 & 0.9248 & 0.9321 \\
& 0.8 & 0.8799 & 0.9714 & 0.9109 & 0.9046 & 0.9572 & 0.9298 & 0.9228 & 0.9252 \\
\midrule

\multirow{3}{*}{deepseek-v4-pro}
& 0.2 & \textbf{0.9691} & \textbf{0.9828} & \textbf{0.937} & \textbf{0.8926} & \textbf{0.9796} & \textbf{0.9421} & \textbf{0.9414} & \textbf{0.9492} \\
& 0.5 & 0.9456 & 0.9742 & 0.9145 & 0.8871 & 0.9654 & 0.9262 & 0.9323 & 0.935 \\
& 0.8 & 0.8849 & 0.9665 & 0.8999 & 0.8609 & 0.9548 & 0.9145 & 0.9233 & 0.915 \\
\midrule

\multirow{3}{*}{gemini-3.1-pro}
& 0.2 & 0.8707 & 0.9792 & \textbf{0.9326} & \textbf{0.9284} & \textbf{0.9732} & \textbf{0.9701} & \textbf{0.9363} & \textbf{0.9415} \\
& 0.5 & 0.9109 & \textbf{0.9813} & 0.9197 & 0.913 & 0.9708 & 0.9588 & 0.9304 & 0.9407 \\
& 0.8 & \textbf{0.9124} & 0.9782 & 0.9209 & 0.9101 & 0.9632 & 0.9403 & 0.9214 & 0.9352 \\
\midrule

\multirow{3}{*}{gpt-5.4}
& 0.2 & \textbf{0.9466} & 0.983 & \textbf{0.9312} & 0.8953 & \textbf{0.9562} & \textbf{0.9338} & 0.9284 & \textbf{0.9392} \\
& 0.5 & 0.9403 & \textbf{0.9841} & 0.9294 & \textbf{0.9016} & 0.9502 & 0.93 & \textbf{0.9319} & 0.9382 \\
& 0.8 & 0.9133 & 0.9807 & 0.9228 & 0.8917 & 0.9477 & 0.932 & 0.9305 & 0.9312 \\
\bottomrule

\end{tabular}
\caption{BERTScore similarity results of repeated evidence extraction under different temperature settings during Memory-augmented Extraction phase. \textbf{Bold} indicates the highest score across all temperature settings for each LLM. Higher scores indicate greater semantic similarity and more stable evidence extraction across repeated runs.}
\label{Table:pre_experiment_bertscore}
\end{table*}

\begin{table*}[!t]
\centering
\small
\setlength{\tabcolsep}{4pt}
\renewcommand{\arraystretch}{1.15}
\begin{tabular}{l |c |cccc ccc c}
\toprule
\multirow{2}{*}{Model} & \multirow{2}{*}{Temperature} & \multicolumn{4}{c}{Step-1} & \multicolumn{3}{c}{Step-2} & \multirow{2}{*}{AVG} \\
\cmidrule(lr){3-6} \cmidrule(lr){7-9}
& & Fluency & Flexibility & Elaboration & Originality & Integrity & Focus & Adequacy & \\
\midrule

\multirow{3}{*}{qwen3.6-plus}
& 0.2 & \textbf{0.2424} & \textbf{0.6169} & 0.0728 & 0.0587 & \textbf{0.4134} & \textbf{0.1262} & \textbf{0.0757} & \textbf{0.2294} \\
& 0.5 & 0.1546 & 0.6064 & 0.0707 & 0.0645 & 0.3747 & 0.1135 & 0.0091 & 0.1991 \\
& 0.8 & 0.1278 & 0.5796 & \textbf{0.0796} & \textbf{0.075} & 0.2894 & 0.0505 & 0.0091 & 0.173 \\
\midrule

\multirow{3}{*}{deepseek-v4-pro}
& 0.2 & \textbf{0.5805} & \textbf{0.6053} & 0.0063 & 0.0252 & \textbf{0.4311} & \textbf{0.0631} & 0 & \textbf{0.2445} \\
& 0.5 & 0.5112 & 0.5966 & 0.0067 & 0.0302 & 0.2257 & 0 & 0 & 0.1958 \\
& 0.8 & 0.4469 & 0.5924 & \textbf{0.0075} & \textbf{0.0327} & 0.128 & 0 & 0 & 0.1725 \\
\midrule

\multirow{3}{*}{gemini-3.1-pro}
& 0.2 & \textbf{0.1598} & 0.6136 & 0.0256 & 0.0225 & \textbf{0.1854} & \textbf{0.2776} & 0 & \textbf{0.1835} \\
& 0.5 & 0.005 & \textbf{0.615} & \textbf{0.0313} & \textbf{0.0313} & 0.1603 & 0.1387 & \textbf{0.0145} & 0.1423 \\
& 0.8 & 0.0113 & 0.5814 & 0.0234 & 0.0234 & 0.0618 & 0 & 0 & 0.1002 \\
\midrule

\multirow{3}{*}{gpt-5.4}
& 0.2 & \textbf{0.3585} & \textbf{0.6619} & \textbf{0.1315} & \textbf{0.1997} & 0.1233 & 0.0724 & 0.0724 & \textbf{0.2314} \\
& 0.5 & 0.3095 & 0.6603 & 0.1203 & 0.1823 & \textbf{0.1261} & \textbf{0.087} & \textbf{0.0816} & 0.2239 \\
& 0.8 & 0.2745 & 0.6549 & 0.1186 & 0.1396 & 0.0626 & 0.0507 & 0.0468 & 0.1925 \\
\bottomrule

\end{tabular}
\caption{Self-BLEU similarity results of repeated evidence extraction under different temperature settings during Memory-augmented Extraction phase. \textbf{Bold} indicates the highest score across all temperature settings for each LLM. Higher scores indicate greater lexical similarity and more stable evidence extraction across repeated runs.}
\label{Table:pre_experiment_selfbleu}
\end{table*}

\begin{table*}[!t]
\centering
\small
\setlength{\tabcolsep}{4pt}
\renewcommand{\arraystretch}{1.15}
\begin{tabular}{l |c |cccc ccc c}
\toprule
\multirow{2}{*}{Model} & \multirow{2}{*}{Temperature} & \multicolumn{4}{c}{Step-1} & \multicolumn{3}{c}{Step-2} & \multirow{2}{*}{AVG} \\
\cmidrule(lr){3-6} \cmidrule(lr){7-9}
& & Fluency & Flexibility & Elaboration & Originality & Integrity & Focus & Adequacy & \\
\midrule
\multirow{3}{*}{qwen3.6-plus}
& 0.2 & 0 & \textbf{0.025} & 0 & \textbf{0.0094} & \textbf{0} & \textbf{0.025} & 0 & \textbf{0.0085} \\
& 0.5 & 0 & \textbf{0.025} & 0 & 0.0156 & 0.0125 & 0.05 & 0 & 0.0147 \\
& 0.8 & 0 & 0.125 & 0 & 0.0188 & 0.0375 & 0.075 & 0 & 0.0366 \\
\midrule
\multirow{3}{*}{deepseek-v4-pro}
& 0.2 & 0 & 0.025 & \textbf{0.0062} & \textbf{0.0062} & \textbf{0} & \textbf{0.125} & \textbf{0.1} & \textbf{0.0375} \\
& 0.5 & 0 & \textbf{0} & 0.0094 & 0.0281 & 0.0062 & 0.275 & 0.2 & 0.0741 \\
& 0.8 & 0 & 0.025 & 0.025 & 0.0094 & 0.0375 & 0.525 & 0.125 & 0.1067 \\
\midrule
\multirow{3}{*}{gemini-3.1-pro}
& 0.2 & 0 & 0 & 0.0094 & \textbf{0} & 0 & \textbf{0} & \textbf{0} & \textbf{0.0013} \\
& 0.5 & 0 & 0 & 0.025 & 0.0062 & 0 & \textbf{0} & 0.05 & 0.0116 \\
& 0.8 & 0 & 0 & \textbf{0.0031} & \textbf{0} & 0 & 0.05 & 0.1 & 0.0219 \\
\midrule
\multirow{3}{*}{gpt-5.4}
& 0.2 & 0 & \textbf{0} & 0.0062 & 0.0031 & \textbf{0} & \textbf{0.025} & 0 & \textbf{0.0049} \\
& 0.5 & 0 & 0.025 & 0.0062 & \textbf{0} & 0.0062 & \textbf{0.025} & 0 & 0.089 \\
& 0.8 & 0 & \textbf{0} & \textbf{0} & \textbf{0} & 0.0062 & 0.05 & 0 & 0.008 \\
\bottomrule
\end{tabular}
\caption{Variance results of repeated scoring under different temperature settings during Evidence-based Judging phase. \textbf{Bold} indicates the best (lowest) score across all temperature settings for each LLM.}
\label{Table:pre_experiment_variance}
\end{table*}

\begin{table*}[!t]
\centering

\begin{subtable}{\textwidth}
\centering
\resizebox{\textwidth}{!}{
\begin{tabular}{l|
cccc
ccc
cccc}
\toprule

\multirow{2}{*}{\textbf{Method}}
& \multicolumn{4}{c}{Step-1}
& \multicolumn{3}{c}{Step-2}
& \multicolumn{4}{c}{Step-3} \\

\cmidrule(lr){2-5}
\cmidrule(lr){6-8}
\cmidrule(lr){9-12}

& Fluency & Flexibility & Elaboration & Originality
& Integrity & Focus & Adequacy
& Fluency & Flexibility & Elaboration & Originality \\

\midrule

Direct Score
& \underline{0.5646} & 0.2243 & 0.2969 & 0.4969
& 0.2364 & 0.0649 & 0.0422
& 0.7296 & 0.3862 & 0.4211 & 0.3844 \\

CoT
& 0.4958 & 0.2548 & 0.3504 & 0.4459
& 0.2265 & 0.041 & 0.0452
& 0.7154 & 0.3521 & 0.4086 & 0.3939 \\

ToT
& 0.5521 & 0.2569 & 0.3367 & 0.4713
& 0.2014 & 0.0862 & 0.0434
& 0.7209 & 0.3824 & 0.4001 & 0.3679 \\

GoT
& 0.529 & 0.2519 & 0.3459 & 0.4529
& 0.1953 & 0.0444 & 0.0032
& \underline{0.7722} & 0.3936 & 0.3997 & 0.387 \\

TaT
& 0.428 & 0.2149 & 0.3186 & 0.4268
& 0.2556 & 0.0611 & 0.0283
& 0.6565 & 0.3425 & 0.4211 & 0.3481 \\

SaMer
& 0.2098 & 0.1523 & 0.316 & 0.4481
& 0.341 & 0.1425 & 0.1716
& 0.3342 & 0.2797 & 0.4164 & 0.2961 \\

SFT
& 0.5536 & \underline{0.487} & \underline{0.5642} & \underline{0.5827}
& \underline{0.563} & \underline{0.5502} & \underline{0.448}
& 0.5248 & \underline{0.5138} & \underline{0.5742} & \underline{0.5743} \\

CreaEval
& \textbf{0.7512$^{*}$} & \textbf{0.7045$^{*}$} & \textbf{0.7193$^{*}$} & \textbf{0.6994$^{*}$}
& \textbf{0.7865$^{*}$} & \textbf{0.5825} & \textbf{0.5575$^{*}$}
& \textbf{0.844$^{*}$} & \textbf{0.7359$^{*}$} & \textbf{0.7022$^{*}$} & \textbf{0.7243$^{*}$} \\

\bottomrule
\end{tabular}
}
\end{subtable}

\vspace{3mm}

\begin{subtable}{\textwidth}
\centering
\resizebox{\textwidth}{!}{
\begin{tabular}{l|
cc
c
cccccc
c}
\toprule

\multirow{2}{*}{\textbf{Method}}
& \multicolumn{2}{c}{Step-4}
& \multicolumn{1}{c}{Step-5}
& \multicolumn{6}{c}{Step-6}
& \multirow{2}{*}{AVG} \\

\cmidrule(lr){2-3}
\cmidrule(lr){4-4}
\cmidrule(lr){5-10}

& Correctly Written & Relevance
& Correctly Used
& Relevance & Effectiveness & Criteria & Impact & Humaneness & Development
& \\

\midrule

Direct Score
& 0.5091 & 0.2789
& 0.4962
& 0.1532 & 0.2075 & 0.2542 & 0.22 & 0.1665 & 0.5239
& 0.3329 \\

CoT
& 0.4114 & 0.2471
& 0.339
& 0.0954 & 0.2301 & 0.256 & 0.2079 & 0.1224 & 0.4861
& 0.3062 \\

TaT
& 0.424 & 0.2198
& 0.3814
& 0.1135 & 0.232 & 0.2828 & 0.2119 & 0.1309 & 0.5185
& 0.3167 \\

GoT
& 0.4786 & 0.2004
& 0.1762
& 0.0803 & 0.2371 & 0.3114 & 0.2345 & 0.1288 & \underline{0.5573}
& 0.309 \\

TaT
& 0.3453 & 0.2355
& 0.3887
& 0.0754 & 0.2215 & 0.2264 & 0.1648 & 0.1089 & 0.5032
& 0.2886 \\

SaMer
& 0.1742 & 0.2547
& 0.3984
& 0.1945 & 0.2081 & 0.2875 & 0.1984 & 0.1815 & 0.1557
& 0.258 \\

SFT
& \underline{0.5384} & \underline{0.5365}
& \underline{0.5958}
& \underline{0.5598} & \underline{0.51} & \underline{0.5503} & \underline{0.5534} & \underline{0.4245} & 0.5474
& \underline{0.5376} \\

CreaEval
& \textbf{0.8146$^{*}$} & \textbf{0.6111$^{*}$}
& \textbf{0.945$^{*}$}
& \textbf{0.6114} & \textbf{0.5519$^{*}$} & \textbf{0.5958$^{*}$} & \textbf{0.5685} & \textbf{0.5281$^{*}$} & \textbf{0.7144$^{*}$}
& \textbf{0.6874$^{*}$} \\

\bottomrule
\end{tabular}
}
\end{subtable}

\caption{Consistency (PCC) results across different methods and dimensions. \textbf{Bold} indicates the highest score and \underline{underline} indicates the second highest score. The AVG column summarizes the overall average score. The asterisk ($^{*}$) marks statistically significant improvements (p < 0.05, t-test) over the second-best method.}
\label{Table:PCC_results}
\end{table*}

\begin{table*}[!t]
\centering

\begin{subtable}{\textwidth}
\centering
\resizebox{\textwidth}{!}{
\begin{tabular}{l|
cccc
ccc
cccc}
\toprule

\multirow{2}{*}{\textbf{Method}}
& \multicolumn{4}{c}{Step-1}
& \multicolumn{3}{c}{Step-2}
& \multicolumn{4}{c}{Step-3} \\

\cmidrule(lr){2-5}
\cmidrule(lr){6-8}
\cmidrule(lr){9-12}

& Fluency & Flexibility & Elaboration & Originality
& Integrity & Focus & Adequacy
& Fluency & Flexibility & Elaboration & Originality \\

\midrule

Direct Score
& \underline{0.5539} & 0.1778 & 0.1079 & 0.3571
& 0.1937 & 0.0462 & 0.0339
& 0.7149 & 0.2975 & 0.3187 & 0.3171 \\

CoT
& 0.4761 & 0.1787 & 0.1303 & 0.2994
& 0.1761 & 0.0334 & 0.0377
& 0.7031 & 0.2757 & 0.3259 & 0.3203 \\

ToT
& 0.5277 & 0.1751 & 0.1275 & 0.3166
& 0.1596 & 0.0648 & 0.0351
& 0.7084 & 0.3024 & 0.3165 & 0.2993 \\

GoT
& 0.5061 & 0.1785 & 0.1285 & 0.3379
& 0.1724 & 0.0299 & 0.0029
& \underline{0.7474} & 0.3199 & 0.292 & 0.3254 \\

TaT
& 0.4111 & 0.1861 & 0.1341 & 0.2909
& 0.1961 & 0.0423 & 0.0174
& 0.623 & 0.3036 & 0.3317 & 0.265 \\

SaMer
& 0.1887 & 0.1709 & 0.28 & 0.2765
& 0.301 & 0.1217 & 0.1532
& 0.3034 & 0.2731 & 0.3803 & 0.2895 \\

SFT
& 0.4801 & \underline{0.4012} & \underline{0.4602} & \underline{0.5049}
& \underline{0.445} & \underline{0.4311} & \underline{0.41}
& 0.479 & \underline{0.3957} & \underline{0.4865} & \underline{0.4284} \\

CreaEval
& \textbf{0.7514$^{*}$} & \textbf{0.6675$^{*}$} & \textbf{0.6571$^{*}$} & \textbf{0.6665$^{*}$}
& \textbf{0.7765$^{*}$} & \textbf{0.5565$^{*}$} & \textbf{0.4844}
& \textbf{0.8324$^{*}$} & \textbf{0.6986$^{*}$} & \textbf{0.6967$^{*}$} & \textbf{0.6975$^{*}$} \\

\bottomrule
\end{tabular}
}
\end{subtable}

\vspace{3mm}

\begin{subtable}{\textwidth}
\centering
\resizebox{\textwidth}{!}{
\begin{tabular}{l|
cc
c
cccccc
c}
\toprule

\multirow{2}{*}{\textbf{Method}}
& \multicolumn{2}{c}{Step-4}
& \multicolumn{1}{c}{Step-5}
& \multicolumn{6}{c}{Step-6}
& \multirow{2}{*}{AVG} \\

\cmidrule(lr){2-3}
\cmidrule(lr){4-4}
\cmidrule(lr){5-10}

& Correctly Written & Relevance
& Correctly Used
& Relevance & Effectiveness & Criteria & Impact & Humaneness & Development
& \\

\midrule

Direct Score
& \underline{0.4848} & 0.0799
& 0.4635
& 0.0747 & 0.122 & 0.1497 & 0.1109 & 0.0759 & 0.1721
& 0.2426 \\

CoT
& 0.3734 & 0.102
& 0.2839
& 0.0329 & 0.1095 & 0.165 & 0.1041 & 0.055 & 0.1832
& 0.2183 \\

ToT
& 0.3804 & 0.1001
& 0.3131
& 0.0417 & 0.1037 & 0.1641 & 0.0979 & 0.0546 & 0.1821
& 0.2235 \\

GoT
& 0.4562 & 0.044
& 0.083
& 0.0203 & 0.1081 & 0.1839 & 0.1144 & 0.0562 & 0.1829
& 0.2145 \\

TaT
& 0.3199 & 0.0676
& 0.3716
& 0.0325 & 0.1113 & 0.1285 & 0.0742 & 0.042 & 0.1507
& 0.205 \\

SaMer
& 0.1659 & 0.2198
& 0.394
& 0.1936 & 0.2067 & 0.1967 & 0.1604 & 0.1808 & 0.1443
& 0.23 \\

SFT
& 0.4637 & \underline{0.4211}
& \underline{0.4733}
& \underline{0.4175} & \underline{0.4513} & \underline{0.3975} & \underline{0.4153} & \underline{0.4124} & \underline{0.4596}
& \underline{0.4417} \\

CreaEval
& \textbf{0.794$^{*}$} & \textbf{0.536$^{*}$}
& \textbf{0.9441$^{*}$}
& \textbf{0.538$^{*}$} & \textbf{0.5225$^{*}$} & \textbf{0.5289$^{*}$} & \textbf{0.4966$^{*}$} & \textbf{0.4403} & \textbf{0.5131$^{*}$}
& \textbf{0.6399$^{*}$} \\

\bottomrule
\end{tabular}
}
\end{subtable}

\caption{Consistency (ICC) results across different methods and dimensions. \textbf{Bold} indicates the highest score and \underline{underline} indicates the second highest score. The AVG column summarizes the overall average score. The asterisk ($^{*}$) marks statistically significant improvements (p < 0.05, t-test) over the second-best method.}
\label{Table:ICC_results}
\end{table*}

\begin{table*}
    \centering
    \begin{tabular}{>{\footnotesize}p{1.5cm}>{\footnotesize}p{5.5cm}>{\footnotesize}p{6cm}>{\footnotesize}p{1cm}}
        \toprule
        \textbf{Dimension} & \textbf{Description} & \textbf{Scoring Rubrics} & \textbf{Score Range} \\
        \midrule
        Fluency & 
        Whether each “challenge” is clearly articulated, semantically unambiguous, and able to establish a reasonable causal relationship with the future scenario.

The reasons for unreasonable challenges are as follows:

\begin{enumerate}[itemsep=0pt, parsep=0pt, topsep=1pt]
    \item Perhaps: The statement is vague or semantically unclear, making its intended meaning difficult to determine. 
    \item Why: It is unrelated to the future scenario or lacks a reasonable connection to it.
    \item Solution: The content describes a solution rather than a challenge itself.
    \item Duplicate: It is redundant or semantically equivalent to an existing “Yes” challenge.
    \item Blank: No valid content is provided.
\end{enumerate} & 
\textbf{0:} Unreasonable Challenge

\textbf{1:} Reasonable Challenge & 0-8 \\
        Flexibility & The number of distinct types represented by all reasonable challenges. & \textbackslash & 0-8 \\
        Elaboration & The complexity of the information expansion structure of each valid challenge, focusing on the information density and level of detail in the description. & 
        \textbf{0:} The description is unclear or incomplete, failing to specify the challenge or lacking relevance to the scenario.

\textbf{1:} The description is present but lacks sufficient detail, or the connection to the scenario is not fully explained.

\textbf{2:} The description is clear and complete, explicitly stating the challenge and its significance, and establishing a clear logical connection with the future scenario. & 0-16 \\
        Originality & Whether each challenge deviates from common or easily anticipated typical problem formulations, reflecting a non-typical perspective or uncommon problem construction. & 
        \textbf{0:} The challenge lacks novelty and is conventional or template-like in content.

\textbf{1:} The challenge shows some originality but remains relatively common or only moderately innovative.

\textbf{2:} The challenge demonstrates clear uniqueness or a novel perspective, with strong novelty. & 0-16 \\
        
        \bottomrule
    \end{tabular}
    \caption{Detailed evaluation dimensions and scoring rubrics for Step-1 on CGPST. Step-1 requires item-wise scoring of challenges, with up to 8 challenges. Therefore, the maximum score for \textit{Fluency} is 1×8=8, and similarly for other dimensions.}
    \label{Table:Step-1 Rubrics}
\end{table*}

\begin{table*}
    \centering
    \begin{tabular}{>{\footnotesize}p{1.5cm}>{\footnotesize}p{4.3cm}>{\footnotesize}p{7.2cm}>{\footnotesize}p{1cm}}
        \toprule
        \textbf{Dimension} & \textbf{Description} & \textbf{Scoring Rubrics} & \textbf{Score Range} \\
        \midrule
        Integrity: Condition Phrase & 
        The condition phrase used to connect the future scenario within the underlying problem, which reflects how the problem is linked to the future context. & 
\textbf{0:} No condition phrase is included; the underlying problem is not linked to the future scenario.

\textbf{1:} The future-scenario information is inaccurate or not associated with the key verb phrase.

\textbf{2:} Accurate information from the future scenario is used and is properly connected to the key verb phrase. & 0-2 \\
        Integrity: Stem \& KVP & The stem is typically “How can we” or “In what way can we”, and the key verb phrase (KVP) appears after the stem, expressed as a verb–object structure (limited to one active verb and one object) to represent the core action. &         
        \textbf{0:} The key verb phrase is not provided.
        
\textbf{1:} A key verb phrase exists but contains multiple unrelated active verbs.

\textbf{2:} A key verb phrase exists but contains multiple objects or modifiers.

\textbf{3:} The key verb phrase contains only one clear active verb. & 0-3 \\
        Integrity: Purpose & The purpose or intention implied or explicitly stated in the underlying problem, indicating the direction the problem aims to address or focus on. It should contain only one purpose. & 
        \textbf{0:} No purpose is expressed.
        
\textbf{1:} Multiple purposes are present, or the purpose overlaps with the key verb phrase.

\textbf{2:} A purpose is present but has no clear logical connection with the key verb phrase.

\textbf{3:} A single, clearly defined purpose is present and is reasonably related to the key verb phrase. & 0-3 \\
        Integrity: Future Scene Parameters & Whether the underlying problem reflects the core informational elements of the future scenario, including theme, location, and time as three key scenario parameters. & 
        \textbf{0:} 0 or 1 parameter is reflected.
        
\textbf{1:} 2 parameters are reflected.

\textbf{2:} All three parameters—theme, location, and time—are clearly reflected. & 0-2 \\

Focus & Whether the underlying problem has a clear action structure in its overall expression, including whether the condition phrase, key verb phrase, and purpose meet the required criteria. & 
        \textbf{1/2/3:} The future scenario is restated, generalized, or ignored; there is no purpose or it is unrelated to the key verb phrase, or the purpose is redundant with the key verb phrase or condition phrase.
        
\textbf{4/5/6:} The key verb phrase and purpose are overly broad or overly narrow; the core problem is unclear; or multiple key verb phrases or purposes are included.

\textbf{7/8:} The core problem contains a well-formed key verb phrase; the purpose is clear and responds to the future scenario task.

\textbf{9/10:} Excellent key verb phrase, closely aligned with a clear purpose, and strongly responsive to the future scenario task. & 1-10 \\

Adequacy & The scope of impact and level of criticality of the underlying problem within the future scenario, reflecting its relative importance in the overall context. & 
       \textbf{1/2/3:} The future scenario is restated, generalized, or ignored; there is no purpose or it is unrelated to the key verb phrase, or the purpose is redundant with the key verb phrase or conditional phrase.
        
\textbf{4/5/6:} Identifies a secondary issue within the future scenario.

\textbf{7/8:} Identifies an appropriate issue within the future scenario.

\textbf{9/10:} Identifies a major and highly important issue within the future scenario. & 1-10 \\
        
        \bottomrule
    \end{tabular}
    \caption{Detailed evaluation dimensions and scoring rubrics for Step-2 on CGPST. \textit{Integrity} consists of four sub-dimensions: \textit{Condition Phrase}, \textit{Stem \& KVP}, \textit{Purpose}, and \textit{FS Parameters}. Each sub-dimension must be evaluated during scoring. Therefore, the score for \textit{Integrity} ranges from 0 to 10.}
    \label{Table:Step-2 Rubrics}
\end{table*}

\begin{table*}
    \centering
    \begin{tabular}{>{\footnotesize}p{1.5cm}>{\footnotesize}p{5.5cm}>{\footnotesize}p{6cm}>{\footnotesize}p{1cm}}
        \toprule
        \textbf{Dimension} & \textbf{Description} & \textbf{Scoring Rubrics} & \textbf{Score Range} \\
        \midrule
        Fluency & 
        Whether each solution can establish a clear correspondence with the underlying problem in Step-2 and semantically respond to the key action direction.

Invalid solution reasons are as follows:

\begin{enumerate}[itemsep=0pt, parsep=0pt, topsep=1pt]
    \item Perhaps — The relationship between the solution and the key verb phrase and purpose is unclear.  
    \item Why — The solution is unrelated to the potential problem.
    \item Duplicate — The solution is overly similar to another “Yes” solution.
    \item Blank: No valid content is provided.
\end{enumerate} & 
\textbf{0:} Invalid Solution

\textbf{1:} Valid Solution & 0-8 \\
        Flexibility & The number of distinct types represented by all valid solutions. & \textbackslash & 0-8 \\
        Elaboration & The complexity of the information expansion structure of each valid solution, focusing on the information density and level of detail in the description. & 
        \textbf{0:} The description is unclear or incomplete, failing to specify the solution or lacking relevance to the scenario.

\textbf{1:} The description is present but lacks sufficient detail, or the connection to the scenario is not sufficiently explained.

\textbf{2:} The description is clear and complete, explicitly stating the solution and its significance, and establishing a clear logical connection with the future scenario. & 0-16 \\
        Originality & Whether each solution deviates from common or easily anticipated classical solution approaches, reflecting non-traditional or non-obvious problem-solving strategies. & 
        \textbf{0:} The solution lacks novelty and is conventional or template-like in content.

\textbf{1:} The solution shows some originality but remains relatively common or only moderately innovative.

\textbf{2:} The solution demonstrates clear uniqueness or a novel perspective, with strong novelty. & 0-16 \\
        
        \bottomrule
    \end{tabular}
    \caption{Detailed evaluation dimensions and scoring rubrics for Step-3 on CGPST. Step-3 requires item-wise scoring of solutions, with up to 8 solution. Therefore, the maximum score for \textit{Fluency} is 1×8=8, and similarly for other dimensions.}
    \label{Table:Step-3 Rubrics}
\end{table*}

\begin{table*}
    \centering
    \begin{tabular}{>{\footnotesize}p{1.5cm}>{\footnotesize}p{5.5cm}>{\footnotesize}p{6cm}>{\footnotesize}p{1cm}}
        \toprule
        \textbf{Dimension} & \textbf{Description} & \textbf{Scoring Rubrics} & \textbf{Score Range} \\
        \midrule
        Correctly Written & 
        The structural characteristics of the evaluation criteria in its expression form, which should satisfy all of the following four conditions simultaneously:

\begin{enumerate}[itemsep=0pt, parsep=0pt, topsep=1pt]
    \item Whether superlative expressions are used (e.g., “most ...”);   
    \item Whether a single optimization objective is clearly specified;
    \item Whether the desired direction is explicitly stated; 
    \item Whether it is formulated in the form of a question.
\end{enumerate} & 
\textbf{0:} Fails to satisfy at least one of the above conditions.

\textbf{1:} All of the above conditions are met. & 0-5 \\
        Relevance & The degree of semantic relevance between the evaluation criteria and the underlying problem identified in Step-2, i.e., whether the criteria are designed around the core focus or key aspects of the problem. & \textbf{0:} The evaluation criteria are unrelated to the underlying problem or merely repeat it.
        
\textbf{1:} The evaluation criteria are overly general and non-specific, applicable to many types of problems.

\textbf{2:} The evaluation criteria are relatively specific but still have room for improvement.

\textbf{3:} The evaluation criteria are clear, specific, and highly relevant to the underlying problem. & 0-15 \\
        
        \bottomrule
    \end{tabular}
    \caption{Detailed evaluation dimensions and scoring rubrics for Step-4 on CGPST. Step-4 requires item-wise scoring of criteria, with 5 criteria. Therefore, the maximum score for \textit{Correctly Written} is 1×5=5, and similarly for \textit{Relevance}.}
    \label{Table:Step-4 Rubrics}
\end{table*}

\begin{table*}
    \centering
    \begin{tabular}{>{\footnotesize}p{2cm}>{\footnotesize}p{6cm}>{\footnotesize}p{5cm}>{\footnotesize}p{1cm}}
        \toprule
        \textbf{Dimension} & \textbf{Description} & \textbf{Scoring Rubrics} & \textbf{Score Range} \\
        \midrule
        Correctly Used & 
        For each evaluation criterion, the scores of all solutions must form a non-repeating set of integers from 1 to x (where x is the number of solutions), and scoring should be based on the number of errors in the evaluation grid. & 

\textbf{1:} The grid contains four or more errors.

\textbf{2:} The grid contains three errors. 

\textbf{3:} The grid contains two errors.

\textbf{4:} The grid contains one error.

\textbf{5:} The grid contains no errors.

& 1-5 \\
        
        \bottomrule
    \end{tabular}
    \caption{Detailed evaluation dimensions and scoring rubrics for Step-5 on CGPST.}
    \label{Table:Step-5 Rubrics}
\end{table*}

\begin{table*}
    \centering
    \begin{tabular}{>{\footnotesize}p{1.5cm}>{\footnotesize}p{5.5cm}>{\footnotesize}p{6cm}>{\footnotesize}p{1cm}}
        \toprule
        \textbf{Dimension} & \textbf{Description} & \textbf{Scoring Rubrics} & \textbf{Score Range} \\
        \midrule
        Relevance & 
        The degree of semantic correspondence between the action plan and the underlying problem identified in Step-2, i.e., whether the plan is specifically designed to address and respond to the problem. & 

        \textbf{1:} The action plan does not address the underlying problem.
        
\textbf{2/3:} The action plan is somewhat related to the underlying problem, but better alternatives may exist.

\textbf{4:} The action plan responds well to the underlying problem.

\textbf{5:} The action plan is highly relevant to the underlying problem.

& 1-5 \\

Effectiveness & 
        The potential problem-solving effectiveness of the action plan in terms of logical reasoning, including the extent to which it covers key aspects of the problem and the completeness of its solution pathway. & 

        \textbf{1:} The action plan can hardly solve the underlying problem.
        
\textbf{2/3:} The action plan addresses only part of the underlying problem.

\textbf{4:} The action plan is able to sufficiently address most aspects of the underlying problem.

\textbf{5:} The action plan can comprehensively and effectively resolve the underlying problem.

& 1-5 \\

Criteria & 
        The degree of correspondence between the action plan and the evaluation criteria generated in Step-4, i.e., whether the plan reflects the core elements emphasized by these criteria. & 

        \textbf{1:} The action plan does not reflect any of the evaluation criteria.
        
\textbf{2/3:} The connection between the action plan and the evaluation criteria is weak or unclear.

\textbf{4:} The action plan establishes clear and reasonable links with some of the evaluation criteria.

\textbf{5:} The action plan effectively addresses all evaluation criteria in a clear and comprehensive manner.

& 1-5 \\

Impact & 
        The potential positive impact direction of the action plan within the future scenario, including the scope and direction of its influence on systems, environments, or relevant stakeholders. & 

        \textbf{1:} The action plan has no noticeable impact on the future scenario.
        
\textbf{2/3:} The action plan has a limited impact on the future scenario.

\textbf{4:} The action plan has a certain positive impact on the future scenario.

\textbf{5:} The action plan has a significant and positive impact on the future scenario.

& 1-5 \\

Humaneness & 
        Whether the action plan reflects a human-centered value orientation, including attention to human well-being, safety, development, or positive social values. & 

        \textbf{1:} The action plan has a negative or destructive orientation.
        
\textbf{2/3:} The action plan is neutral, with neither positive nor negative effects.

\textbf{4:} The action plan shows some constructive and positive potential.

\textbf{5:} The action plan clearly demonstrates human-centered care and is positive and highly constructive.

& 1-5 \\

Development & 
        The degree of semantic correspondence between the action plan and the underlying problem identified in Step-2, i.e., whether the plan is specifically designed to address and respond to the problem. & 

        \textbf{1/2/3:} The description is extremely brief, merely repeating the solutions from Step-3.
        
\textbf{4/5/6:} The action plan is somewhat developed but lacks sufficient supporting details.

\textbf{7/8:} Clearly explains key elements of the action plan, including “who does what, why, and how,” with some supporting details.

\textbf{9/10:} The structure is clear and highly elaborated, going far beyond basic elements, with strong executability and completeness.

& 1-10 \\
        
        \bottomrule
    \end{tabular}
    \caption{Detailed evaluation dimensions and scoring rubrics for Step-6 on CGPST.}
    \label{Table:Step-6 Rubrics}
\end{table*}

\end{document}